\documentclass[acmsmall]{acmart}
\usepackage[explicit]{titlesec}
\usepackage{subfigure}
\usepackage{colortbl}
\usepackage{xcolor}
\usepackage{pifont}
\usepackage{footmisc}
\usepackage{amsmath, amsfonts}
\usepackage[linesnumbered,ruled,vlined]{algorithm2e}
\SetKwInOut{Input}{Input}
\SetKwInOut{Output}{Output}

\SetCommentSty{mycommfont}
\DeclareMathOperator*{\argmax}{arg\,max}

\usepackage{multirow}

\definecolor{answercolor}{RGB}{240, 240, 240}
\AtBeginDocument{%
  \providecommand\BibTeX{{%
    \normalfont B\kern-0.5em{\scshape i\kern-0.25em b}\kern-0.8em\TeX}}}

\setcopyright{acmcopyright}
\copyrightyear{2018}
\acmYear{2018}
\acmDOI{XXXXXXX.XXXXXXX}

\acmJournal{JACM}
\acmVolume{37}
\acmNumber{4}
\acmArticle{0}
\acmMonth{8}

\SetCommentSty{mycommfont}
\definecolor{answercolor}{RGB}{240, 240, 240}

\begin{document}

\title{LaF: Labeling-Free Model Selection for Automated Deep Neural Network Reusing}
\author{Qiang Hu $^*$}
\affiliation{%
  \institution{SnT, University of Luxembourg}
  \country{Luxembourg}}

\author{Yuejun Guo$^+\ ^*$}\thanks{$^*$ Equal Contribution. $^+$Work partially done while at SnT, University of Luxembourg.}
\affiliation{%
  \institution{Luxembourg Institute of Science and Technology}
  \country{Luxembourg}}

\author{Xiaofei Xie}
\affiliation{%
 \institution{Singapore Management University}
  \country{Singapore}}
\author{Maxime Cordy}
\affiliation{%
 \institution{SnT, University of Luxembourg}
  \country{Luxembourg}}
\author{Mike Papadakis}
\affiliation{%
  \institution{SnT, University of Luxembourg}
  \country{Luxembourg}}
\author{Yves Le Traon}
\affiliation{%
  \institution{SnT, University of Luxembourg}
  \country{Luxembourg}}

\renewcommand{\shortauthors}{Hu and Guo et al.}


\begin{abstract}

Applying deep learning to science is a new trend in recent years which leads DL engineering to become an important problem. Although training data preparation, model architecture design, and model training are the normal processes to build DL models, all of them are complex and costly. Therefore, reusing the open-sourced pre-trained model is a practical way to bypass this hurdle for developers. Given a specific task, developers can collect massive pre-trained deep neural networks from public sources for re-using. However, testing the performance (e.g., accuracy and robustness) of multiple DNNs and recommending which model should be used is challenging regarding the scarcity of labeled data and the demand for domain expertise.  In this paper, we propose a labeling-free (LaF) model selection approach to overcome the limitations of labeling efforts for automated model reusing. The main idea is to statistically learn a Bayesian model to infer the models' specialty only based on predicted labels. We evaluate LaF using 9 benchmark datasets including image, text, and source code, and 165 DNNs, considering both the accuracy and robustness of models. The experimental results demonstrate that LaF outperforms the baseline methods by up to 0.74 and 0.53 on Spearman's correlation and Kendall's $\tau$, respectively.

\end{abstract}




\maketitle

\section{Introduction}
\label{sec:intro}

Deep learning (DL) is helping solve all sorts of real-world problems in various domains, such as computer vision~\cite{Vision2020Mahony}, natural language processing (NLP)~\cite{text2021Minaee}, code understanding~\cite{codenet2021}, and autonomous driving~\cite{huang2018apolloscape}. Due to the outstanding performance of deep neural networks (DNNs), software engineering (SE) researchers have attempted to apply DNNs to solve various SE tasks, such as source code processing~\cite{codenet2021, alon2019code2vec}, automatic software testing~\cite{9402046, 8952543}, and GUI designs~\cite{zhao2021guigan}. Building machine learning (or deep learning) systems generally requires model architecture design and data preparation, model training, and model deployment and maintenance, known as machine learning operations (MLOps)~\cite{mlops}. However, since 1) designing new DNN architectures requires tremendous experimentation and DL knowledge, 2) preparing massive labeled training data is labor-intensive, and 3) training the DL model needs a considerable amount of time and resources, developers generally reuse generic and publicly available pre-trained models to solve their given task. Ultimately, we expect model reuse to become the norm in DL-based SE, just like it has been in other fields like NLP.

Due to the convenience of multiple public open sources such as GitHub~\cite{github2022}, engineers can today access a massive number of DNNs, in the form of either pre-trained model files (e.g., .h5 and .pth). While this is practical, there is no prior evidence of which model will be capable of solving the targeted task the most effectively. Indeed, different models have been developed by different contributors and in different development settings, while they have been evaluated in unknown or incomparable testing conditions. For example, only 21 of the 165 DNNs we collected and evaluated in this paper have a performance report.

Model selection is a process of determining the best fit from multiple models for a given test set. Selecting candidate models for the targeted task raises two challenges. First, test data that can be found in the real world (e.g. source code from public repositories) are generally unlabelled~\cite{sun2017unlabel}, while data labeling requires significant manual effort. For example, the AIZU online programming challenge~\cite{aoj2018} receives submissions in different programming languages all the time. A Java developer can easily annotate the source code in Java but may have difficulties with other programming languages, such as C++. \emph{The effective selection of pre-trained models to solve various tasks, therefore, requires a precise and efficient method to \textbf{compare and rank} DNN candidates \textbf{without} data labels}.

An additional challenge originates from the fact that the chosen models will likely be confronted with data that have a different distribution from the data they were trained/tested on, i.e. out-of-distribution (OOD) data~\cite{Cats2020david,hu2021understanding,dola2021distribution}. For many application domains -- including software~\cite{wilds2021} --  the distribution shift phenomena that yield OOD data are inevitable. As a result, these models may exhibit remarkably different performances over time, which raises the concern of quality and reliability~\cite{book2020alexander}. For instance, for the same dataset iWildCam (please refer to Section~\ref{subsec:data} for more details), two DNNs exhibit 75.74\% and 76.60\% accuracy on the initial test data, while their performance turns to 76.82\% and 65.30\%, respectively, on OOD data. In the end, this implies that \emph{ideal selection methods should reliably estimate model performance on OOD data as well}.

Such methods have been proposed by a recent approach named sample discrimination-based selection (SDS)~\cite{sds2021meng}. SDS achieves positive model ranking results on three benchmark datasets. SDS selects a set of data to label based on the majority voting~\cite{omer2018voting} and item discrimination~\cite{robert1954item}. These data are considered the most discriminative in terms of distinguishing the accuracy between DNNs. DNNs are then ranked based on their accuracy on these selected and labeled data. Figure~\ref{fig:problem} illustrates how this sampling-based approach works. Although promising, SDS still suffers from manual labeling. Besides, it has only considered ranking based on model accuracy with in-distribution (ID) data (i.e. data that follow the same distribution as the data the models were trained on) -- it has disregarded OOD data. Third, it has been applied to the image domain only; its effectiveness for other domains remains unclear.

In this paper, we aim to overcome the above limitations and propose LaF, a labeling-free model selection method, for DNNs that is effective with both ID and OOD data. Given a sample, only the predicted labels of multiple DNNs are available. The main idea of LaF is to build a Bayesian model that incorporates the data difficulty and model specialty to estimate the likelihood of a predicted label being the true label. The data difficulty implies how difficult a sample is for all DNNs to predict correctly, which is reflected by the prediction difference across multiple models. The model specialty indicates how good a model is to infer the correct labels of all samples, which is reflected by the ability to have the same predictions as the majority of models. Via optimizing the Bayesian model, we infer the model specialty to perform the model selection. The optimization is achieved by the expectation-maximization (EM) algorithm~\cite{dempster1977em}, which is efficient in finding maximum likelihood parameters (the data difficulty and model specialty in our case). To summarize, the main contributions of our work are:
\begin{enumerate}
   \item We propose a novel approach, LaF, for automatically ranking multiple DNN models to facilitate the reuse of DNNs from public sources. 
    \item We demonstrate the effectiveness of LaF on ID and OOD data. Both the artificial and natural distribution shifts are considered.
    \item LaF is labeling-free, which makes it practical and feasible in real-world applications.
    \item We experiment on 9 benchmark datasets spanning different domains including image, text, and source code with different programming languages (Java and C++). To the best of our knowledge, this is the first DNN model selection work containing datasets other than images.
    \item All the used models, datasets, and implementations of LaF and baseline methods are publicly available at \url{https://github.com/testing-cs/LaF-model-selection.git}.
\end{enumerate}

The rest of this paper is organized as follows. Section~\ref{sec:back} introduces preliminary knowledge behind this work. Section~\ref{sec:methodology} presents the problem statement and our proposed solution. Section~\ref{sec:design} explains our experimental design. We present the empirical results and corresponding discussions in Section~\ref{sec:results}. Section~\ref{sec:threats} details the threats that may affect the validity of conclusions. Section~\ref{sec:related} reviews related work. The last section concludes our work and points out future research directions.

\section{Background}
\label{sec:back}

\subsection{MLOps and Model Reusing}
\begin{figure*}[h]
    \centering
    \includegraphics[scale=0.42]{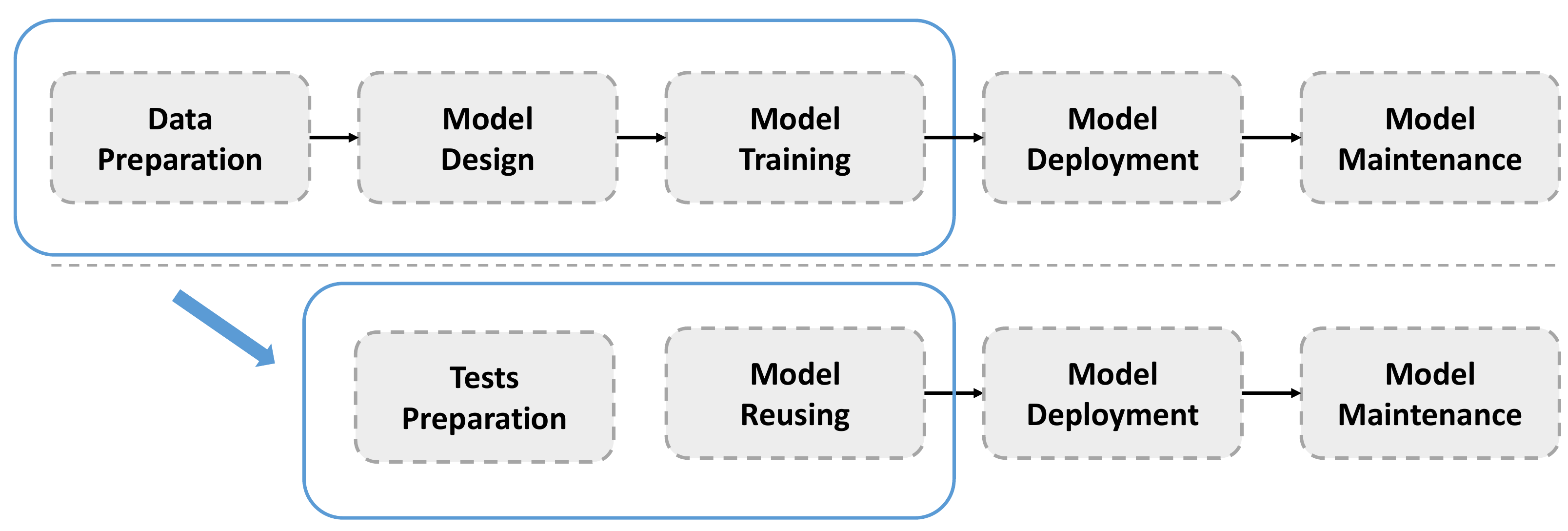}
    \caption{MLOps with and without model reusing.}
    \label{fig:mlops}
\end{figure*}

Generally, building machine learning systems needs a set of operations~\cite{mlops} that are depicted in the first row of Figure~\ref{fig:mlops}. Roughly speaking, first, developers connect their interested data and design the model architecture that is suitable to train the data. Then, developers fed the model and data into the high-performance server to tune the parameters of the model. After a model with expected performance was trained, it will be deployed~~(embedded) into the application and function in the wild. Finally, similar to conventional software systems, machine learning systems also need to be evolved and maintained from time to time. For the first two steps, i.e., data preparation and model design, huge human efforts and expert domain knowledge are needed to label the data and design the model. And for the model training process, expensive computing resource is necessary to handle the complex parameter tuning procedure. In conclusion, the first three steps make the whole process heavy before the model is deployed for real usage.

In practice, model reusing is a commonly adopted way to lighten machine learning operations (MLOps). As shown in the second row of Figure~\ref{fig:mlops}, the original three operations data preparation, model design, and model training are replaced by the two, tests preparation and model reusing. The hidden reason for this replacement is that, nowadays, given a specific task, e.g., face recognition, we can access many well-established pre-trained models from online resources, e.g., GitHub. Thus, the straightforward way to build our machine learning system is to reuse such models. To do so, we only need to prepare some test data (which can be much less than the training data) for our task, then collect potential pre-trained models, and finally, select the best model using the test data to build our system. In this way, we can reduce the human effort the training data labeling and the difficult procedure of model architecture design. In this work, we focus on how to efficiently select the best model from the massive number of available models.

\subsection{Comparison testing and test selection in deep learning}
\label{subsec:comparison}
In conventional software engineering, comparison testing~\cite{selvapriya2013different,kavitha2014software,sawant2012soft} aims at figuring out the strength and weaknesses of a newly developed software product compared with existing products. The end goal is to facilitate the deployment of a product with high functionality and reliability. Recently, Meng \emph{et al.}~\cite{sds2021meng} re-framed ``comparison testing'' as testing methods that aim to compare alternative software artifacts, especially DNNs~\cite{jovanovic2006software}. Concretely, the problem turns into how to find out the most discriminative data that can amply distinguish the difference. In their proposed sample discrimination-based selection method, the majority voting~\cite{omer2018voting} is first applied to produce pseudo labels based on which DNNs are classified to top, middle, and bottom groups following the item discrimination~\cite{robert1954item}. Via the prediction difference between the top and bottom DNNs, each data has a unique discrimination score and the high ones are selected for the final ranking.

A close topic to comparison testing is test selection for DNN model performance estimation. The key idea of this kind of test selection is, given massive unlabeled test data and a DNN model, we estimate the performance of the model on these data by using a subset of data selected by test selection metrics. In this way, the labeling effort can be reduced and the budget can be saved. For instance, Li \emph{et al.}~\cite{ces2019li} proposed the cross Entropy-based sampling method to identify the most representative data of a test set. Similarly, Chen \emph{et al.} \cite{chen2020pace} developed the practical accuracy estimation. The difference is that in test selection, the objective is a single DNN, while in comparison testing, the objective is multiple DNNs. Undoubtedly, one can first approximate the performance of each DNN by selecting its corresponding representative set and then undertake the comparison. However, this will largely increase the effort in labeling and is less practical than selecting once.

Both comparison testing and test selection can be used in the model reusing process to indicate the model with the best performance. 

\subsection{Distribution Shift}
\label{subsec:testing}

Distribution shift is a crucial problem in machine learning which means the train data and test data follow different data distributions. Generally, compared to the in-distribution data~(IID), the model is more difficult to handle the data with distribution shift, which makes the reported performance (using IID) of the model unreliable.

\begin{figure*}
    \centering
    \subfigure[White noise]{
    \includegraphics[scale=0.5]{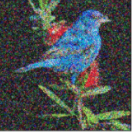}
    }
    \subfigure[Brightness]{
    \includegraphics[scale=0.5]{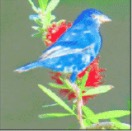}
    }
    \subfigure[Location 1]{
    \includegraphics[scale=0.42]{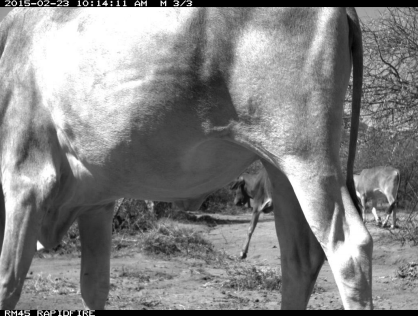}
    }
    \subfigure[Location 2]{
    \includegraphics[scale=0.42]{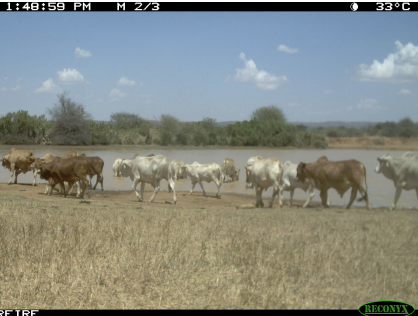}
    \label{fig:clean-camera}
    }
    \caption{Image examples with artificial distribution shifts (noise, brightness, location change).}
    \label{fig:ds_example}
\end{figure*}

Roughly speaking, there are two types of distribution shifts, artificial and natural. Artificial distribution shift mainly comes from adding artificial perturbations (corruptions) into raw data. Dan and Thomas \cite{cifar10c2019dan} proposed to add 15 types of algorithmically generated corruptions with 5 levels of severity to image data to mimic realistic situations, such as noise, blur, snow, and zoom. Based on these corruptions, different benchmark datasets, such as CIFAR-10-C \cite{cifar10c2019dan} and MNIST-C \cite{mnistc2019norman}, have been developed for testing the robustness of DNN models. The first two figures in Figure~\ref{fig:ds_example} show two examples of artificial distribution shift - the original bird picture with two types of noises, white noise and brightness. On the other hand, natural distribution shift is usually induced by the change of environment or population and exists in raw data, such as the change of camera traps \cite{beery2020iwildcam}, new customers \cite{ni2019justifying}, and new repositories \cite{wilds2021}. A recent benchmark \cite{wilds2021} provides in-the-wild distribution shifts covering diverse data domains and applications. The last two figures in Fig~\ref{fig:ds_example} are examples of natural distribution shift. The two pictures have the same label \textit{cow} but the cows are captured by different positions. 

\section{Methodology}
\label{sec:methodology}

\subsection{Problem Formulation}
\label{subsec:problem}
\begin{figure*}[h]
    \centering
    \includegraphics[scale=0.52]{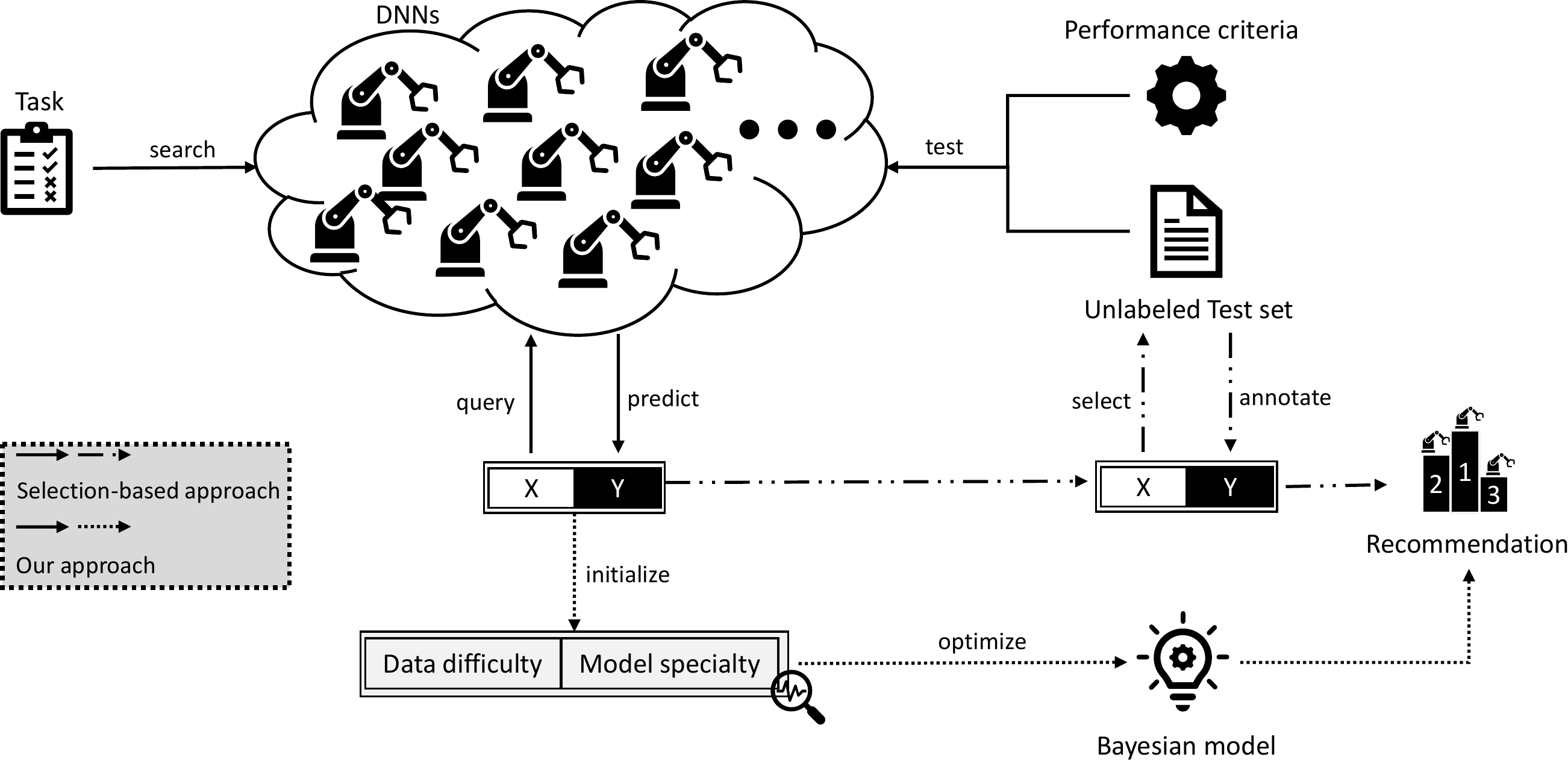}
    \caption{Illustration of studied problem.}
    \label{fig:problem}
\end{figure*}

In this paper, we are interested in the classification task. Given a $C$-class task over a sample space $\mathcal{Z}=\mathcal{X}\times\mathcal{Y}\rightarrow\mathcal{R}$, where $x\in\mathcal{X}$ is an input data and $y\in\mathcal{Y}$ is its class label. Let $f: x\rightarrow y$ be a deep neural network (DNN) that maps $x$ to the problem domain. Given $n$ models, $f_1,f_2,\ldots,f_n$, extracted from public sources and a set of unlabeled test data $T$, the problem we study is to estimate the rank of models regarding their performance on ${\pmb{T}}=\left\{x_1,x_2,\ldots,x_m\right\}$. Figure \ref{fig:problem} illustrates the workflow.

We assume that manual labeling is expensive especially when domain knowledge is required. To this end, we propose to tackle the ranking problem by only querying the predictions, which is highly applicable in practical scenarios. Remarkably, in this paper, we consider the performance of both accuracy and robustness. The accuracy is the correctness ratio of prediction on ID data. The robustness is the correctness ratio of OOD data.

\subsection{Motivating Example}
\label{subsec:motivating}
\begin{figure}[h]
    \centering
    \includegraphics[scale=0.5]{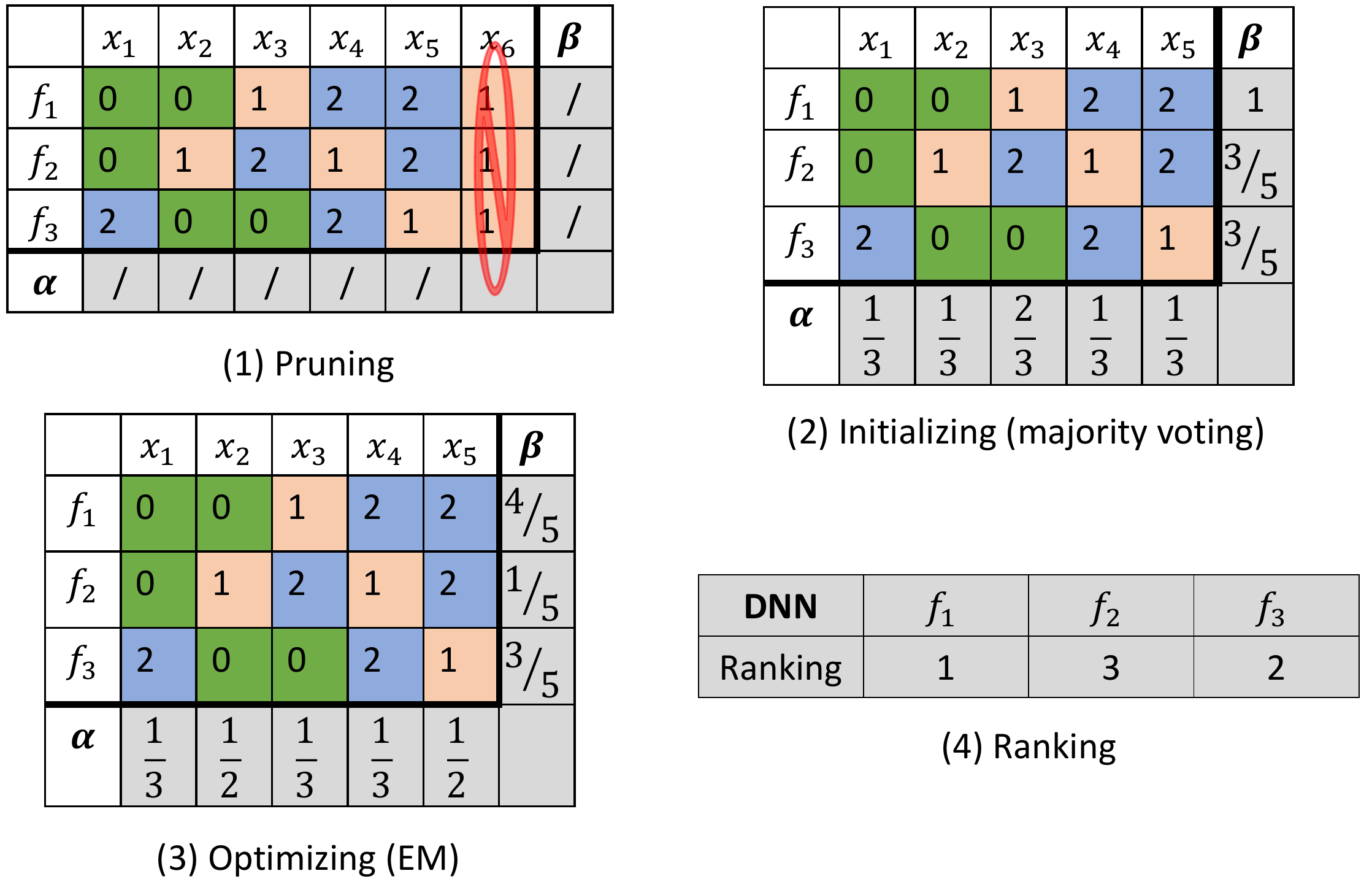}
    \caption{An example of our four-step approach. It ranks 3 DNNs ($f_1, f_2, f_3$) given 6 unlabeled data ($x_1, x_2, x_3, x_5, x_6$) in a 3-class classification task. Numbers (0, 1, 2) highlighted in colors are predicted labels.}
    \label{fig:example}
\end{figure}

Figure \ref{fig:example} gives an example of LaF. In this simple 3-class example, there are 3 DNN models ($f_1,f_2,f_3$) given 6 unlabeled samples ($x_1,x_2,\ldots,x_6$). The goal is to rank the 3 models concerning their accuracy on these samples in the absence of true labels. First, we compute the predicted label of each sample by each model and remove $x_6$ where all the models have the same prediction. Second, we initialize the two parameters, $\pmb{\alpha}$ and $\pmb{\beta}$, contained in our approach. $\pmb{\alpha}$ refers to how difficult a sample is for all models to predict the correct label. $\pmb{\beta}$ indicates how good a model is to output the correct labels of all samples. Initially, we use the simplest and most commonly used majority voting heuristic \cite{omer2018voting} to give a pseudo label to each sample. For instance, the pseudo label of $x_1$ is 0 because 2 ($f_1, f_2$) of 3 models predict the label as 0.  $\pmb{\alpha}$ is defined as the ratio of mismatched models that output a different label instead of the pseudo one. $\pmb{\beta}$ is calculated as the ratio of correctly predicted samples over the entire set. Third, since the pseudo labels are not the true labels, $\pmb{\alpha}$ and $\pmb{\beta}$ cannot truly reflect the data difficulty and model ability. We optimize these two parameters by a likelihood estimation method in the presence of true labels. Finally, based on the optimized $\pmb{\beta}$ ($\frac{4}{5}, \frac{1}{5}, \frac{3}{5}$), we obtain the ranking 1, 3, and 2 for $f_1, f_2$, and $f_3$, respectively. 

\subsection{Our Approach: LaF}
\label{subsec:our}
Given that no label is available in the test data, the main idea of our approach is to infer the specialties of DNNs by approximately maximizing the likelihood between the predictions and true labels via the expectation-maximization (EM) algorithm \cite{dempster1977em}. Let ${\widetilde{\pmb{Y}}}=\left\{\widetilde{y}_{ij}\right\}_{1\leq i\leq m, 1\leq j\leq n}$ be the predicted labels of ${\pmb{T}}$ and ${\pmb{Y}}=\left\{y_i\right\}_{1\leq i\leq m}$ be the true labels. Here, $\widetilde{y}_{ij}$ refers to $x_i$ and model $f_j$. Given the observed ${\widetilde{\pmb{Y}}}$ and latent ${\pmb{Y}}$ governed by unknown parameters ${\pmb{\theta}}$, the likelihood function is defined as $L\left(\pmb{\theta};\widetilde{\pmb{Y}}\right)=p\left(\widetilde{\pmb{Y}}\mid\pmb{\theta}\right)=\sum\limits_{i=1}^{m}p\left(\widetilde{\pmb{Y}},y_i\mid\pmb{\theta}\right)$. The goal is to search the best $\pmb{\theta}$ that maximizes the likelihood, in other words, the probability of observing ${\widetilde{\pmb{Y}}}$. As for $\pmb{\theta}$, inspired by \cite{Jacob2009glad}, we consider two factors, data difficulty $\pmb{\alpha}=\left\{\alpha_i\right\}_{1\leq i\leq m}$ and model specialty $\pmb{\beta}=\left\{\beta_j\right\}_{1\leq j\leq m}$, that influence the performance of DNNs. Namely, $\pmb{\theta}=\left(\pmb{\alpha},\pmb{\beta}\right)$. Algorithm \ref{alg:our} presents the pseudo-code of our approach. 

\textbf{Step 1: Pruning.} Inevitably, some data will receive the same predictions by all models, which is useless for discriminating the performance and causes computational cost. For this reason, we filter these data without losing any information for ranking and obtain a smaller set $\pmb{T'}$ (Lines 1-6 in Algorithm \ref{alg:our}).

\textbf{Step 2: Initializing.} First, for each data $x_i$, a pseudo label is voted by a majority of DNNs, namely, $y_i'={\rm mode}\left(\left\{\widetilde{y}_{ij}\right\}_{1\leq j\leq n}\right)$ (Lines 7-9). Next, $\alpha_i$ is the number of DNNs that gives a different label from the pseudo label and $\beta_j$ is the accuracy based on pseudo labels (Lines 10-15). Formally, the definitions are:
\begin{equation}
\label{equ:alpha}
    \alpha_i=\frac{\left\{\widetilde{y}_{ij}\mid\widetilde{y}_{ij}\neq y_i', 1\leq j\leq n\right\}}{n},
    \beta_j=\frac{\left\{\widetilde{y}_{ij}\mid\widetilde{y}_{ij}=y_i',1\leq i\leq \left|{\pmb{T'}}\right|\right\}}{\left|{\pmb{T'}}\right|}
\end{equation}

\textbf{Step 3: Optimizing.} The EM algorithm solves the optimization problem by iteratively performing an expectation (E) step and a maximization (M) step (Lines 16-26). In the E-step, it estimates the expected value of the log-likelihood:
\begin{equation}
\label{equ:estep}
\begin{split}
Q(\pmb\theta,\pmb{\theta_{last}})&={\rm E}\left[\log L\left(\pmb\theta;\widetilde{\pmb Y}, \pmb Y\right)\right]\\
&=\sum\limits_{i=1}^{\left|{\pmb{T'}}\right|}{\rm E}\left[\log\left(p\left(y_i\right)\right)\right]+\sum\limits_{i=1}^{\left|{\pmb{T'}}\right|}\sum\limits_{j=1}^{n}{\rm E}\left[p\left(\widetilde{y}_{ij}\mid y_i,\alpha_i,\beta_j\right)\right]
\end{split}
\end{equation}
where $\pmb\theta=\left(\pmb\alpha,\pmb\beta\right)$ and $\pmb{\theta_{last}}$ is from the last E-step. For the computation, we use the definition from \cite{Jacob2009glad} where $p\left(\widetilde{y}_{ij}=y_j\mid\alpha_i,\beta_j\right)=\frac{1}{1+e^{-\alpha_i\beta_j}}$. Besides, as $\widetilde{y}_{ij}$ and $\pmb\beta$ are independent given $\pmb\alpha$, $p\left(y_i\right)=p\left(y_i\mid\pmb\alpha,\beta_j\right)$. Remarkably, $y_i$ represents the true label of a sample. In the ranking problem, $y_i$ is absent but the probability of taking it as a true label is can be inferred by $p\left(y_i\mid\pmb\alpha,\beta_j\right)$.

In the M-step, the gradient ascent is applied to search for $\pmb\alpha$ and $\pmb\beta$ that maximize $Q$:
\begin{equation}
\label{equ:mstep}
\pmb{\theta_{new}}=\argmax_{\pmb{\theta}}Q(\pmb\theta,\pmb{\theta_{last}})
\end{equation}
where $\pmb{\theta_{new}}$ is the updated parameters for the next iteration.

\textbf{Step 4: Ranking.} Finally, as $\pmb\beta$ well estimate the abilities of each DNN given the observed labels, we use this vector to rank DNNs (Line 27). A high specialty indicates a good performance on the data. 

\begin{algorithm}[htpb]
\small
\caption{LaF: Labeling-free comparison testing}
\label{alg:our}
\SetAlgoLined
\Input{$\left\{f_1,f_2,\ldots,f_n\right\}$: DNNs for comparison\\
${\pmb{T}}=\left\{x_1,x_2,\ldots,x_m\right\}$ : test set\\
$\Gamma$: performance criterion}
\Output{$\left\{r'\left(f_1\right), r'\left(f_2\right), \ldots, r'\left(f_n\right)\right\}$: Rank of DNNs}

\tcc{Step1: Pruning}
 ${\pmb{T'}}=\left\{\right\}$
  
  \For{$i=1 \to m$}
  {
    \If{$\left|\left\{\left\{\widetilde{y}_{ij}\right\}_{1\leq j\leq n}\right\}\right|>1$} 
    {
        ${\pmb{T'}}\leftarrow x_i$ \tcp*{$\widetilde{y}_{ij}$ is the predicted label by $f_j$}
    }
  }
  \tcc{Step 2: Initializing}
  \For{$i=1 \to \left|{\pmb{T'}}\right|$}
  { 
    $y_i'={\rm mode}\left(\left\{\widetilde{y}_{ij}\right\}_{1\leq j\leq n}\right)$\tcp*{Majority voting}
  }
  \For{$i=1 \to \left|{\pmb{T'}}\right|$}
  {
      $\alpha_i=\frac{\left\{\widetilde{y}_{ij}\mid\widetilde{y}_{ij}\neq y_i', 1\leq j\leq n\right\}}{n}$ \tcp*{Data difficulty}
    }
    \For{$j=1 \to n$}
  {
      $\beta_j=\frac{\left\{\widetilde{y}_{ij}\mid\widetilde{y}_{ij}=y_i',1\leq i\leq \left|{\pmb{T'}}\right|\right\}}{\left|{\pmb{T'}}\right|}$ \tcp*{Model specialty}
    }
 
  \tcc{Step 3: Optimizing}
  ${\pmb\alpha_{last}}=\left\{\alpha_i\right\}_{1\leq i\leq \left|{\pmb{T'}}\right|}$
  
  ${\pmb\beta_{last}}=\left\{\beta_j\right\}_{1\leq j\leq n}$
  
  $Q_{last}=computeQ({\pmb\alpha_{last}}, {\pmb\beta_{last}})$ \tcp*{$ComputeQ$ estimates the log likelihood based on Equation (\ref{equ:estep})}
  
  $\pmb\alpha, \pmb\beta=gradientAscent\left({\pmb\alpha_{last}}, {\pmb\beta_{last}}, Q_{last}\right)$
  
  $Q=computeQ(\pmb\alpha, \pmb\beta)$
  
  \While{$\left|\frac{Q - Q_{last}}{Q_{last}}\right| > 1E-5$}
  {
    $Q_{last}=Q$
    
    ${\pmb\alpha_{last}}, {\pmb\beta_{last}}=\pmb\alpha, \pmb\beta$
    
    $\pmb\alpha, \pmb\beta=gradientAscent\left({\pmb\alpha_{last}}, {\pmb\beta_{last}}, Q_{last}\right)$
  
  $Q=computeQ(\pmb\alpha, \pmb\beta)$
  }
  \tcc{Step 4: Ranking}
  $r'\left(f_1\right), r'\left(f_2\right), \ldots, r'\left(f_n\right)=Sort({\pmb\beta})$ 
  
  \Return
  $\left\{r'\left(f_1\right), r'\left(f_2\right), \ldots, r'\left(f_n\right)\right\}$
\end{algorithm}

\section{Experimental Setup}
\label{sec:design}

\subsection{Implementation}
\label{subsec:hardware}
All experiments were conducted on a high-performance computer cluster and each cluster node runs a 2.6 GHz Intel Xeon Gold 6132 CPU with an NVIDIA Tesla V100 16G SXM2 GPU. We implement the proposed approach and baseline methods based on the state-of-the-art frameworks, Tensorflow 2.3.0 and PyTorch 1.6.0. For artificial data distribution shift, we consider 2 benchmark datasets where each includes 15 types of natural corruption with 5 levels of severity. In total, we test on $2*15*5=150$ datasets with the artificial distribution shift. Due to the space limitation, we only report the average results on corrupted data for baseline methods. The remaining results corroborate our findings and are available on our companion project website \cite{homepage}.

\subsection{Research Questions}
\label{subsec:rqs}
In this study, we focus on the following three research questions:
\begin{itemize}
    \item RQ1 (\textbf{effectiveness given ID test}): How is LaF ranking multiple DNNs given ID test data?
    \item RQ2 (\textbf{effectiveness under distribution shift}): How is LaF ranking multiple DNNs given OOD test data (including artificial and natural distribution shifts)?
    \item RQ3 (\textbf{impact factors of LaF}): What is the impact of model quality and diversity on ranking DNNs?
\end{itemize}

The first two research questions successively evaluate the effectiveness of our proposed solution given test data with and without distribution shift. The second one also tends to show how flexible and practical LaF is in real-world applications, especially by the test data with natural distribution shifts. The last one investigates the impact factors that may affect the ranking performance.


\subsection{Datasets and DNNs}
\label{subsec:data}
\textbf{Datasets.} We choose 7 datasets, MNIST \cite{Lecun1998gradient}, Fashion-MNIST \cite{fashionMnist}, CIFAR-10 \cite{Alex2009techm}, iWildCam \cite{beery2020iwildcam}, Amazon \cite{ni2019justifying}, Java250, and C++1000 \cite{codenet2021} that are widely studied in previous work. These datasets cover the image (first 4), text (Amazon), and source code (Java250 and C++1000) domains. The test data that follow the same distribution as the training set are the so-called in-distribution (ID) data. The test data with data distribution shift are out-of-distribution (OOD). In our work, we consider two types of distribution shifts: artificial and natural. For the artificial distribution shift, we use two benchmark datasets, MNIST-C \cite{mnistc2019norman} and CIFAR-10-C \cite{cifar10c2019dan}, for MNIST and CIFAR-10, respectively. Each benchmark includes 75 datasets with 15 types of natural corruptions, such as Gaussian noise, shot noise, impulse noise, defocus blur, frosted glass blur, motion blur, zoom blur, snow, frost, fog, brightness, contrast, elastic, pixelate, and jpeg. Besides, each type of corruption has 5 levels of severity. For the natural distribution shift, we use two datasets for iWildCam and Amazon, respectively, from a recent-published benchmark, WILDS \cite{wilds2021}. The distribution shift comes from new came traps in iWildCam and new users in Amazon. For Java250, we manually collect the OOD dataset based on the definition in WILDS that the distribution shift of source code comes from new repositories. For each class in Java250, we extract java files from \cite{codenet2021} under the constraint that the corresponding users do not exist in ID data. Table \ref{tab:data-summary} lists the details of datasets.

\begin{table*}[htpb]
\caption{Summary of datasets. ``\#ID'' is the number of in-distribution test data. ``\#OOD'' is the number of out-of-distribution test data with artificial or natural distribution shifts.}
\label{tab:data-summary}
\resizebox{\textwidth}{!}{
\begin{tabular}{lrrrrrr}
\hline
\textbf{Dataset} & \textbf{Domain}& \textbf{Data Type} & \textbf{\#Classes} & \textbf{\#ID} & \textbf{\#OOD} & \textbf{Distribution Shift}\\ \hline
MNIST & Computer vision & Image of handwritten digits & 10 & 10,000 & 750,000 & Artificial \\
Fashion-MNIST & Computer vision & Image of fashion products & 10 & 10,000 & - & - \\
CIFAR-10 & Computer vision & Image of animals and vehicles & 10 & 10,000 & 750,000 & Artificial \\
iWildCam & Computer vision & Image of wildlife & 182 & 8,154 & 42,791 & Natural \\
Amazon & Natural language processing & Text of comments & 5 & 46,950 & 100,050 & Natural \\
Java250 & Source code analysis & Source in Java & 250 & 15,000 & 15,000 & Natural \\ 
C++1000 & Source code analysis & Source code in C++ & 1,000 & 99.997 & - & - \\ \hline
\end{tabular}
}
\end{table*}

\textbf{DNNs.} From Github, we collect, in total, 165 models, 30 for MNIST, 25 for Fashion-MNIST, 30 for CIFAR-10, 20 for iWildCam, 20 for Amazon, 20 for Java250, and 20 for C++1000. In concrete, the models of MNIST and CIFAR-10 are extracted using the GitHub links in \cite{sds2021meng}. The 25 models of Fashion-MNIST are extracted from \cite{fashion2021model} \cite{sds2021meng}. For iWildCam and Amazon, we train models using the implementation in the benchmark WILDS \cite{wilds2021}. For Java250 and C++1000, we train models using the implementation in the benchmark Project CodeNet \cite{codenet2021} by different optimizers (sgd, rmsprop, adam, adadelta, adagrad, adamax, nadam, ftrl) and architectures (basic, doublePool). Table \ref{tab:model-summary} presents the accuracy and robustness of DNNs on ID test data and OOD test data with natural distribution shift, respectively.

\begin{table*}[htpb]
\caption{Summary of models. ``\#DNNs'' is the number of DNNs collected for each dataset. ``\#Parameters'' shows the minimum and the maximum number of parameters of collected DNNs. ``Accuracy'' and ``Robustness'' lists the lowest and highest accuracy and robustness on test data with and without distribution shift, respectively. MNIST-C and CIFAR-10-C are two benchmark datasets and corresponding robustness is summarized in Table \ref{tab:robustness}.}
\label{tab:model-summary}
\resizebox{.7\textwidth}{!}{
\begin{tabular}{lrrrr}
\hline
\textbf{Dataset} & \textbf{\#DNNs} & \textbf{\#Parameters} & \textbf{Accuracy (\%)} & \textbf{Robustness (\%)} \\ \hline
MNIST & 30 & 7,206-3,274,634 & 85.27-99.54 & MNIST-C \\
Fashion-MNIST & 25 & 258,826-1,256,080 & 90.09-93.38 & - \\
CIFAR-10 & 30 & 62,006-45,294,194 & 69.90-95.92 & CIFAR-10-C \\
iWildCam & 20 & 7,224,054-23,960,630 & 75.72-77.26 & 65.30-76.82 \\
Amazon & 20 & 7,037,504-23,587,712 & 73.60-74.84 & 71.35-72.35 \\
Java250 & 20 & 2,927,290 & 64.73-87.39 & 57.24-81.67 \\
C++1000 & 20 & 4,409,288 & 71.39-92.10 & - \\ \hline
\end{tabular}
}
\end{table*}

\begin{table}[htpb]
\caption{Summary of MNIST-C and CIFAR-10-C with the artificial distribution shift and robustness. Each dataset includes 15 types of natural corruptions (e.g., Gaussian Noise) with 5 levels of severity (1-5). The number in each cell presents the minimum and maximum robustness of multiple DNNs given the corruption type and severity.}
\label{tab:robustness}
\resizebox{.8\columnwidth}{!}{
\begin{tabular}{lccccc}
\hline
\textbf{Corruption Type} & \textbf{Severity=1} & \textbf{Severity=2} & \textbf{Severity=3} & \textbf{Severity=4} & \textbf{Severity=5} \\ \hline
\multicolumn{6}{c}{\textbf{MNIST-C}} \\ \hline
Gaussian Noise & 84.85-99.49 & 80.49-99.25 & 55.89-99.05 & 29.86-98.77 & 16.92-96.02 \\
Shot Noise & 84.89-99.53 & 84.73-99.49 & 84.87-99.38 & 84.32-99.00 & 83.12-98.68 \\
Impulse Noise & 84.29-99.20 & 71.33-98.91 & 56.50-98.62 & 27.77-96.39 & 16.80-88.67 \\
Defocus Blur & 58.63-95.69 & 31.76-84.91 & 9.73-43.40 & 3.73-20.46 & 1.96-18.30 \\
Frosted Glass Blur & 65.06-96.52 & 54.02-94.33 & 19.06-75.43 & 15.63-70.29 & 11.12-54.46 \\
Motion Blur & 56.71-97.29 & 36.28-90.04 & 25.23-78.19 & 20.29-65.58 & 18.59-61.07 \\
Zoom Blur & 82.68-99.54 & 81.50-99.45 & 80.42-99.35 & 78.30-99.22 & 74.93-98.7 \\
Snow & 51.43-99.06 & 17.78-98.38 & 19.27-96.03 & 16.53-93.73 & 11.39-95.73 \\
Frost & 17.25-98.99 & 10.60-97.91 & 9.52-96.72 & 9.18-96.63 & 9.05-95.36 \\
Fog & 9.74-97.07 & 9.61-95.32 & 9.75-89.51 & 9.73-85.28 & 9.69-73.37 \\
Brightness & 66.18-99.53 & 21.94-99.22 & 12.55-98.99 & 9.96-98.60 & 9.10-97.06 \\
Contrast & 37.47-99.22 & 20.38-99.17 & 10.29-99.04 & 9.74-98.08 & 8.02-92.11 \\
Elastic & 49.09-99.07 & 12.80-17.58 & 79.89-97.75 & 44.52-94.33 & 12.32-71.43 \\
Pixelate & 80.40-98.64 & 84.37-99.11 & 77.65-98.47 & 62.11-92.34 & 54.96-89.33 \\
JPEG & 84.90-99.53 & 84.93-99.56 & 85.10-99.54 & 84.78-99.44 & 84.85-99.36 \\ \hline
\multicolumn{6}{c}{\textbf{CIFAR-10-C}} \\ \hline
Gaussian Noise & \multicolumn{1}{l}{59.02-82.46} & \multicolumn{1}{l}{45.70-75.57} & \multicolumn{1}{l}{34.70-73.70} & \multicolumn{1}{l}{30.07-72.54} & \multicolumn{1}{l}{25.08-71.14} \\
Shot Noise & 66.72-88.90 & 57.29-81.71 & 42.12-75.33 & 37.34-74.12 & 31.77-71.97 \\
Impulse Noise & 63.82-88.70 & 52.58-83.19 & 42.74-76.43 & 24.28-66.53 & 15.32-61.54 \\
Defocus Blur & 69.62-96.13 & 68.32-96.19 & 66.24-95.85 & 52.11-91.50 & 31.98-86.14 \\
Frosted Glass Blur & 35.53-74.17 & 37.23-73.88 & 40.17-73.61 & 30.33-68.81 & 32.24-68.36 \\
Motion Blur & 67.30-94.31 & 59.71-91.27 & 49.23-86.62 & 48.20-86.27 & 40.05-81.53 \\
Zoom Blur & 64.45-93.94 & 57.73-94.29 & 49.07-92.93 & 42.70-91.18 & 35.24-87.35 \\
Snow & 67.65-93.63 & 54.93-87.77 & 59.97-88.74 & 58.91-86.40 & 54.09-82.56 \\
Frost & 66.73-92.93 & 62.62-89.90 & 54.64-84.75 & 53.15-83.61 & 44.76-76.56 \\
Fog & 68.71-95.87 & 61.93-95.16 & 55.08-93.99 & 45.97-91.18 & 32.22-75.58 \\
Brightness & 69.55-95.76 & 68.88-95.65 & 67.51-94.98 & 65.73-94.44 & 60.59-92.64 \\
Contrast & 67.12-95.77 & 49.64-93.63 & 38.69-91.56 & 30.37-87.35 & 15.53-64.79 \\
Elastic & 62.71-93.92 & 63.91-94.03 & 63.78-93.46 & 55.80-86.46 & 53.64-80.07 \\
Pixelate & 69.62-94.77 & 67.38-91.97 & 60.46-89.24 & 42.55-78.97 & 28.45-75.55 \\
JPEG & 69.29-87.71 & 64.43-83.60 & 61.23-81.53 & 59.08-80.11 & 54.93-77.30 \\ \hline
\end{tabular}
}
\end{table}

\subsection{Baseline Methods}
\label{subsec:baselines}

In our study, we compare LaF to 3 baseline methods, random sampling, SDS, and CES.  All baseline methods are sample-selection-based. Following \cite{sds2021meng}, the labeling budget of the baseline methods ranges from the number of DNNs (i.e., 30 DNNs in MNIST) to 180 at intervals of 5. 

\textbf{Random sampling} is a basic and model-independent method for data selection where each data has an equal probability to be considered. A subset of data is randomly selected and annotated to rank DNNs.
    
\textbf{Sample discrimination based selection (SDS)} \cite{sds2021meng} is the state-of-the-art approach in ranking multiple DNNs with respect to accuracy. Following \cite{sds2021meng}, among data in the top 25\% with high discrimination scores, we randomly select a given budget of data to label and annotate to perform the ranking task.
    
\textbf{Cross Entropy-based Sampling (CES)} \cite{ces2019li} is designed to select a set of representative data to approximate the actual performance given a single DNN. We follow the same procedure as \cite{sds2021meng} to adapt CES for multi-DNN comparison. 

Due to the random manner in the sampling methodology, each experiment of the baseline methods is repeated 50 times and we report the average result.

\subsection{Evaluation Measures}
\label{subsec:eval}
To evaluate the effectiveness of each method, we follow the baseline work \cite{sds2021meng} and apply the statistical analysis, Spearman's rank-order correlation \cite{daniel1990applied}, and Jaccard similarity \cite{sds2021meng}. The first one evaluates the general ranking of all models, while the last one specifically estimates the ranking of top-$k$ DNNs. In addition, we add the evaluation on Kendall's $\tau$ rank correlation \cite{daniel1990applied}. Similar to Spearman's rank-order correlation, Kendall's $\tau$ measures the non-parametric rank correlation. However, Kendall's $\tau$ calculates based on concordant and discordant pairs and is insensitive to errors (if any) in data. By contrast, Spearman's rank-order correlation calculates based on deviations and is more sensitive to errors (if any) in data.

Given $n$ DNNs, $f_1, f_2, \ldots, f_n$, let $r\left(f_1\right), r\left(f_2\right), \ldots, r\left(f_n\right)$ be the ground truth ranking and \\ $r'\left(f_1\right), r'\left(f_2\right), \ldots, r'\left(f_n\right)$ be the estimated ranking. The Spearman's rank-order correlation coefficient  is computed as
\begin{equation}
\label{equ:spearman}
    \rho=\frac{n\sum\limits_{i=1}^nr\left(f_i\right)r'\left(f_i\right)-\left(\sum\limits_{i=1}^nr\left(f_i\right)\right)\left(\sum\limits_{i=1}^nr'\left(f_i\right)\right)}{\sqrt{\left[n\sum\limits_{i=1}^nr\left(f_i\right)^2-\left(\sum\limits_{i=1}^nr\left(f_i\right)\right)^2\right]\left[n\sum\limits_{i=1}^nr'\left(f_i\right)^2-\left(\sum\limits_{i=1}^nr'\left(f_i\right)\right)^2\right]}}
\end{equation}
A large $\rho$ indicates that the correlation between the ground truth and estimation is strong.

Kendall's $\tau$ is 
\begin{equation}
\label{equ:tau}
\tau=\frac{P-Q}{\sqrt{\left(P+Q+T\right)\left(P+Q+U\right)}}
\end{equation}
where $P$ and $Q$ are the numbers of ordered and disordered pairs in $\left\{r\left(f_i\right), r'\left(f_i\right)\right\}$, respectively. $T$ and $U$ are the numbers of ties in $\left\{r\left(f_i\right)\right\}$ and $\left\{r'\left(f_i\right)\right\}$, respectively. A large $\tau$ indicates a strong agreement between the ground truth and estimation.

Meng \emph{et al.} proposed to apply the Jaccard similarity for measuring the similarity between the top-$k$ models. The similarity coefficient is defined as:
\begin{equation}
\label{equ:jac}
    J_k=\frac{\mid\left\{f_i\mid r\left(f_i\right)<=k\right\}\cap\left\{f_i\mid r'\left(f_i\right)<=k\right\}\mid}{\mid\left\{f_i\mid r\left(f_i\right)<=k\right\}\cup\left\{f_i\mid r'\left(f_i\right)<=k\right\}\mid},1\leq i\leq n
\end{equation}
A large $J_k$ implies a high success in identifying the top-$k$ models.

\section{Results and Discussion}
\label{sec:results}

\subsection{RQ1: Effectiveness Given ID Test Data}
\label{subsec:accuracy}
\begin{figure*}
    \centering
    \subfigure[MNIST]{
    \includegraphics[scale=0.29]{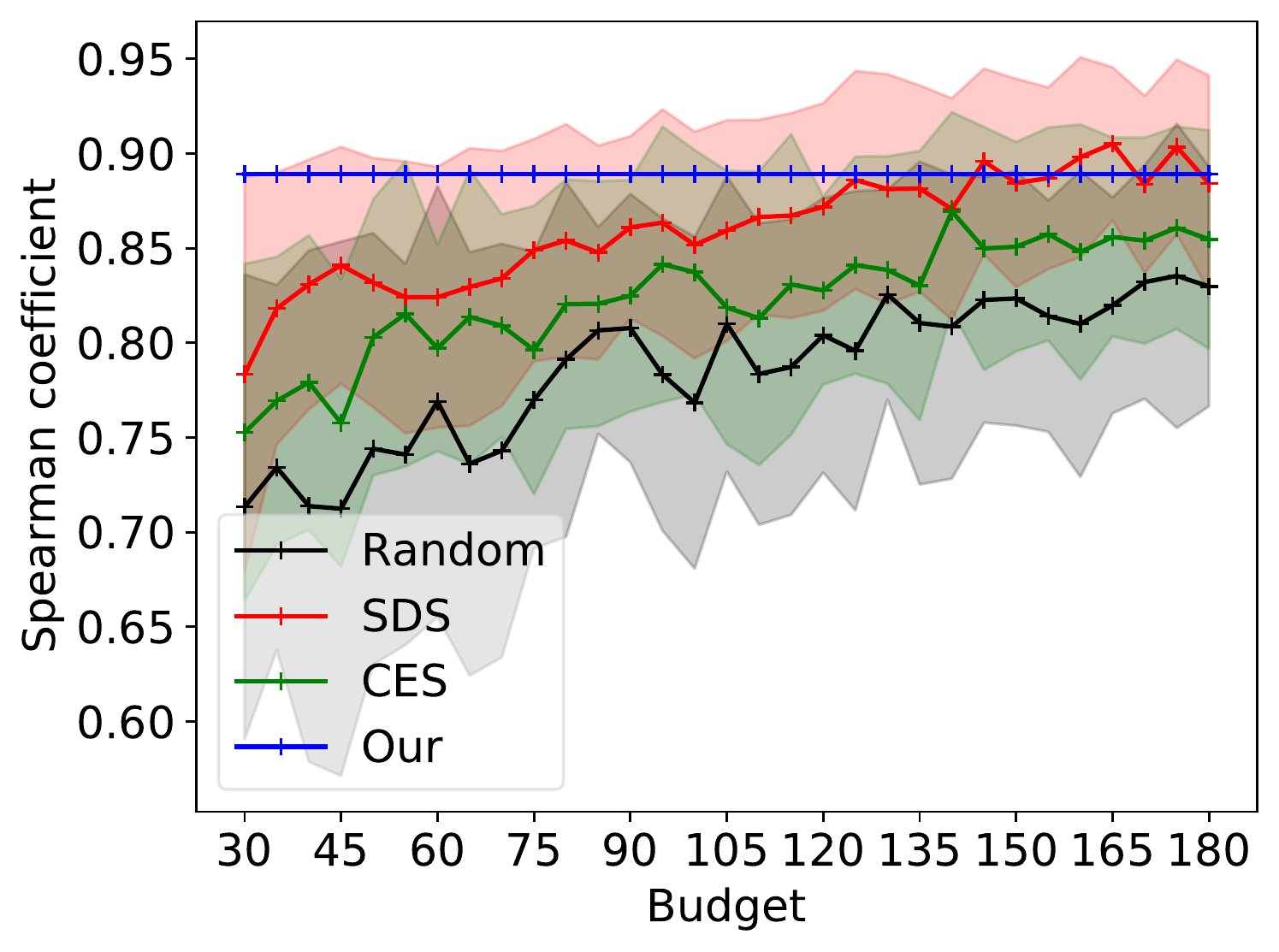}
    }
    \subfigure[Fashion-MNIST]{
    \includegraphics[scale=0.29]{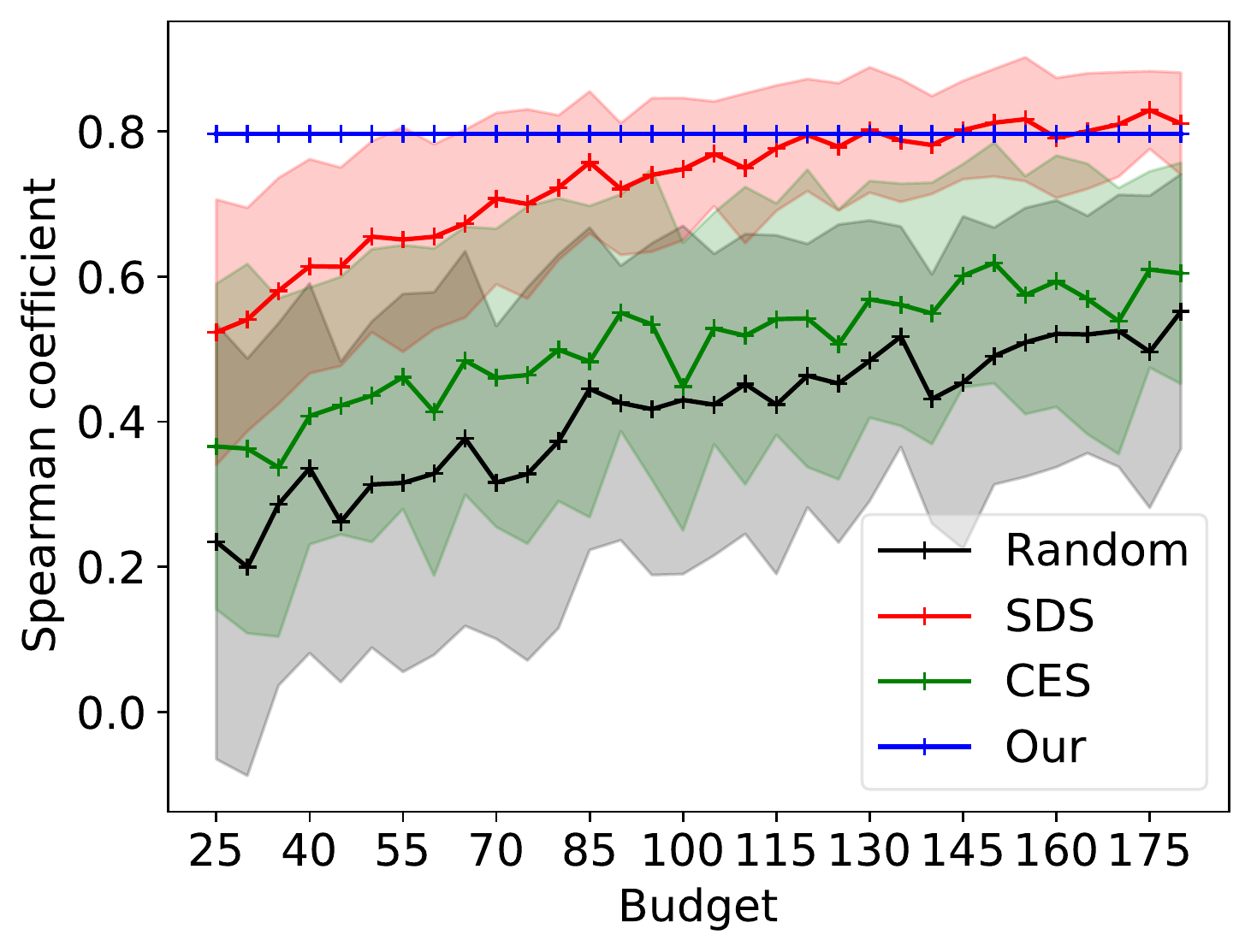}
    }
    \subfigure[CIFAR-10]{
    \includegraphics[scale=0.29]{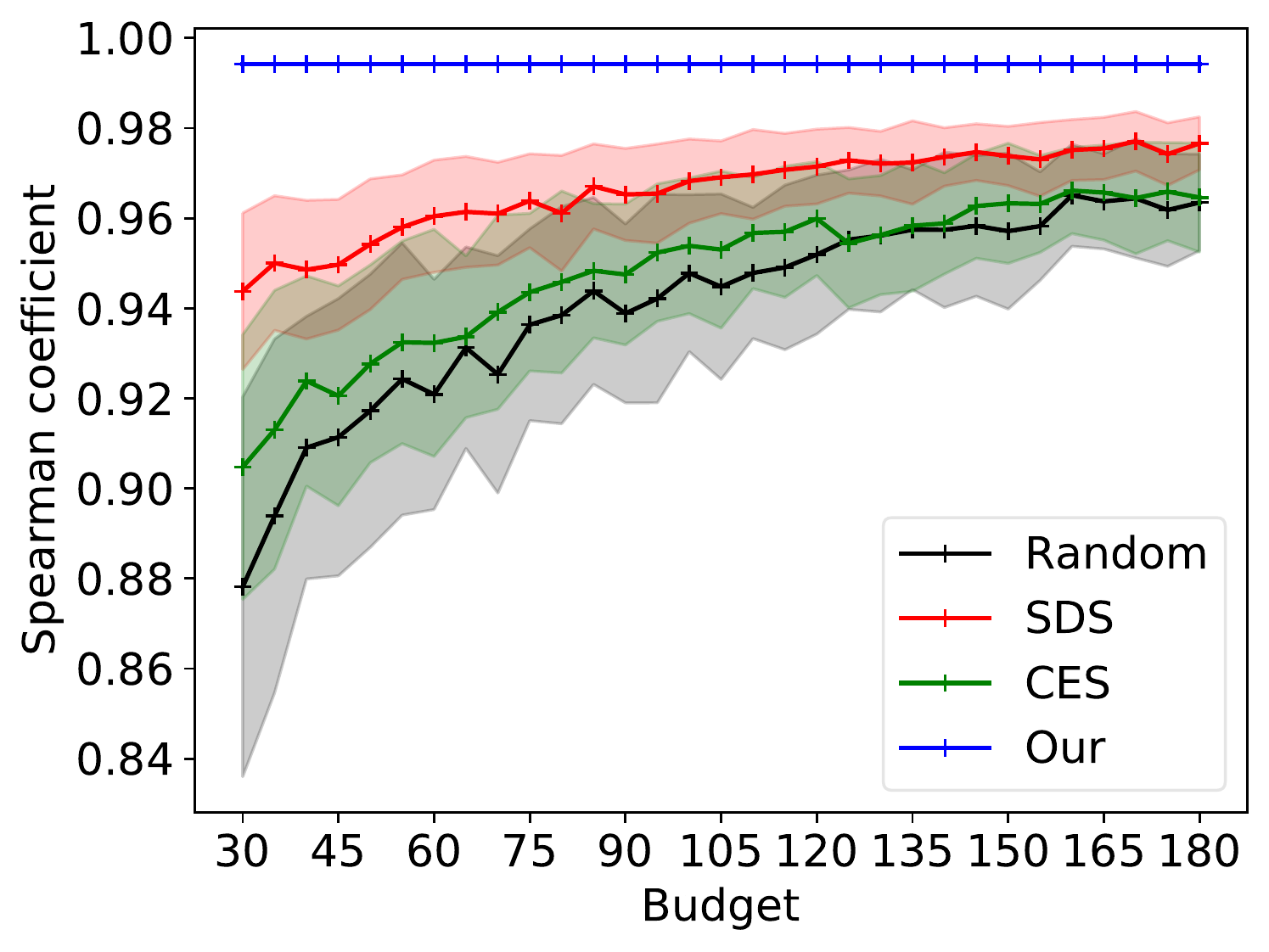}
    }
    \subfigure[iWildCam-ID]{
    \includegraphics[scale=0.29]{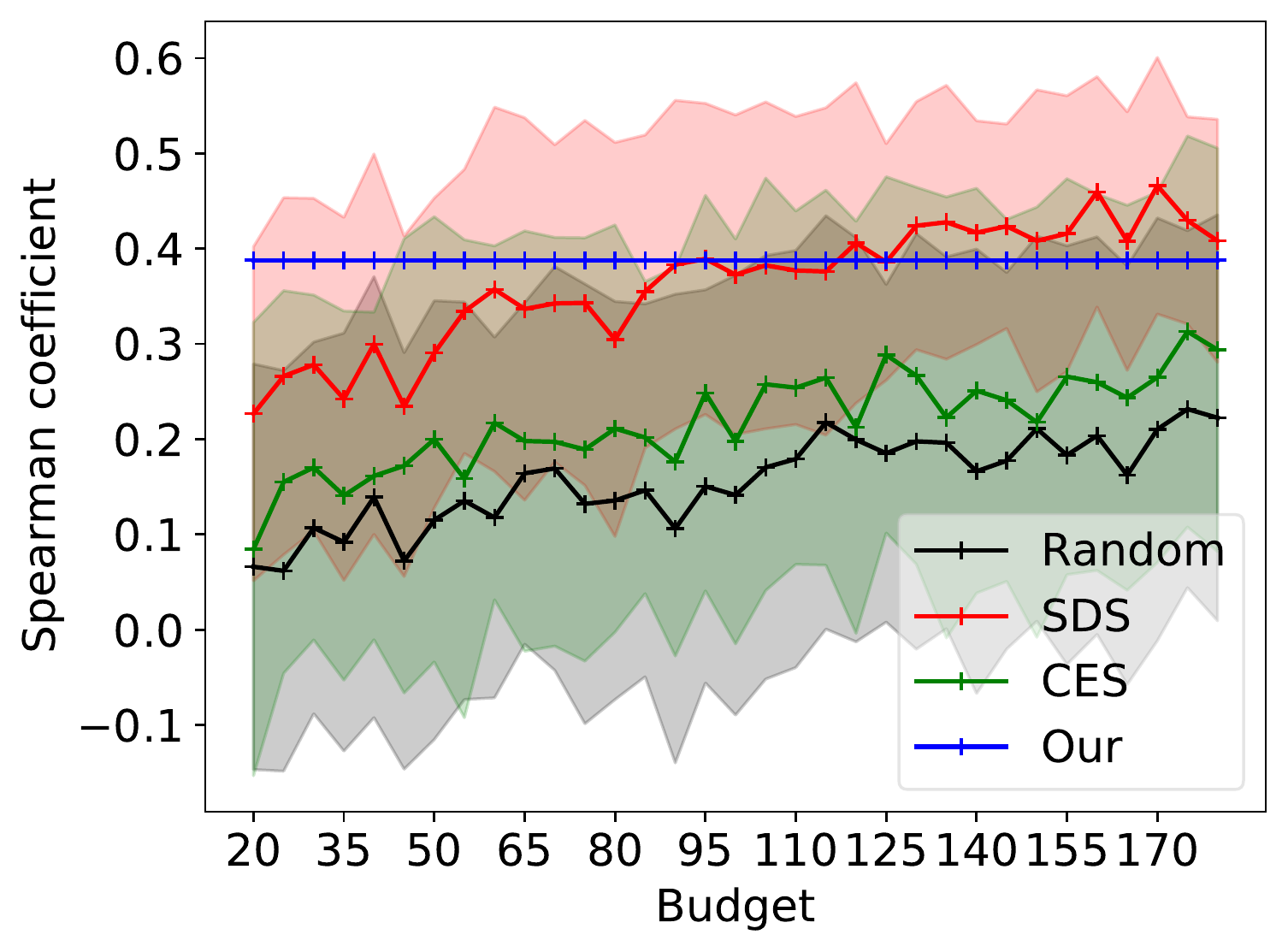}
    \label{fig:clean-camera}
    }
    \subfigure[Amazon-ID]{
    \includegraphics[scale=0.29]{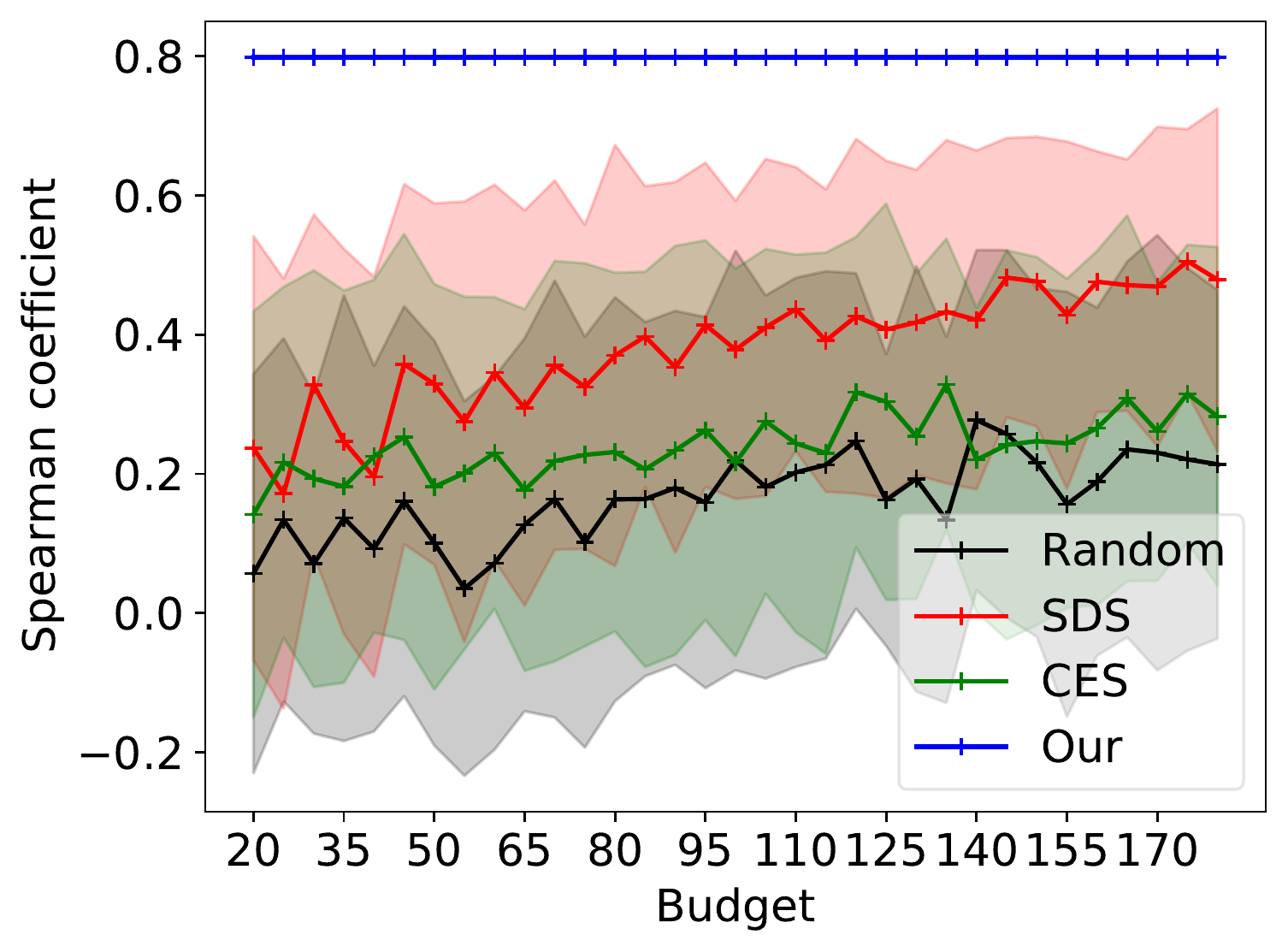}
    }
    \subfigure[Java250]{
    \includegraphics[scale=0.29]{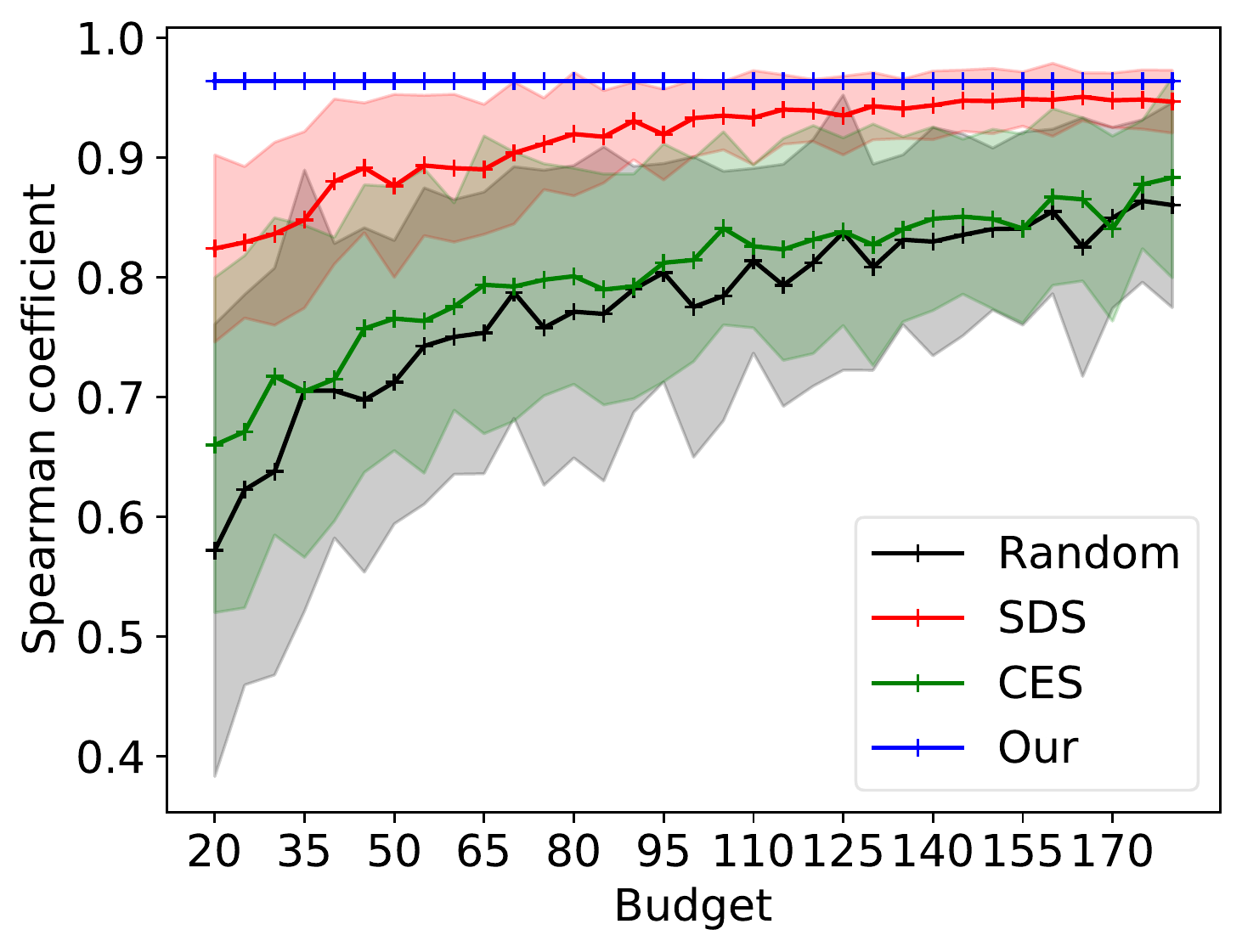}
    }
    \subfigure[C++1000]{
    \includegraphics[scale=0.29]{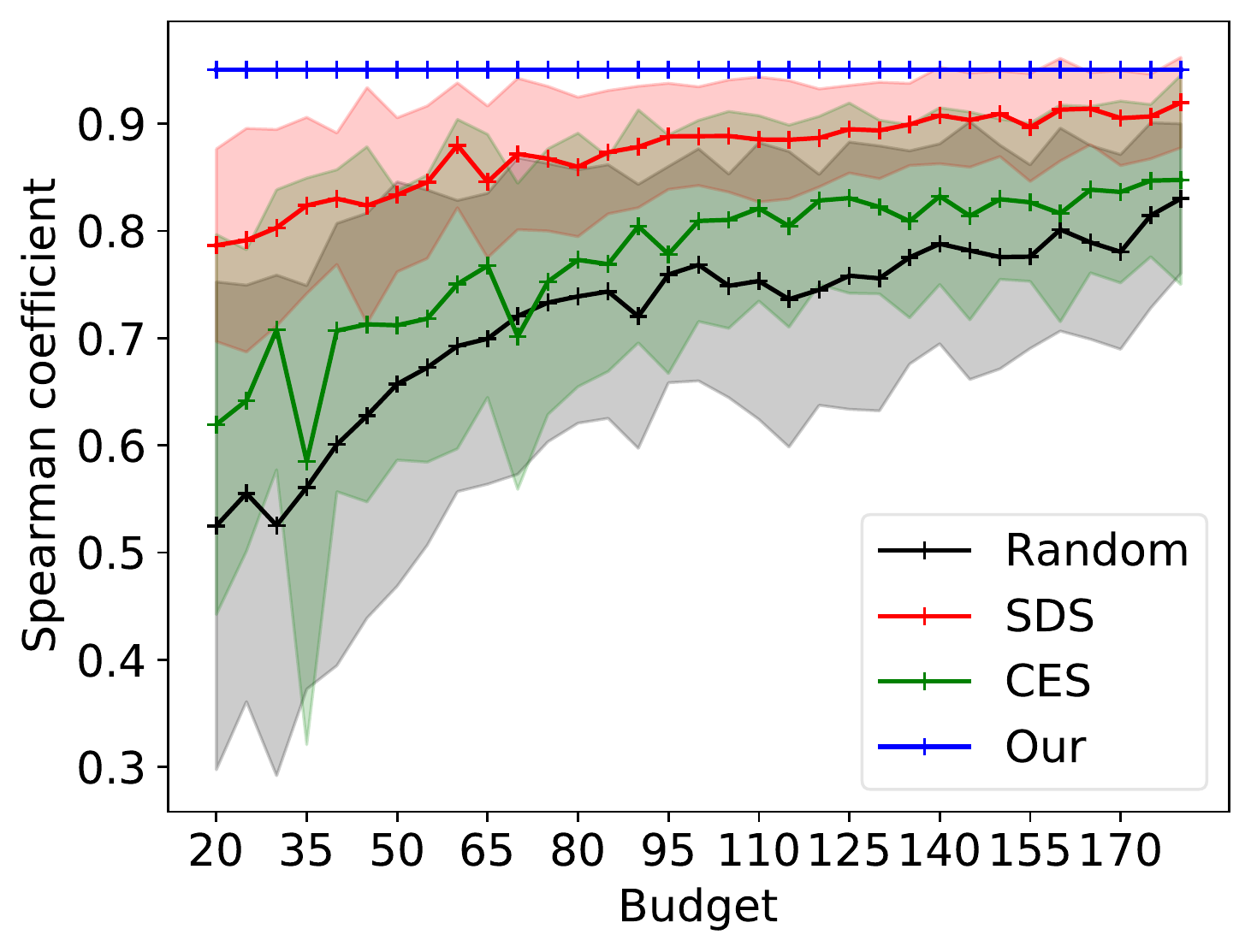}
    }
    \caption{Spearman's correlation coefficient of ranking results based on ID test data. The higher the better. The shaded area represents the standard deviation. ``Budget'' is the number of labeled data.}
    \label{fig:clean-spearman}
\end{figure*}

First, we compare the effectiveness of four methods in ranking multiple DNNs based on the accuracy of ID data. Figure \ref{fig:clean-spearman} shows the result measured by Spearman's rank-order correlation. The first conclusion we can draw is that, over seven datasets, all methods succeed in outputting positively correlated rankings. By comparison, LaF continuously outperforms (by up to 0.74) the baseline methods regardless of the labeling budget. Namely, the ranking by LaF is strongly correlated with the ground truth. In general, for the three sample-selection-based baseline methods, the correlation between the estimated rank and the ground truth increases when more data are labeled. However, for some datasets, the performance is still far from LaF. For example, in Amazon, LaF obtains a correlation coefficient of 0.80, while the best baseline, SDS, only achieves 0.48 using the maximum labeling budget of 180. Besides, due to the sampling randomness, each baseline method obtains different ranking results over 50 experiments, which is indicated by the large standard deviation (up to 0.36, shaded area in the figure) at each labeling budget. As a result, the rank by one experiment is not reliable by occasionally being good and poor. In particular, the standard deviation becomes smaller when more data are labeled, which means the ranking method highly relies on the labeling budget. By contrast, since LaF is labeling-free, there is no sampling randomness, in other words, the rank is deterministic.

\begin{figure*}
    \centering
    \subfigure[MNIST]{
    \includegraphics[scale=0.285]{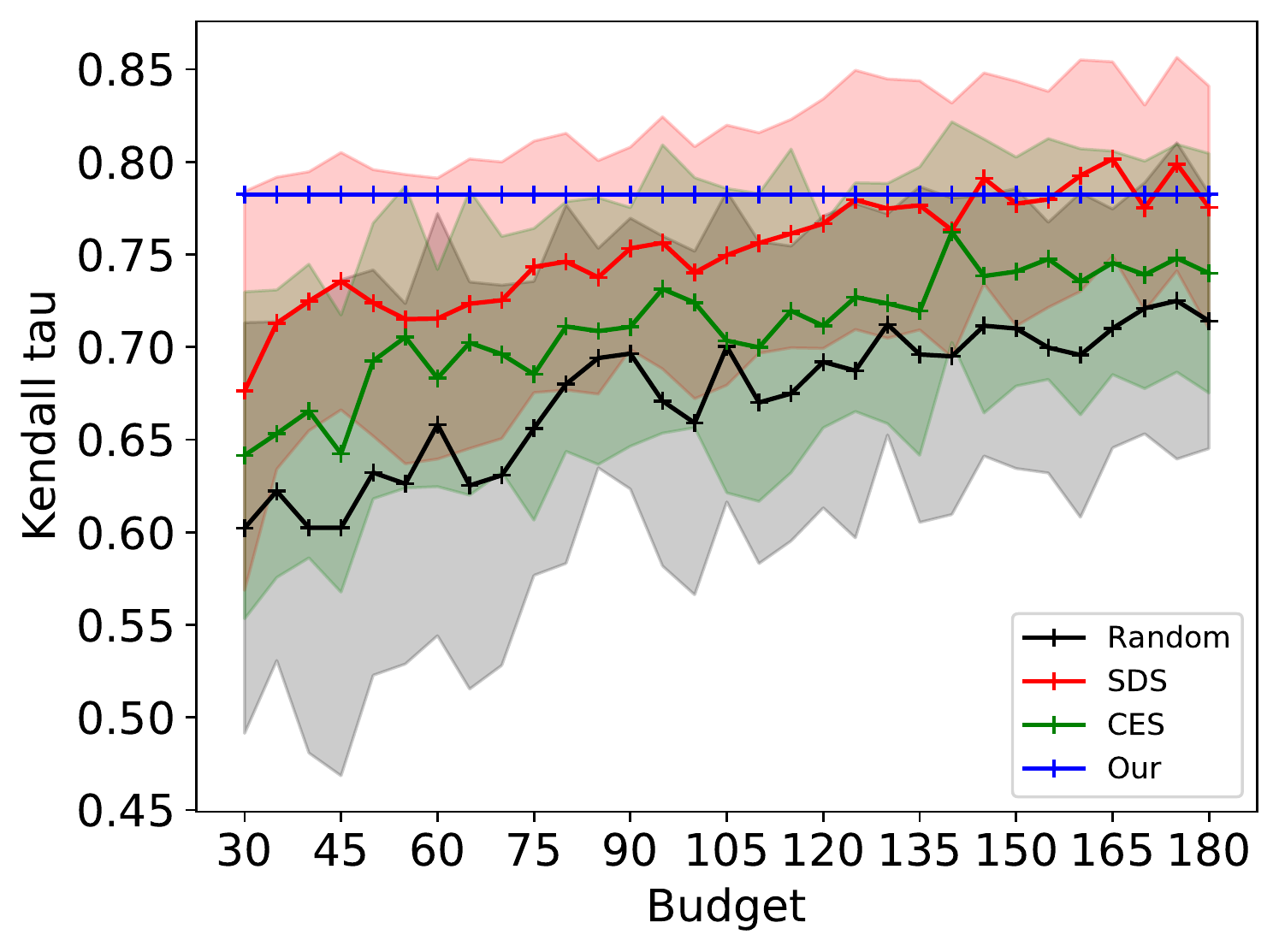}
    }
    \subfigure[Fashion-MNIST]{
    \includegraphics[scale=0.29]{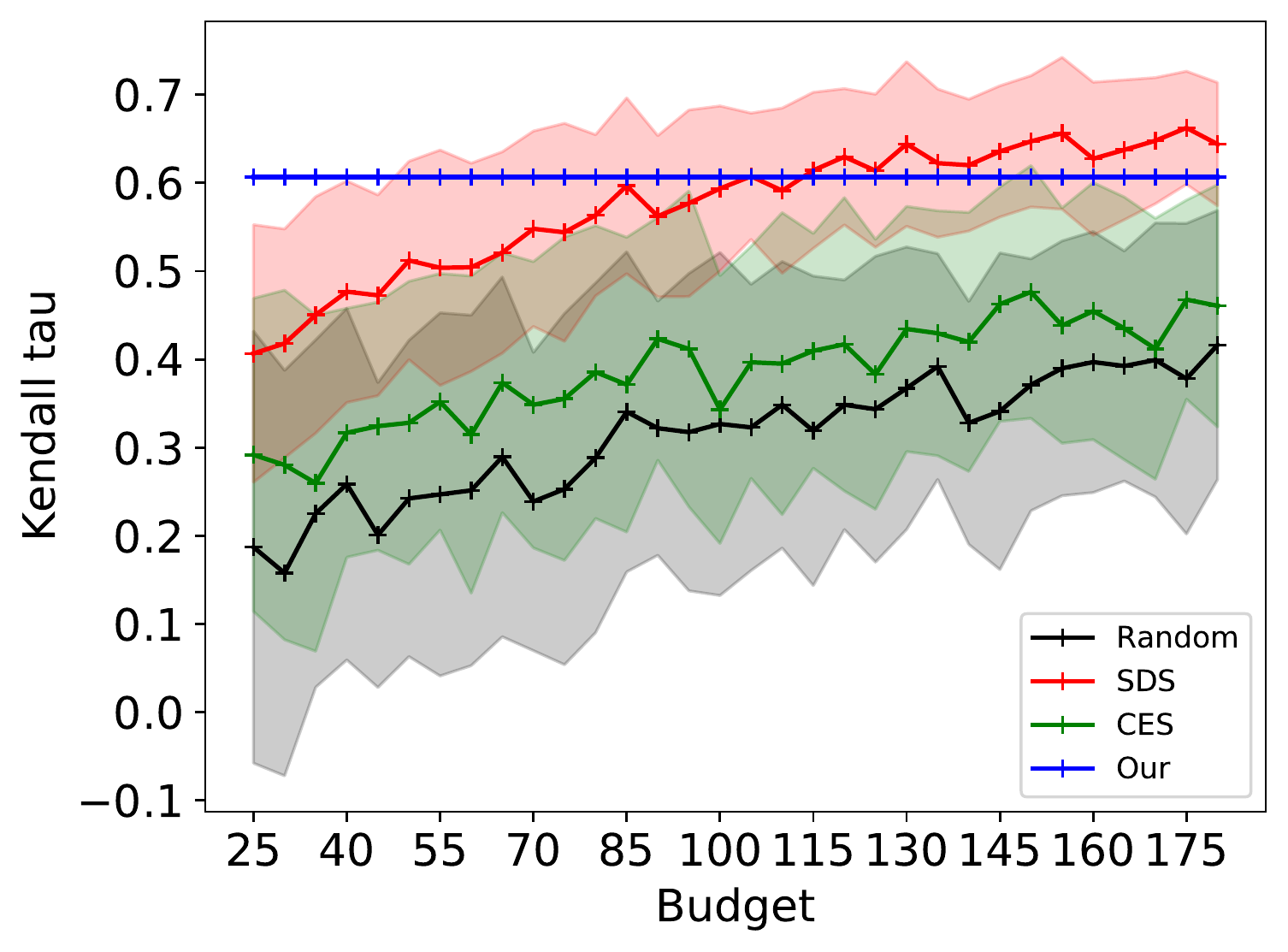}
    }
    \subfigure[CIFAR-10]{
    \includegraphics[scale=0.29]{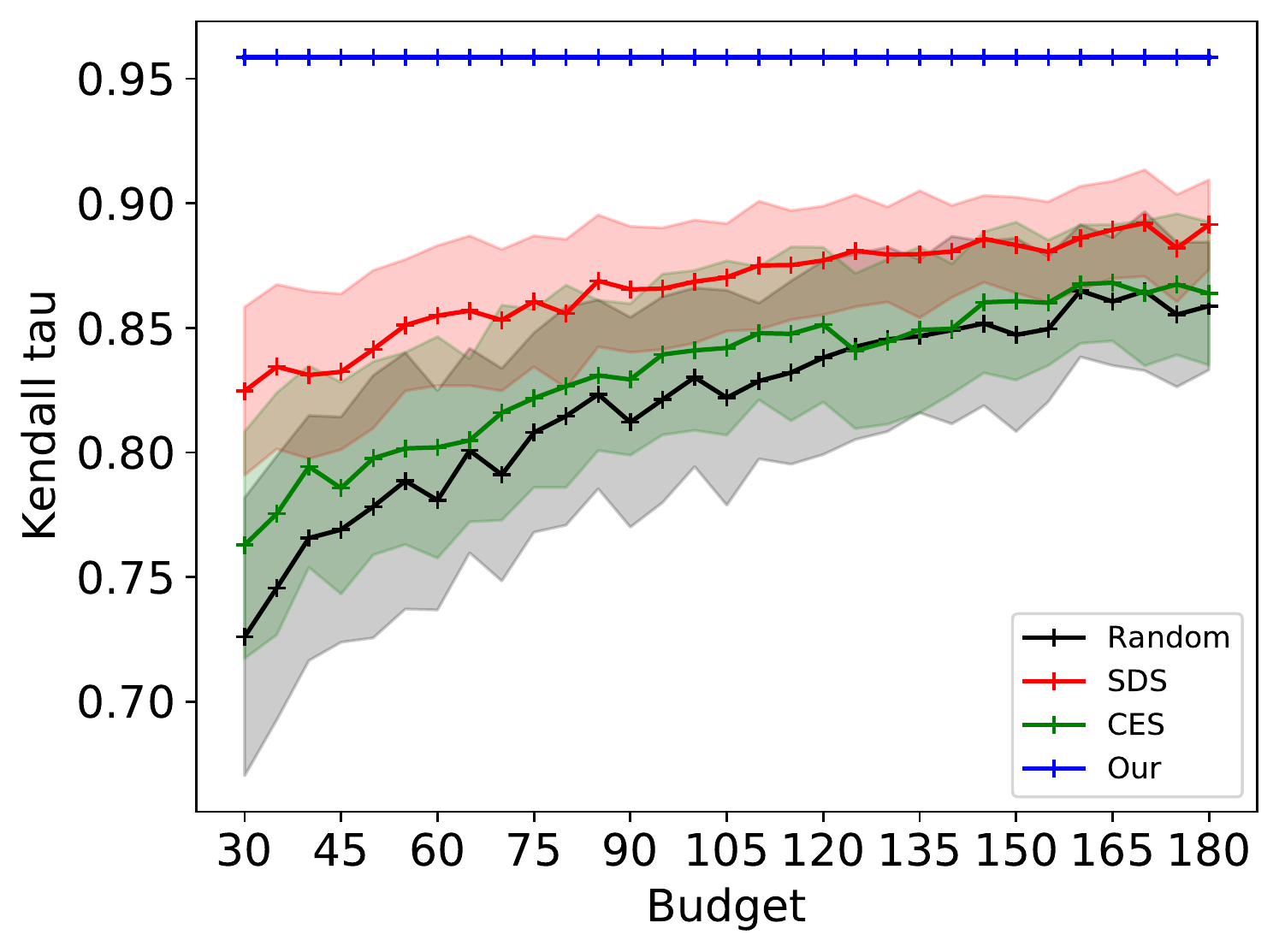}
    }\\
    \subfigure[iWildCam-ID]{
    \includegraphics[scale=0.29]{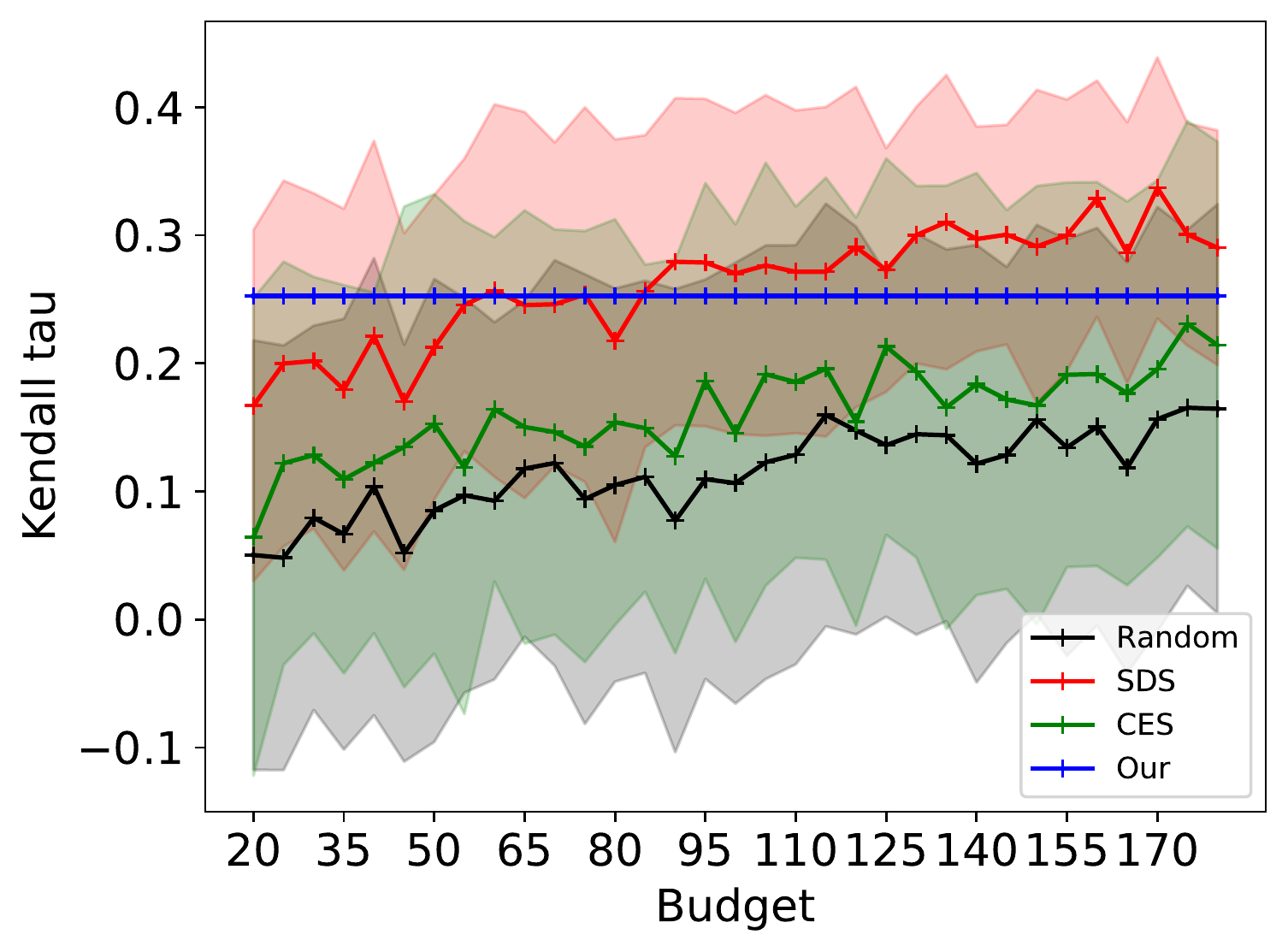} \label{fig:kentall-camera}
    }
    \subfigure[Amazon-ID]{
    \includegraphics[scale=0.29]{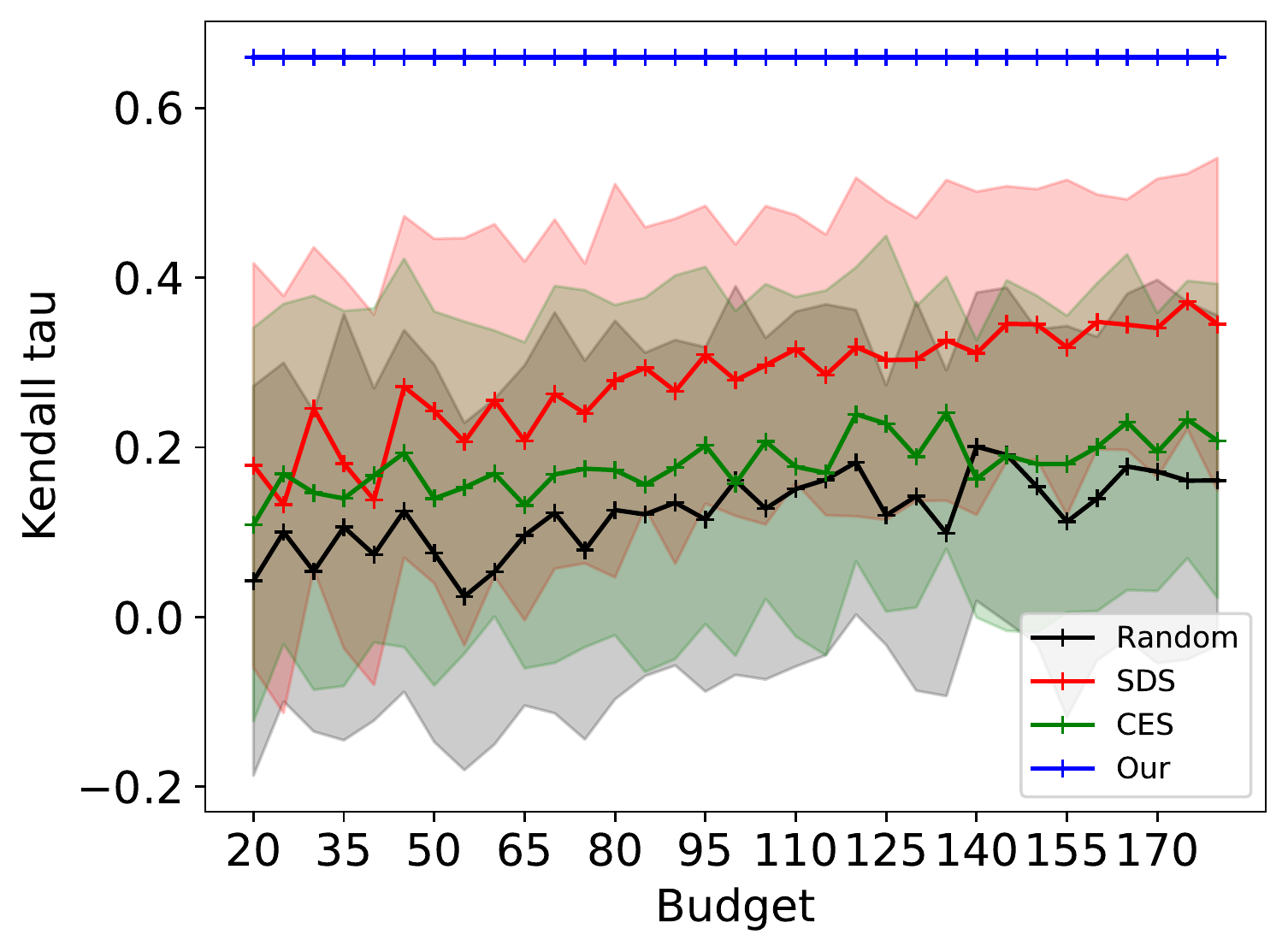}
    }
    \subfigure[Java250]{
    \includegraphics[scale=0.29]{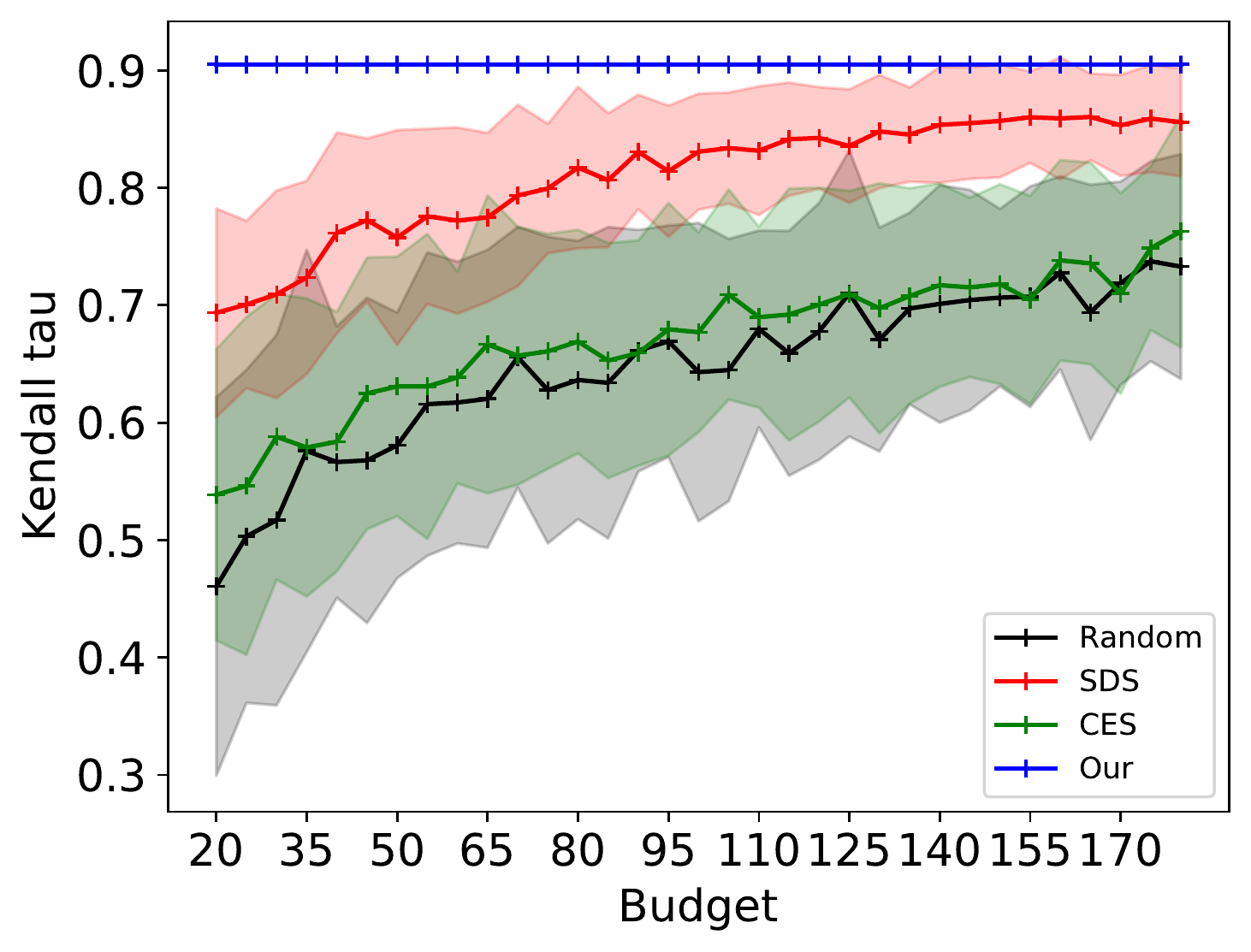}
    }
    \subfigure[C++1000]{
    \includegraphics[scale=0.29]{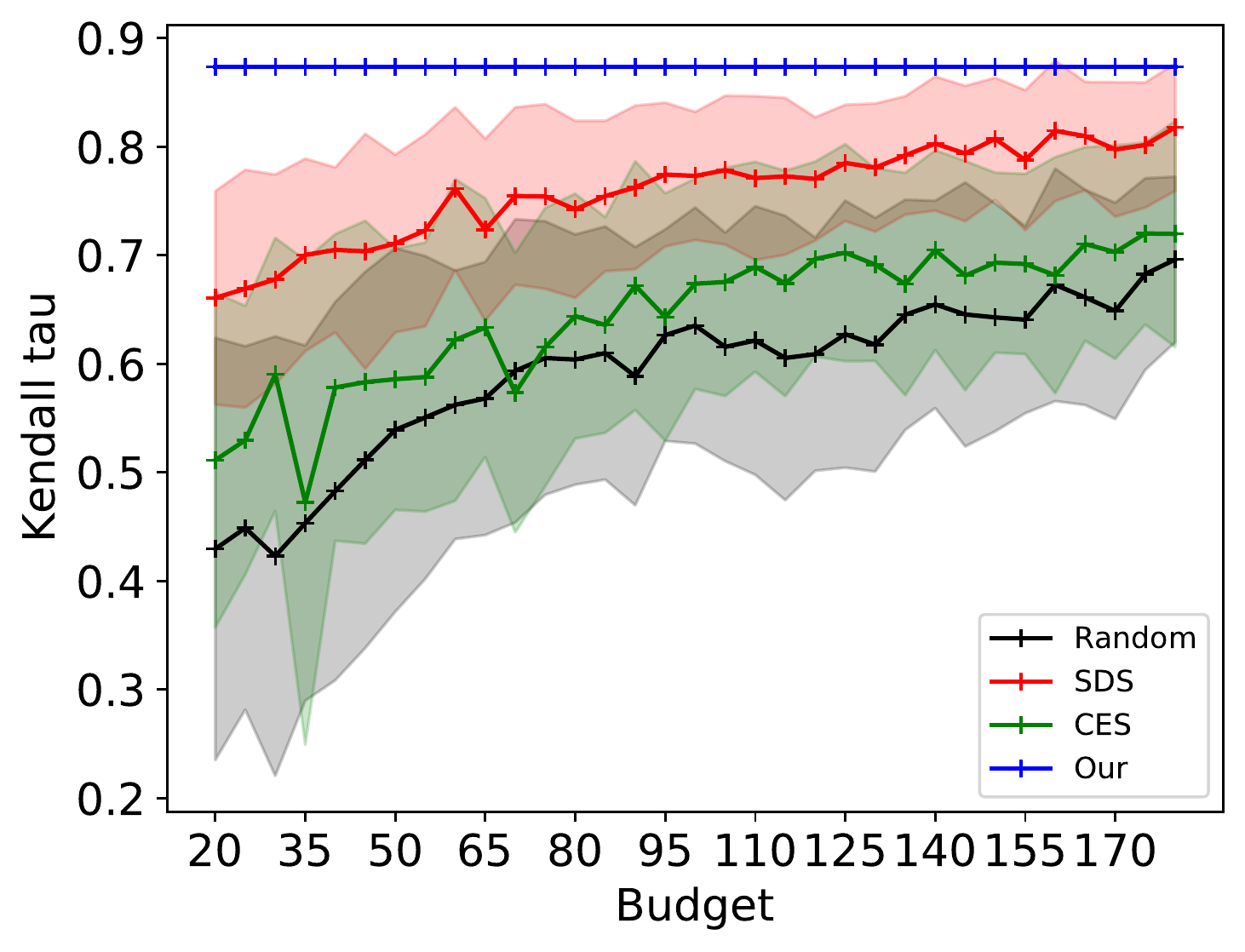}
    }
    \caption{Kendall's $\tau$ of ranking results based on ID test data. The higher the better. The shaded area represents the standard deviation. ``Budget'' is the number of labeled data (only apply to Random, SDS, and CES).}
    \label{fig:clean-kendall}
\end{figure*}

Additionally, Figure \ref{fig:clean-kendall} presents the effectiveness of all ranking methods based on Kendall's $\tau$ rank correlation. By comparison, the result confirms the conclusion drawn from the analysis based on Spearman's rank-order correlation. Namely, our approach stands out concerning the effectiveness without sampling randomness. 

Besides, to demonstrate the significance of the two statistical analyses, we calculate the corresponding $p$-value of all methods. A $p$-value lower than the common significance level of 0.05 indicates that the ranking is strongly correlated with the ground truth. Except for the iWildCam dataset, the ranking results by LaF are all strongly correlated. However, due to the effectiveness and sampling randomness, the baseline methods always achieve insignificant rankings. For the iWildCam dataset, we believe the reason is that the difference between multiple DNNs is too slight given 182 classes. For instance, the accuracy difference between the best and worst is only 1.54\% (Table \ref{tab:data-summary}). The impact of the accuracy/robustness on the ranking is investigated in Section \ref{subsec:impact}.

On the other hand, we evaluate different methods concerning identifying the top-$k$ DNNs ($k=1,3,5,10$). Table \ref{tab:clean-jaccard} lists the result of Jaccard similarity. On average, LaF achieves the best result regardless of the datasets. It is better than the worst performance by up to 0.33, 0.32, 0.33, and 0.27 in the top 1, 3, 5, and 10 rankings, respectively. Concretely, in the top-1 ranking, for datasets MNIST, Fashion-MNIST, and iWildCam, all methods (Random, SDS, and ours) are not effective (under 0.08). Remark that CES takes the best results of all models for each labeling budget when knowing the ground truth. Specifically, it takes $n$ (number of DNNs) times of labeling budget. Therefore, it sometimes outperforms others but is not applicable in practice.

\begin{table}[ht]
\caption{Jaccard similarity of ranking the top-$k$ DNNs based on the clean accuracy. For baseline methods, we report the average results over all labeling budgets. The best performance is highlighted in gray. The higher the better.}
\label{tab:clean-jaccard}
\resizebox{.9\columnwidth}{!}{
\begin{tabular}{llrrrrrll|r}
\hline
\textbf{Jaccard} & \textbf{Method} & \textbf{MNIST} & \textbf{Fashion-MNIST} & \textbf{CIFAR-10} & \textbf{iWildCam} & \textbf{Amazon} & \textbf{Java250} & \textbf{C++1000} & \textbf{Average} \\ \hline
 & Random & 0.01 & 0.02 & 0.19 & 0.07 & 0.12 & 0.20 & 0.09 & 0.10 \\
 & SDS & 0.03 & 0.07 & 0.17 & 0.02 & 0.20 & 0.36 & 0.20 & 0.15 \\
 & CES & \cellcolor[HTML]{C0C0C0}0.65 & \cellcolor[HTML]{C0C0C0}0.23 & 0.21 & \cellcolor[HTML]{C0C0C0}0.10 & 0.15 & \cellcolor[HTML]{C0C0C0}0.85 & 0.94 & \cellcolor[HTML]{C0C0C0}0.45 \\
\multirow{-4}{*}{k=1} & Our & 0.00 & 0.00 & \cellcolor[HTML]{C0C0C0}1.00 & 0.00 & \cellcolor[HTML]{C0C0C0}1.00 & 0.00 & \cellcolor[HTML]{C0C0C0}1.00 & 0.43 \\ \hline
 & Random & 0.07 & 0.12 & 0.36 & 0.14 & 0.17 & 0.27 & 0.19 & 0.19 \\
 & SDS & 0.11 & 0.24 & 0.37 & \cellcolor[HTML]{C0C0C0}0.20 & 0.31 & 0.40 & 0.29 & 0.27 \\
 & CES & \cellcolor[HTML]{C0C0C0}0.69 & \cellcolor[HTML]{C0C0C0}0.28 & 0.39 & 0.15 & 0.21 & \cellcolor[HTML]{C0C0C0}0.72 & \cellcolor[HTML]{C0C0C0}0.80 & 0.46 \\
\multirow{-4}{*}{k=3} & Our & 0.20 & 0.20 & \cellcolor[HTML]{C0C0C0}1.00 & \cellcolor[HTML]{C0C0C0}0.20 & \cellcolor[HTML]{C0C0C0}1.00 & 0.50 & 0.50 & \cellcolor[HTML]{C0C0C0}0.51 \\ \hline
 & Random & 0.11 & 0.19 & 0.50 & 0.22 & 0.25 & 0.38 & 0.30 & 0.28 \\
 & SDS & 0.17 & 0.33 & 0.60 & \cellcolor[HTML]{C0C0C0}0.33 & 0.39 & 0.50 & 0.44 & 0.39 \\
 & CES & \cellcolor[HTML]{C0C0C0}0.89 & 0.36 & 0.55 & 0.24 & 0.29 & \cellcolor[HTML]{C0C0C0}0.72 & \cellcolor[HTML]{C0C0C0}0.68 & 0.53 \\
\multirow{-4}{*}{k=5} & Our & 0.25 & \cellcolor[HTML]{C0C0C0}0.43 & \cellcolor[HTML]{C0C0C0}1.00 & 0.25 & \cellcolor[HTML]{C0C0C0}1.00 & 0.67 & 0.67 & \cellcolor[HTML]{C0C0C0}0.61 \\ \hline
 & Random & 0.23 & 0.35 & 0.80 & 0.43 & 0.43 & 0.63 & 0.58 & 0.49 \\
 & SDS & 0.41 & \cellcolor[HTML]{C0C0C0}0.55 & 0.85 & \cellcolor[HTML]{C0C0C0}0.46 & 0.49 & 0.80 & 0.79 & 0.62 \\
 & CES & \cellcolor[HTML]{C0C0C0}0.73 & 0.46 & 0.82 & 0.44 & 0.46 & 0.68 & 0.68 & 0.61 \\
\multirow{-4}{*}{k=10} & Our & 0.67 & 0.54 & \cellcolor[HTML]{C0C0C0}1.00 & 0.43 & \cellcolor[HTML]{C0C0C0}0.67 & \cellcolor[HTML]{C0C0C0}1.00 & \cellcolor[HTML]{C0C0C0}1.00 & \cellcolor[HTML]{C0C0C0}0.76 \\ \hline
\end{tabular}
}
\end{table}

\vspace{1mm}
\noindent\fcolorbox{black}{answercolor}{\begin{minipage}{\columnwidth} \textbf{Answer to RQ1}: Based on the accuracy of ID test data, LaF outperforms all the 3 selection-based baseline methods in outputting strongly correlated ranking. In addition, statistical analysis demonstrates that outperforming is significant.
\end{minipage}}

\subsection{RQ2: Effectiveness Under Distribution Shift}
\label{subsec:robustness}
For the synthetic distribution shift, Tables \ref{tab:corrup-spearman} and \ref{tab:corrup-kendall} summarize the results of Spearman's rank-order correlation and Kendall's $\tau$ on MNIST-C and CIFAR-10-C, respectively. We observe that our approach achieves the best performance in most cases, for instance, 291 of 300 cases in MNIST-C and 289 of 300 cases in CIFAR-10-C concerning Spearman's correlation, and 287 of 300 and 288 of 300, respectively, in two benchmarks concerning Kendall's $\tau$. Furthermore, as shown in RQ1, SDS performs the second best among the four ranking approaches. However, compared to random and CES, SDS tends to lose its performance in these two tables (highlighted in yellow). For example, in MNIST-C with Defocus Blur, severity-2, SDS ranks the models wrongly with a correlation of -0.03 (Table \ref{tab:corrup-spearman}), while both random and CES can achieve comparable ranking performance as LaF. In short, this existing state-of-the-art approach is sensitive to artificial distribution shifts, which calls for the testing under distribution shifts of existing approaches.  Concerning the Jaccard similarity, in the 75 corruptions of MNIST-C, both LaF and CES outperform the random sampling and SDS to identify the top DNNs precisely. In CIFAR-10-C, LaF achieves the best performance (similarity of 1) in most cases (173 of 300).

\begin{table*}[ht]
\caption{Spearman's correlation coefficient of ranking results based on MNIST-C and CIFAR-10-C. For baseline methods, we compute the average and standard deviation over all labeling budgets and 50-repetition experiments. The best performance is highlighted in gray. Values highlighted in yellow indicate CES or random outperform SDS. The higher the better.}
\label{tab:corrup-spearman}
\resizebox{\textwidth}{!}{
\begin{tabular}{lcccccccccccccccccccc}
\hline
 & \multicolumn{4}{c|}{\textbf{Severity=1}} & \multicolumn{4}{c|}{\textbf{Severity=2}} & \multicolumn{4}{c|}{\textbf{Severity=3}} & \multicolumn{4}{c|}{\textbf{Severity=4}} & \multicolumn{4}{c}{\textbf{Severity=5}} \\ \cline{2-21} 
\multirow{-2}{*}{\textbf{Corruption Type}} & \textbf{Random} & \textbf{SDS} & \textbf{CES} & \multicolumn{1}{c|}{\textbf{Our}} & \textbf{Random} & \textbf{SDS} & \textbf{CES} & \multicolumn{1}{c|}{\textbf{Our}} & \textbf{Random} & \textbf{SDS} & \textbf{CES} & \multicolumn{1}{c|}{\textbf{Our}} & \textbf{Random} & \textbf{SDS} & \textbf{CES} & \multicolumn{1}{c|}{\textbf{Our}} & \textbf{Random} & \textbf{SDS} & \textbf{CES} & \textbf{Our} \\ \hline
\multicolumn{21}{c}{\textbf{MNIST-C}} \\ \hline
Gaussian Noise & { 0.80$\pm$0.08} & { 0.89$\pm$0.04} & { 0.83$\pm$0.06} & \multicolumn{1}{c|}{\cellcolor[HTML]{C0C0C0}{ 0.94}} & { 0.86$\pm$0.06} & { 0.93$\pm$0.03} & { 0.88$\pm$0.05} & \multicolumn{1}{c|}{\cellcolor[HTML]{C0C0C0}{ 0.96}} & { 0.91$\pm$0.04} & { 0.93$\pm$0.03} & { 0.92$\pm$0.03} & \multicolumn{1}{c|}{\cellcolor[HTML]{C0C0C0}{ 0.97}} & { 0.94$\pm$0.02} & { 0.96$\pm$0.02} & { 0.95$\pm$0.02} & \multicolumn{1}{c|}{\cellcolor[HTML]{C0C0C0}{ 0.98}} & { 0.98$\pm$0.01} & { 0.98$\pm$0.01} & { 0.98$\pm$0.01} & \cellcolor[HTML]{C0C0C0}{ 0.99} \\
Shot Noise & { 0.79$\pm$0.09} & { 0.86$\pm$0.06} & { 0.83$\pm$0.07} & \multicolumn{1}{c|}{\cellcolor[HTML]{C0C0C0}{ 0.89}} & { 0.78$\pm$0.09} & { 0.86$\pm$0.06} & { 0.83$\pm$0.07} & \multicolumn{1}{c|}{\cellcolor[HTML]{C0C0C0}{ 0.89}} & { 0.78$\pm$0.09} & { 0.86$\pm$0.06} & { 0.83$\pm$0.07} & \multicolumn{1}{c|}{\cellcolor[HTML]{C0C0C0}{ 0.86}} & { 0.75$\pm$0.09} & { 0.82$\pm$0.06} & { 0.81$\pm$0.07} & \multicolumn{1}{c|}{\cellcolor[HTML]{C0C0C0}{ 0.89}} & { 0.69$\pm$0.09} & \cellcolor[HTML]{FFFFC7}{ 0.75$\pm$0.06} & { 0.84$\pm$0.06} & \cellcolor[HTML]{C0C0C0}{ 0.97} \\
Impulse Noise & { 0.69$\pm$0.09} & \cellcolor[HTML]{FFFFC7}{ 0.75$\pm$0.07} & { 0.89$\pm$0.05} & \multicolumn{1}{c|}{\cellcolor[HTML]{C0C0C0}{ 0.95}} & { 0.56$\pm$0.10} & \cellcolor[HTML]{FFFFC7}{ 0.63$\pm$0.08} & { 0.92$\pm$0.03} & \multicolumn{1}{c|}{\cellcolor[HTML]{C0C0C0}{ 0.96}} & { 0.43$\pm$0.10} & \cellcolor[HTML]{FFFFC7}{ 0.49$\pm$0.09} & { 0.94$\pm$0.02} & \multicolumn{1}{c|}{\cellcolor[HTML]{C0C0C0}{ 0.98}} & { 0.20$\pm$0.11} & \cellcolor[HTML]{FFFFC7}{ 0.25$\pm$0.10} & { 0.97$\pm$0.01} & \multicolumn{1}{c|}{\cellcolor[HTML]{C0C0C0}{ 0.98}} & { 0.01$\pm$0.11} & \cellcolor[HTML]{FFFFC7}{ 0.06$\pm$0.10} & { 0.98$\pm$0.01} & \cellcolor[HTML]{C0C0C0}{ 0.98} \\
Defocus Blur & { 0.92$\pm$0.03} & \cellcolor[HTML]{FFFFC7}{ 0.90$\pm$0.18} & { 0.93$\pm$0.03} & \multicolumn{1}{c|}{\cellcolor[HTML]{C0C0C0}{ 0.98}} & { 0.94$\pm$0.02} & \cellcolor[HTML]{FFFFC7}{ -0.03$\pm$0.10} & { 0.95$\pm$0.02} & \multicolumn{1}{c|}{\cellcolor[HTML]{C0C0C0}{ 0.97}} & { 0.90$\pm$0.05} & \cellcolor[HTML]{FFFFC7}{ -0.23$\pm$0.07} & \cellcolor[HTML]{C0C0C0}{ 0.92$\pm$0.03} & \multicolumn{1}{c|}{{ -0.02}} & { 0.75$\pm$0.10} & \cellcolor[HTML]{FFFFC7}{ -0.17$\pm$0.09} & \cellcolor[HTML]{C0C0C0}{ 0.80$\pm$0.08} & \multicolumn{1}{c|}{{ -0.07}} & { 0.69$\pm$0.14} & \cellcolor[HTML]{FFFFC7}{ 0.03$\pm$0.09} & \cellcolor[HTML]{C0C0C0}{ 0.76$\pm$0.10} & { 0.03} \\
Frosted Glass Blur & { 0.31$\pm$0.12} & \cellcolor[HTML]{FFFFC7}{ 0.34$\pm$0.10} & { 0.92$\pm$0.03} & \multicolumn{1}{c|}{\cellcolor[HTML]{C0C0C0}{ 0.98}} & { 0.18$\pm$0.12} & \cellcolor[HTML]{FFFFC7}{ 0.22$\pm$0.10} & { 0.92$\pm$0.03} & \multicolumn{1}{c|}{\cellcolor[HTML]{C0C0C0}{ 0.98}} & { -0.48$\pm$0.10} & \cellcolor[HTML]{FFFFC7}{ -0.49$\pm$0.08} & { 0.95$\pm$0.02} & \multicolumn{1}{c|}{\cellcolor[HTML]{C0C0C0}{ 0.93}} & { -0.50$\pm$0.09} & \cellcolor[HTML]{FFFFC7}{ -0.50$\pm$0.07} & { 0.95$\pm$0.02} & \multicolumn{1}{c|}{\cellcolor[HTML]{C0C0C0}{ 0.87}} & { -0.31$\pm$0.10} & \cellcolor[HTML]{FFFFC7}{ -0.34$\pm$0.08} & { 0.92$\pm$0.03} & \cellcolor[HTML]{C0C0C0}{ 0.09} \\
Motion Blur & { 0.73$\pm$0.08} & \cellcolor[HTML]{FFFFC7}{ 0.78$\pm$0.05} & { 0.93$\pm$0.03} & \multicolumn{1}{c|}{\cellcolor[HTML]{C0C0C0}{ 0.94}} & { 0.70$\pm$0.08} & \cellcolor[HTML]{FFFFC7}{ 0.73$\pm$0.06} & { 0.96$\pm$0.01} & \multicolumn{1}{c|}{\cellcolor[HTML]{C0C0C0}{ 0.90}} & { 0.56$\pm$0.08} & \cellcolor[HTML]{FFFFC7}{ 0.59$\pm$0.07} & { 0.96$\pm$0.01} & \multicolumn{1}{c|}{\cellcolor[HTML]{C0C0C0}{ 0.87}} & { 0.46$\pm$0.09} & \cellcolor[HTML]{FFFFC7}{ 0.49$\pm$0.08} & { 0.95$\pm$0.02} & \multicolumn{1}{c|}{\cellcolor[HTML]{C0C0C0}{ 0.50}} & { 0.45$\pm$0.09} & \cellcolor[HTML]{FFFFC7}{ 0.49$\pm$0.08} & \cellcolor[HTML]{C0C0C0}{ 0.94$\pm$0.03} & { -0.18} \\
Zoom Blur & { 0.78$\pm$0.08} & \cellcolor[HTML]{FFFFC7}{ 0.84$\pm$0.06} & { 0.86$\pm$0.06} & \multicolumn{1}{c|}{\cellcolor[HTML]{C0C0C0}{ 0.92}} & { 0.77$\pm$0.08} & \cellcolor[HTML]{FFFFC7}{ 0.84$\pm$0.05} & { 0.87$\pm$0.05} & \multicolumn{1}{c|}{\cellcolor[HTML]{C0C0C0}0.91} & 0.77$\pm$0.08 & \cellcolor[HTML]{FFFFC7}0.83$\pm$0.06 & 0.87$\pm$0.05 & \multicolumn{1}{c|}{\cellcolor[HTML]{C0C0C0}0.91} & 0.77$\pm$0.08 & \cellcolor[HTML]{FFFFC7}0.82$\pm$0.06 & 0.89$\pm$0.04 & \multicolumn{1}{c|}{\cellcolor[HTML]{C0C0C0}0.90} & 0.75$\pm$0.08 & \cellcolor[HTML]{FFFFC7}0.81$\pm$0.06 & { 0.89$\pm$0.04} & \cellcolor[HTML]{C0C0C0}{ 0.91} \\
Snow & { 0.55$\pm$0.09} & \cellcolor[HTML]{FFFFC7}{ 0.57$\pm$0.08} & { 0.90$\pm$0.04} & \multicolumn{1}{c|}{\cellcolor[HTML]{C0C0C0}{ 0.95}} & { 0.48$\pm$0.10} & \cellcolor[HTML]{FFFFC7}{ 0.48$\pm$0.09} & { 0.96$\pm$0.02} & \multicolumn{1}{c|}{\cellcolor[HTML]{C0C0C0}0.99} & 0.50$\pm$0.10 & \cellcolor[HTML]{FFFFC7}0.50$\pm$0.08 & 0.96$\pm$0.02 & \multicolumn{1}{c|}{\cellcolor[HTML]{C0C0C0}0.99} & 0.51$\pm$0.10 & \cellcolor[HTML]{FFFFC7}0.51$\pm$0.08 & 0.96$\pm$0.02 & \multicolumn{1}{c|}{\cellcolor[HTML]{C0C0C0}0.98} & 0.48$\pm$0.10 & \cellcolor[HTML]{FFFFC7}0.47$\pm$0.09 & { 0.97$\pm$0.01} & \cellcolor[HTML]{C0C0C0}{ 0.99} \\
Frost & { 0.44$\pm$0.10} & \cellcolor[HTML]{FFFFC7}{ 0.43$\pm$0.09} & { 0.96$\pm$0.02} & \multicolumn{1}{c|}{\cellcolor[HTML]{C0C0C0}{ 0.99}} & { 0.44$\pm$0.10} & \cellcolor[HTML]{FFFFC7}{ 0.42$\pm$0.09} & { 0.98$\pm$0.01} & \multicolumn{1}{c|}{\cellcolor[HTML]{C0C0C0}0.99} & 0.44$\pm$0.11 & \cellcolor[HTML]{FFFFC7}0.42$\pm$0.08 & 0.98$\pm$0.01 & \multicolumn{1}{c|}{\cellcolor[HTML]{C0C0C0}0.99} & 0.43$\pm$0.11 & \cellcolor[HTML]{FFFFC7}0.42$\pm$0.09 & 0.99$\pm$0.01 & \multicolumn{1}{c|}{\cellcolor[HTML]{C0C0C0}0.99} & 0.44$\pm$0.11 & \cellcolor[HTML]{FFFFC7}0.42$\pm$0.08 & 0.98$\pm$0.01 & \cellcolor[HTML]{C0C0C0}0.99 \\
Fog & { 0.43$\pm$0.10} & \cellcolor[HTML]{FFFFC7}{ 0.43$\pm$0.09} & { 0.99} & \multicolumn{1}{c|}{\cellcolor[HTML]{C0C0C0}{ 0.99}} & { 0.43$\pm$0.11} & \cellcolor[HTML]{FFFFC7}{ 0.42$\pm$0.09} & { 0.99$\pm$0.01} & \multicolumn{1}{c|}{\cellcolor[HTML]{C0C0C0}0.99} & 0.39$\pm$0.10 & \cellcolor[HTML]{FFFFC7}0.38$\pm$0.08 & 0.98$\pm$0.01 & \multicolumn{1}{c|}{\cellcolor[HTML]{C0C0C0}0.95} & 0.4$\pm$0.11 & \cellcolor[HTML]{FFFFC7}0.39$\pm$0.08 & 0.98$\pm$0.01 & \multicolumn{1}{c|}{\cellcolor[HTML]{C0C0C0}0.91} & 0.33$\pm$0.11 & \cellcolor[HTML]{FFFFC7}0.32$\pm$0.09 & \cellcolor[HTML]{C0C0C0}0.97$\pm$0.01 & -0.14 \\
Brightness & { 0.87$\pm$0.05} & \cellcolor[HTML]{FFFFC7}{ 0.58$\pm$0.08} & { 0.89$\pm$0.04} & \multicolumn{1}{c|}{\cellcolor[HTML]{C0C0C0}{ 0.96}} & { 0.92$\pm$0.03} & \cellcolor[HTML]{FFFFC7}{ 0.45$\pm$0.09} & { 0.93$\pm$0.03} & \multicolumn{1}{c|}{\cellcolor[HTML]{C0C0C0}0.98} & 0.96$\pm$0.02 & \cellcolor[HTML]{FFFFC7}0.44$\pm$0.09 & 0.97$\pm$0.01 & \multicolumn{1}{c|}{\cellcolor[HTML]{C0C0C0}0.99} & 0.98$\pm$0.01 & \cellcolor[HTML]{FFFFC7}0.43$\pm$0.09 & 0.98$\pm$0.01 & \multicolumn{1}{c|}{\cellcolor[HTML]{C0C0C0}1} & 0.98$\pm$0.01 & \cellcolor[HTML]{FFFFC7}0.44$\pm$0.08 & 0.99$\pm$0.01 & \cellcolor[HTML]{C0C0C0}0.98 \\
Contrast & { 0.91$\pm$0.04} & \cellcolor[HTML]{FFFFC7}{ 0.52$\pm$0.08} & { 0.93$\pm$0.03} & \multicolumn{1}{c|}{\cellcolor[HTML]{C0C0C0}{ 0.98}} & { 0.94$\pm$0.03} & \cellcolor[HTML]{FFFFC7}{ 0.48$\pm$0.09} & { 0.95$\pm$0.02} & \multicolumn{1}{c|}{\cellcolor[HTML]{C0C0C0}0.99} & 0.98$\pm$0.01 & \cellcolor[HTML]{FFFFC7}0.39$\pm$0.10 & 0.98$\pm$0.01 & \multicolumn{1}{c|}{\cellcolor[HTML]{C0C0C0}1} & 0.96$\pm$0.02 & \cellcolor[HTML]{FFFFC7}0.24$\pm$0.09 & 0.97$\pm$0.01 & \multicolumn{1}{c|}{\cellcolor[HTML]{C0C0C0}0.96} & 0.89$\pm$0.06 & \cellcolor[HTML]{FFFFC7}0.13$\pm$0.08 & \cellcolor[HTML]{C0C0C0}0.91$\pm$0.05 & 0.23 \\
Elastic & { 0.94$\pm$0.03} & \cellcolor[HTML]{FFFFC7}{ 0.83$\pm$0.06} & \cellcolor[HTML]{C0C0C0}{ 0.95$\pm$0.02} & \multicolumn{1}{c|}{0.83} & { 0.30$\pm$0.20} & \cellcolor[HTML]{C0C0C0}{ 0.40$\pm$0.08} & { 0.38$\pm$0.19} & \multicolumn{1}{c|}{-0.01} & 0.88$\pm$0.05 & \cellcolor[HTML]{FFFFC7}0.52$\pm$0.08 & 0.89$\pm$0.04 & \multicolumn{1}{c|}{\cellcolor[HTML]{C0C0C0}0.97} & 0.94$\pm$0.03 & \cellcolor[HTML]{FFFFC7}-0.29$\pm$0.08 & 0.94$\pm$0.02 & \multicolumn{1}{c|}{\cellcolor[HTML]{C0C0C0}0.97} & 0.96$\pm$0.01 & \cellcolor[HTML]{FFFFC7}-0.65$\pm$0.06 & \cellcolor[HTML]{C0C0C0}0.97$\pm$0.01 & -0.45 \\
Pixelate & { 0.77$\pm$0.08} & { 0.83$\pm$0.05} & { 0.82$\pm$0.06} & \multicolumn{1}{c|}{\cellcolor[HTML]{C0C0C0}{ 0.86}} & { 0.76$\pm$0.08} & \cellcolor[HTML]{FFFFC7}{ 0.82$\pm$0.06} & { 0.84$\pm$0.06} & \multicolumn{1}{c|}{\cellcolor[HTML]{C0C0C0}0.91} & 0.57$\pm$0.10 & \cellcolor[HTML]{FFFFC7}0.58$\pm$0.08 & 0.89$\pm$0.04 & \multicolumn{1}{c|}{\cellcolor[HTML]{C0C0C0}0.97} & 0.13$\pm$0.13 & \cellcolor[HTML]{FFFFC7}0.09$\pm$0.11 & 0.87$\pm$0.05 & \multicolumn{1}{c|}{\cellcolor[HTML]{C0C0C0}0.98} & -0.31$\pm$0.13 & \cellcolor[HTML]{FFFFC7}-0.34$\pm$0.10 & 0.90$\pm$0.04 & \cellcolor[HTML]{C0C0C0}0.88 \\
JPEG & { 0.78$\pm$0.09} & { 0.86$\pm$0.06} & { 0.83$\pm$0.07} & \multicolumn{1}{c|}{\cellcolor[HTML]{C0C0C0}{ 0.91}} & { 0.78$\pm$0.08} & { 0.85$\pm$0.06} & { 0.80$\pm$0.06} & \multicolumn{1}{c|}{\cellcolor[HTML]{C0C0C0}0.89} & 0.78$\pm$0.08 & 0.85$\pm$0.05 & 0.82$\pm$0.07 & \multicolumn{1}{c|}{\cellcolor[HTML]{C0C0C0}0.90} & 0.77$\pm$0.08 & 0.84$\pm$0.05 & 0.82$\pm$0.06 & \multicolumn{1}{c|}{\cellcolor[HTML]{C0C0C0}0.90} & 0.77$\pm$0.08 & 0.83$\pm$0.06 & 0.81$\pm$0.06 & \cellcolor[HTML]{C0C0C0}0.89 \\ \hline
\multicolumn{21}{c}{\textbf{CIFAR-10-C}} \\ \hline
Gaussian Noise & 0.83$\pm$0.07 & 0.88$\pm$0.03 & 0.86$\pm$0.05 & \multicolumn{1}{c|}{\cellcolor[HTML]{C0C0C0}0.98} & 0.88$\pm$0.06 & \cellcolor[HTML]{C0C0C0}0.95$\pm$0.02 & 0.91$\pm$0.04 & \multicolumn{1}{c|}{0.91} & 0.94$\pm$0.02 & \cellcolor[HTML]{FFFFC7}0.94$\pm$0.01 & 0.95$\pm$0.02 & \multicolumn{1}{c|}{\cellcolor[HTML]{C0C0C0}0.95} & 0.94$\pm$0.02 & \cellcolor[HTML]{FFFFC7}0.93$\pm$0.01 & 0.95$\pm$0.01 & \multicolumn{1}{c|}{\cellcolor[HTML]{C0C0C0}0.94} & 0.95$\pm$0.02 & \cellcolor[HTML]{FFFFC7}0.93$\pm$0.01 & \cellcolor[HTML]{C0C0C0}0.95$\pm$0.01 & 0.93 \\
Shot Noise & 0.90$\pm$0.04 & 0.94$\pm$0.01 & 0.91$\pm$0.03 & \multicolumn{1}{c|}{\cellcolor[HTML]{C0C0C0}0.99} & 0.81$\pm$0.08 & 0.86$\pm$0.04 & 0.84$\pm$0.06 & \multicolumn{1}{c|}{\cellcolor[HTML]{C0C0C0}0.97} & 0.91$\pm$0.05 & \cellcolor[HTML]{C0C0C0}0.94$\pm$0.01 & 0.93$\pm$0.03 & \multicolumn{1}{c|}{0.93} & 0.93$\pm$0.03 & \cellcolor[HTML]{FFFFC7}0.93$\pm$0.01 & 0.94$\pm$0.02 & \multicolumn{1}{c|}{\cellcolor[HTML]{C0C0C0}0.96} & 0.94$\pm$0.02 & \cellcolor[HTML]{FFFFC7}0.92$\pm$0.02 & \cellcolor[HTML]{C0C0C0}0.95$\pm$0.02 & 0.91 \\
Impulse Noise & 0.88$\pm$0.05 & 0.92$\pm$0.02 & 0.90$\pm$0.04 & \multicolumn{1}{c|}{\cellcolor[HTML]{C0C0C0}0.99} & 0.84$\pm$0.06 & 0.90$\pm$0.04 & 0.86$\pm$0.05 & \multicolumn{1}{c|}{\cellcolor[HTML]{C0C0C0}0.97} & 0.86$\pm$0.06 & 0.89$\pm$0.03 & 0.88$\pm$0.04 & \multicolumn{1}{c|}{\cellcolor[HTML]{C0C0C0}0.92} & 0.92$\pm$0.03 & \cellcolor[HTML]{FFFFC7}0.86$\pm$0.02 & \cellcolor[HTML]{C0C0C0}0.93$\pm$0.02 & \multicolumn{1}{c|}{0.85} & 0.93$\pm$0.03 & \cellcolor[HTML]{FFFFC7}0.92$\pm$0.02 & \cellcolor[HTML]{C0C0C0}0.95$\pm$0.02 & 0.85 \\
Defocus Blur & 0.94$\pm$0.02 & 0.97$\pm$0.01 & 0.95$\pm$0.02 & \multicolumn{1}{c|}{\cellcolor[HTML]{C0C0C0}1} & 0.93$\pm$0.02 & 0.97$\pm$0.01 & 0.94$\pm$0.02 & \multicolumn{1}{c|}{\cellcolor[HTML]{C0C0C0}0.98} & 0.92$\pm$0.03 & 0.93$\pm$0.02 & 0.93$\pm$0.02 & \multicolumn{1}{c|}{\cellcolor[HTML]{C0C0C0}0.99} & 0.93$\pm$0.02 & 0.94$\pm$0.02 & 0.94$\pm$0.02 & \multicolumn{1}{c|}{\cellcolor[HTML]{C0C0C0}0.99} & 0.94$\pm$0.02 & 0.96$\pm$0.01 & 0.95$\pm$0.02 & \cellcolor[HTML]{C0C0C0}0.98 \\
Frosted Glass Blur & 0.83$\pm$0.05 & 0.89$\pm$0.03 & 0.85$\pm$0.05 & \multicolumn{1}{c|}{\cellcolor[HTML]{C0C0C0}0.97} & 0.83$\pm$0.06 & 0.93$\pm$0.02 & 0.85$\pm$0.05 & \multicolumn{1}{c|}{\cellcolor[HTML]{C0C0C0}0.98} & 0.80$\pm$0.07 & 0.91$\pm$0.03 & 0.83$\pm$0.06 & \multicolumn{1}{c|}{\cellcolor[HTML]{C0C0C0}0.98} & 0.89$\pm$0.06 & \cellcolor[HTML]{C0C0C0}0.96$\pm$0.01 & 0.92$\pm$0.03 & \multicolumn{1}{c|}{0.94} & 0.86$\pm$0.07 & \cellcolor[HTML]{C0C0C0}0.96$\pm$0.02 & 0.90$\pm$0.04 & 0.93 \\
Motion Blur & 0.91$\pm$0.03 & 0.95$\pm$0.01 & 0.92$\pm$0.03 & \multicolumn{1}{c|}{\cellcolor[HTML]{C0C0C0}0.95} & 0.91$\pm$0.03 & \cellcolor[HTML]{FFFFC7}0.91$\pm$0.02 & 0.92$\pm$0.02 & \multicolumn{1}{c|}{\cellcolor[HTML]{C0C0C0}0.99} & 0.92$\pm$0.03 & 0.94$\pm$0.02 & 0.94$\pm$0.02 & \multicolumn{1}{c|}{\cellcolor[HTML]{C0C0C0}0.99} & 0.93$\pm$0.02 & 0.94$\pm$0.02 & 0.94$\pm$0.02 & \multicolumn{1}{c|}{\cellcolor[HTML]{C0C0C0}0.98} & 0.93$\pm$0.02 & 0.95$\pm$0.02 & 0.94$\pm$0.02 & \cellcolor[HTML]{C0C0C0}0.99 \\
Zoom Blur & 0.91$\pm$0.03 & \cellcolor[HTML]{FFFFC7}0.91$\pm$0.02 & 0.92$\pm$0.02 & \multicolumn{1}{c|}{\cellcolor[HTML]{C0C0C0}1} & 0.92$\pm$0.02 & 0.93$\pm$0.02 & 0.93$\pm$0.02 & \multicolumn{1}{c|}{\cellcolor[HTML]{C0C0C0}0.99} & 0.93$\pm$0.02 & 0.94$\pm$0.02 & 0.94$\pm$0.02 & \multicolumn{1}{c|}{\cellcolor[HTML]{C0C0C0}0.99} & 0.93$\pm$0.02 & 0.95$\pm$0.02 & 0.94$\pm$0.02 & \multicolumn{1}{c|}{\cellcolor[HTML]{C0C0C0}0.99} & 0.93$\pm$0.02 & 0.96$\pm$0.01 & 0.94$\pm$0.02 & \cellcolor[HTML]{C0C0C0}0.99 \\
Snow & 0.92$\pm$0.03 & 0.94$\pm$0.01 & 0.93$\pm$0.02 & \multicolumn{1}{c|}{\cellcolor[HTML]{C0C0C0}0.99} & 0.91$\pm$0.03 & 0.93$\pm$0.02 & 0.92$\pm$0.03 & \multicolumn{1}{c|}{\cellcolor[HTML]{C0C0C0}0.99} & 0.90$\pm$0.03 & 0.91$\pm$0.02 & 0.91$\pm$0.03 & \multicolumn{1}{c|}{\cellcolor[HTML]{C0C0C0}0.99} & 0.90$\pm$0.04 & 0.93$\pm$0.02 & 0.91$\pm$0.03 & \multicolumn{1}{c|}{\cellcolor[HTML]{C0C0C0}0.99} & 0.90$\pm$0.04 & \cellcolor[HTML]{FFFFC7}0.89$\pm$0.03 & 0.91$\pm$0.03 & \cellcolor[HTML]{C0C0C0}0.99 \\
Frost & 0.93$\pm$0.03 & 0.96$\pm$0.01 & 0.94$\pm$0.02 & \multicolumn{1}{c|}{\cellcolor[HTML]{C0C0C0}0.99} & 0.92$\pm$0.03 & 0.95$\pm$0.01 & 0.93$\pm$0.02 & \multicolumn{1}{c|}{\cellcolor[HTML]{C0C0C0}0.98} & 0.91$\pm$0.03 & \cellcolor[HTML]{FFFFC7}0.90$\pm$0.02 & 0.93$\pm$0.03 & \multicolumn{1}{c|}{\cellcolor[HTML]{C0C0C0}0.98} & 0.90$\pm$0.05 & \cellcolor[HTML]{FFFFC7}0.90$\pm$0.03 & 0.92$\pm$0.03 & \multicolumn{1}{c|}{\cellcolor[HTML]{C0C0C0}0.99} & 0.86$\pm$0.05 & \cellcolor[HTML]{FFFFC7}0.87$\pm$0.03 & 0.89$\pm$0.04 & \cellcolor[HTML]{C0C0C0}0.94 \\
Fog & 0.94$\pm$0.02 & 0.97$\pm$0.01 & 0.95$\pm$0.02 & \multicolumn{1}{c|}{\cellcolor[HTML]{C0C0C0}1} & 0.95$\pm$0.02 & 0.97$\pm$0.01 & 0.95$\pm$0.02 & \multicolumn{1}{c|}{\cellcolor[HTML]{C0C0C0}1} & 0.95$\pm$0.02 & 0.97$\pm$0.01 & 0.96$\pm$0.01 & \multicolumn{1}{c|}{\cellcolor[HTML]{C0C0C0}0.99} & 0.96$\pm$0.02 & 0.97$\pm$0.01 & 0.96$\pm$0.01 & \multicolumn{1}{c|}{\cellcolor[HTML]{C0C0C0}0.99} & 0.95$\pm$0.02 & 0.96$\pm$0.01 & 0.96$\pm$0.01 & \cellcolor[HTML]{C0C0C0}0.96 \\
Brightness & 0.94$\pm$0.02 & 0.97$\pm$0.01 & 0.95$\pm$0.02 & \multicolumn{1}{c|}{\cellcolor[HTML]{C0C0C0}0.99} & 0.94$\pm$0.02 & 0.97$\pm$0.01 & 0.95$\pm$0.02 & \multicolumn{1}{c|}{\cellcolor[HTML]{C0C0C0}0.99} & 0.94$\pm$0.02 & 0.97$\pm$0.01 & 0.95$\pm$0.02 & \multicolumn{1}{c|}{\cellcolor[HTML]{C0C0C0}0.99} & 0.94$\pm$0.02 & 0.97$\pm$0.01 & 0.95$\pm$0.02 & \multicolumn{1}{c|}{\cellcolor[HTML]{C0C0C0}0.99} & 0.94$\pm$0.02 & 0.97$\pm$0.01 & 0.95$\pm$0.02 & \cellcolor[HTML]{C0C0C0}0.99 \\
Contrast & 0.94$\pm$0.02 & 0.97$\pm$0.01 & 0.95$\pm$0.02 & \multicolumn{1}{c|}{\cellcolor[HTML]{C0C0C0}0.99} & 0.95$\pm$0.02 & 0.97$\pm$0.01 & 0.95$\pm$0.02 & \multicolumn{1}{c|}{\cellcolor[HTML]{C0C0C0}0.99} & 0.95$\pm$0.02 & 0.96$\pm$0.01 & 0.95$\pm$0.02 & \multicolumn{1}{c|}{\cellcolor[HTML]{C0C0C0}0.99} & 0.94$\pm$0.02 & \cellcolor[HTML]{C0C0C0}0.96$\pm$0.01 & 0.95$\pm$0.02 & \multicolumn{1}{c|}{0.91} & 0.88$\pm$0.05 & \cellcolor[HTML]{C0C0C0}0.90$\pm$0.02 & 0.90$\pm$0.04 & 0.86 \\
Elastic & 0.93$\pm$0.02 & 0.97$\pm$0.01 & 0.94$\pm$0.02 & \multicolumn{1}{c|}{\cellcolor[HTML]{C0C0C0}0.99} & 0.93$\pm$0.03 & 0.96$\pm$0.01 & 0.94$\pm$0.02 & \multicolumn{1}{c|}{\cellcolor[HTML]{C0C0C0}0.98} & 0.92$\pm$0.03 & \cellcolor[HTML]{FFFFC7}0.92$\pm$0.02 & 0.93$\pm$0.02 & \multicolumn{1}{c|}{\cellcolor[HTML]{C0C0C0}0.99} & 0.91$\pm$0.03 & 0.92$\pm$0.02 & 0.92$\pm$0.03 & \multicolumn{1}{c|}{\cellcolor[HTML]{C0C0C0}0.99} & 0.82$\pm$0.09 & 0.92$\pm$0.02 & 0.85$\pm$0.06 & \cellcolor[HTML]{C0C0C0}0.97 \\
Pixelate & 0.93$\pm$0.03 & 0.97$\pm$0.01 & 0.94$\pm$0.02 & \multicolumn{1}{c|}{\cellcolor[HTML]{C0C0C0}0.99} & 0.91$\pm$0.03 & 0.92$\pm$0.02 & 0.92$\pm$0.02 & \multicolumn{1}{c|}{\cellcolor[HTML]{C0C0C0}0.99} & 0.90$\pm$0.04 & 0.92$\pm$0.02 & 0.92$\pm$0.03 & \multicolumn{1}{c|}{\cellcolor[HTML]{C0C0C0}0.98} & 0.79$\pm$0.07 & \cellcolor[HTML]{FFFFC7}0.79$\pm$0.05 & 0.82$\pm$0.06 & \multicolumn{1}{c|}{\cellcolor[HTML]{C0C0C0}0.87} & 0.93$\pm$0.04 & \cellcolor[HTML]{FFFFC7}0.88$\pm$0.05 & \cellcolor[HTML]{C0C0C0}0.94$\pm$0.02 & 0.72 \\
JPEG & 0.88$\pm$0.05 & 0.93$\pm$0.02 & 0.90$\pm$0.03 & \multicolumn{1}{c|}{\cellcolor[HTML]{C0C0C0}0.97} & 0.83$\pm$0.08 & 0.88$\pm$0.03 & 0.85$\pm$0.05 & \multicolumn{1}{c|}{\cellcolor[HTML]{C0C0C0}0.99} & 0.79$\pm$0.10 & 0.90$\pm$0.03 & 0.82$\pm$0.07 & \multicolumn{1}{c|}{\cellcolor[HTML]{C0C0C0}0.98} & 0.75$\pm$0.10 & 0.88$\pm$0.03 & 0.79$\pm$0.08 & \multicolumn{1}{c|}{\cellcolor[HTML]{C0C0C0}0.99} & 0.70$\pm$0.12 & 0.86$\pm$0.04 & 0.74$\pm$0.09 & \cellcolor[HTML]{C0C0C0}0.98 \\ \hline
\end{tabular}
}
\end{table*}

\begin{table*}[ht]
\caption{Kendall's $\tau$ of ranking results based on MNIST-C and CIFAR-10-C. For Random, SD, and CES, we compute the average and standard deviation over all labeling budgets and 50-time experiments. The best performance is highlighted in gray. The values highlighted in yellow are where CES or random outperform SDS. The higher the better.}
\label{tab:corrup-kendall}
\resizebox{\textwidth}{!}{
\begin{tabular}{lcccccccccccccccccccc}
\hline
 & \multicolumn{4}{c|}{\textbf{Severity=1}} & \multicolumn{4}{c|}{\textbf{Severity=2}} & \multicolumn{4}{c|}{\textbf{Severity=3}} & \multicolumn{4}{c|}{\textbf{Severity=4}} & \multicolumn{4}{c}{\textbf{Severity=5}} \\ \cline{2-21} 
\multirow{-2}{*}{\textbf{Corruption Type}} & \textbf{Random} & \textbf{SDS} & \textbf{CES} & \multicolumn{1}{c|}{\textbf{Our}} & \textbf{Random} & \textbf{SDS} & \textbf{CES} & \multicolumn{1}{c|}{\textbf{Our}} & \textbf{Random} & \textbf{SDS} & \textbf{CES} & \multicolumn{1}{c|}{\textbf{Our}} & \textbf{Random} & \textbf{SDS} & \textbf{CES} & \multicolumn{1}{c|}{\textbf{Our}} & \textbf{Random} & \textbf{SDS} & \textbf{CES} & \textbf{Our} \\ \hline
\multicolumn{21}{c}{\textbf{MNIST-C}} \\ \hline
Gaussian Noise & 0.68$\pm$0.08 & 0.79$\pm$0.05 & 0.71$\pm$0.07 & \multicolumn{1}{c|}{\cellcolor[HTML]{C0C0C0}0.82} & 0.74$\pm$0.07 & 0.82$\pm$0.04 & 0.77$\pm$0.06 & \multicolumn{1}{c|}{\cellcolor[HTML]{C0C0C0} 0.86} & 0.80$\pm$0.05 & 0.83$\pm$0.04 & 0.82$\pm$0.04 & \multicolumn{1}{c|}{\cellcolor[HTML]{C0C0C0}0.87} & 0.84$\pm$0.04 & 0.86$\pm$0.03 & 0.86$\pm$0.03 & \multicolumn{1}{c|}{\cellcolor[HTML]{C0C0C0} 0.91} & 0.91$\pm$0.02 &
\cellcolor[HTML]{FFFFC7} 0.91$\pm$0.02 & 0.92$\pm$0.02 & \multicolumn{1}{r}{\cellcolor[HTML]{C0C0C0}{ 0.93}} \\
Shot Noise & 0.68$\pm$0.09 & 0.75$\pm$0.07 & 0.72$\pm$0.07 & \multicolumn{1}{c|}{\cellcolor[HTML]{C0C0C0} 0.77} & 0.67$\pm$0.09 & 0.75$\pm$0.07 & 0.71$\pm$0.07 & \multicolumn{1}{c|}{\cellcolor[HTML]{C0C0C0} 0.78} & 0.67$\pm$0.10 & \cellcolor[HTML]{C0C0C0}{ 0.75$\pm$0.07} & { 0.72$\pm$0.07} & \multicolumn{1}{c|}{0.74} & { 0.63$\pm$0.09} & { 0.70$\pm$0.06} & { 0.69$\pm$0.07} & \multicolumn{1}{c|}{\cellcolor[HTML]{C0C0C0}{ 0.76}} & { 0.57$\pm$0.09} & \cellcolor[HTML]{FFFFC7}{ 0.61$\pm$0.06} & { 0.71$\pm$0.07} & \multicolumn{1}{r}{\cellcolor[HTML]{C0C0C0}{ 0.87}} \\
Impulse Noise & { 0.58$\pm$0.08} & \cellcolor[HTML]{FFFFC7}{ 0.64$\pm$0.06} & { 0.77$\pm$0.06} & \multicolumn{1}{c|}{\cellcolor[HTML]{C0C0C0}{ 0.86}} & { 0.48$\pm$0.08} & \cellcolor[HTML]{FFFFC7}{ 0.52$\pm$0.06} & { 0.80$\pm$0.05} & \multicolumn{1}{c|}{\cellcolor[HTML]{C0C0C0}{ 0.85}} & { 0.35$\pm$0.08} & \cellcolor[HTML]{FFFFC7}{ 0.39$\pm$0.06} & { 0.83$\pm$0.04} & \multicolumn{1}{c|}{\cellcolor[HTML]{C0C0C0}{ 0.89}} & { 0.16$\pm$0.08} & \cellcolor[HTML]{FFFFC7}{ 0.20$\pm$0.07} & { 0.89$\pm$0.03} & \multicolumn{1}{c|}{\cellcolor[HTML]{C0C0C0}{ 0.89}} & { 0.03$\pm$0.08} & \cellcolor[HTML]{FFFFC7}{ 0.07$\pm$0.07} & { 0.90$\pm$0.02} & \multicolumn{1}{r}{\cellcolor[HTML]{C0C0C0}{ 0.92}}
\\
Defocus Blur & { 0.80$\pm$0.05} & \cellcolor[HTML]{FFFFC7}{ 0.79$\pm$0.16} & { 0.82$\pm$0.04} & \multicolumn{1}{c|}{\cellcolor[HTML]{C0C0C0}{ 0.92}} & { 0.83$\pm$0.04} & \cellcolor[HTML]{FFFFC7}{ -0.02$\pm$0.06} & { 0.84$\pm$0.03} & \multicolumn{1}{c|}{\cellcolor[HTML]{C0C0C0}{ 0.88}} & { 0.77$\pm$0.06} & \cellcolor[HTML]{FFFFC7}{ -0.18$\pm$0.06} & \cellcolor[HTML]{C0C0C0}0.79$\pm$0.05 & \multicolumn{1}{c|}{0.02} & { 0.61$\pm$0.10} & \cellcolor[HTML]{FFFFC7}{ -0.13$\pm$0.06} & \cellcolor[HTML]{C0C0C0}0.65$\pm$0.08 & \multicolumn{1}{c|}{-0.07} & { 0.56$\pm$0.13} & \cellcolor[HTML]{FFFFC7}{ 0.04$\pm$0.06} & \cellcolor[HTML]{C0C0C0}0.61$\pm$0.10 & \multicolumn{1}{r}{0}
\\
Frosted Glass Blur & { 0.27$\pm$0.09} & \cellcolor[HTML]{FFFFC7}{ 0.29$\pm$0.06} & { 0.80$\pm$0.05} & \multicolumn{1}{c|}{\cellcolor[HTML]{C0C0C0}{ 0.89}} & { 0.17$\pm$0.08} & \cellcolor[HTML]{FFFFC7}{ 0.19$\pm$0.07} & { 0.81$\pm$0.04} & \multicolumn{1}{c|}{\cellcolor[HTML]{C0C0C0}{ 0.89}} & -0.37$\pm$0.08 & \cellcolor[HTML]{FFFFC7}{-0.36$\pm$0.07} & 0.85$\pm$0.04 & \multicolumn{1}{c|}{\cellcolor[HTML]{C0C0C0}0.81} & -0.39$\pm$0.08 & \cellcolor[HTML]{FFFFC7}{-0.39$\pm$0.06} & 0.84$\pm$0.04 & \multicolumn{1}{c|}{\cellcolor[HTML]{C0C0C0}0.71} & -0.24$\pm$0.08 & \cellcolor[HTML]{FFFFC7}{-0.25$\pm$0.06} & 0.80$\pm$0.04 & \multicolumn{1}{r}{\cellcolor[HTML]{C0C0C0}0.05}
\\
Motion Blur & { 0.6$\pm$0.08} & \cellcolor[HTML]{FFFFC7}{ 0.65$\pm$0.06} & { 0.81$\pm$0.04} & \multicolumn{1}{c|}{\cellcolor[HTML]{C0C0C0}{ 0.83}} & { 0.58$\pm$0.08} & \cellcolor[HTML]{FFFFC7}{ 0.61$\pm$0.06} & { 0.86$\pm$0.03} & \multicolumn{1}{c|}{\cellcolor[HTML]{C0C0C0}{ 0.76}} & 0.47$\pm$0.07 & \cellcolor[HTML]{FFFFC7}{0.49$\pm$0.06} & 0.85$\pm$0.03 & \multicolumn{1}{c|}{\cellcolor[HTML]{C0C0C0}0.73} & 0.38$\pm$0.07 & \cellcolor[HTML]{FFFFC7}{0.40$\pm$0.06} & \cellcolor[HTML]{C0C0C0}0.83$\pm$0.04 & \multicolumn{1}{c|}{0.36} & 0.36$\pm$0.07 & \cellcolor[HTML]{FFFFC7}{0.39$\pm$0.06} & \cellcolor[HTML]{C0C0C0}0.81$\pm$0.04 & \multicolumn{1}{r}{-0.13}
\\
Zoom Blur & { 0.67$\pm$0.08} & \cellcolor[HTML]{FFFFC7}{ 0.73$\pm$0.06} & { 0.75$\pm$0.06} & \multicolumn{1}{c|}{\cellcolor[HTML]{C0C0C0}{ 0.81}} & { 0.66$\pm$0.09} & \cellcolor[HTML]{FFFFC7}{ 0.73$\pm$0.06} & { 0.76$\pm$0.06} & \multicolumn{1}{c|}{\cellcolor[HTML]{C0C0C0}0.80} & 0.66$\pm$0.08 & \cellcolor[HTML]{FFFFC7}{0.71$\pm$0.06} & 0.76$\pm$0.06 & \multicolumn{1}{c|}{\cellcolor[HTML]{C0C0C0}0.82} & 0.65$\pm$0.08 & \cellcolor[HTML]{FFFFC7}{0.70$\pm$0.06} & 0.78$\pm$0.05 & \multicolumn{1}{c|}{\cellcolor[HTML]{C0C0C0}0.81} & 0.63$\pm$0.08 & \cellcolor[HTML]{FFFFC7}{0.68$\pm$0.06} & 0.78$\pm$0.06 & \multicolumn{1}{r}{\cellcolor[HTML]{C0C0C0}{ 0.84}}
\\
Snow & { 0.47$\pm$0.08} & \cellcolor[HTML]{FFFFC7}{ 0.51$\pm$0.06} & { 0.79$\pm$0.05} & \multicolumn{1}{c|}{\cellcolor[HTML]{C0C0C0}{ 0.84}} & { 0.41$\pm$0.08} & \cellcolor[HTML]{FFFFC7}{ 0.42$\pm$0.06} & { 0.87$\pm$0.03} & \multicolumn{1}{c|}{\cellcolor[HTML]{C0C0C0}0.95} & 0.43$\pm$0.09 & \cellcolor[HTML]{FFFFC7}{0.43$\pm$0.07} & 0.88$\pm$0.03 & \multicolumn{1}{c|}{\cellcolor[HTML]{C0C0C0}0.93} & 0.43$\pm$0.08 & \cellcolor[HTML]{FFFFC7}{0.44$\pm$0.06} & 0.88$\pm$0.03 & \multicolumn{1}{c|}{\cellcolor[HTML]{C0C0C0}0.91} & 0.41$\pm$0.09 & \cellcolor[HTML]{FFFFC7}{0.41$\pm$0.07} & 0.90$\pm$0.03 & \multicolumn{1}{r}{\cellcolor[HTML]{C0C0C0}{ 0.94}}
\\
Frost & { 0.38$\pm$0.08} & \cellcolor[HTML]{FFFFC7}{ 0.38$\pm$0.06} & { 0.88$\pm$0.03} & \multicolumn{1}{c|}{\cellcolor[HTML]{C0C0C0}{ 0.96}} & { 0.37$\pm$0.09} & \cellcolor[HTML]{FFFFC7}{ 0.36$\pm$0.07} & { 0.92$\pm$0.02} & \multicolumn{1}{c|}{\cellcolor[HTML]{C0C0C0}0.94} & 0.36$\pm$0.09 & \cellcolor[HTML]{FFFFC7}{0.35$\pm$0.07} & 0.92$\pm$0.02 & \multicolumn{1}{c|}{\cellcolor[HTML]{C0C0C0}0.93} & 0.36$\pm$0.09 & \cellcolor[HTML]{FFFFC7}0.35$\pm$0.07 & 0.93$\pm$0.02 & \multicolumn{1}{c|}{\cellcolor[HTML]{C0C0C0}0.94} & 0.36$\pm$0.09 & \cellcolor[HTML]{FFFFC7}0.34$\pm$0.06 & 0.92$\pm$0.02 & \multicolumn{1}{r}{\cellcolor[HTML]{C0C0C0}0.93}
\\
Fog & { 0.35$\pm$0.08} & \cellcolor[HTML]{FFFFC7}{ 0.36$\pm$0.07} & { 0.93$\pm$0.02} & \multicolumn{1}{c|}{\cellcolor[HTML]{C0C0C0}{ 0.94}} & { 0.35$\pm$0.09} & \cellcolor[HTML]{FFFFC7}{ 0.34$\pm$0.07} & { 0.93$\pm$0.02} & \multicolumn{1}{c|}{\cellcolor[HTML]{C0C0C0}0.92} & 0.32$\pm$0.08 & \cellcolor[HTML]{FFFFC7}{0.30$\pm$0.06} & 0.91$\pm$0.02 & \multicolumn{1}{c|}{\cellcolor[HTML]{C0C0C0}0.83} & 0.32$\pm$0.09 & \cellcolor[HTML]{FFFFC7}{0.31$\pm$0.06} & 0.90$\pm$0.02 & \multicolumn{1}{c|}{\cellcolor[HTML]{C0C0C0}0.77} & 0.27$\pm$0.08 & \cellcolor[HTML]{FFFFC7}0.26$\pm$0.06 & \cellcolor[HTML]{C0C0C0}0.88$\pm$0.03 & -0.07
\\
Brightness & { 0.77$\pm$0.06} & \cellcolor[HTML]{FFFFC7}{ 0.53$\pm$0.06} & { 0.79$\pm$0.05} & \multicolumn{1}{c|}{\cellcolor[HTML]{C0C0C0}{ 0.85}} & { 0.82$\pm$0.04} & \cellcolor[HTML]{FFFFC7}{ 0.40$\pm$0.06} & { 0.84$\pm$0.04} & \multicolumn{1}{c|}{\cellcolor[HTML]{C0C0C0}0.91} & 0.89$\pm$0.03 & \cellcolor[HTML]{FFFFC7}0.39$\pm$0.06 & 0.90$\pm$0.02 & \multicolumn{1}{c|}{\cellcolor[HTML]{C0C0C0}0.95} & 0.91$\pm$0.02 & \cellcolor[HTML]{FFFFC7}0.36$\pm$0.07 & 0.92$\pm$0.02 & \multicolumn{1}{c|}{\cellcolor[HTML]{C0C0C0}0.97} & 0.91$\pm$0.03 & \cellcolor[HTML]{FFFFC7}0.35$\pm$0.06 & \cellcolor[HTML]{C0C0C0}0.92$\pm$0.02 & 0.89
\\
Contrast & { 0.82$\pm$0.05} & \cellcolor[HTML]{FFFFC7}{ 0.47$\pm$0.06} & { 0.84$\pm$0.04} & \multicolumn{1}{c|}{\cellcolor[HTML]{C0C0C0}{ 0.92}} & { 0.86$\pm$0.04} & \cellcolor[HTML]{FFFFC7}{ 0.43$\pm$0.06} & { 0.87$\pm$0.03} & \multicolumn{1}{c|}{\cellcolor[HTML]{C0C0C0}0.93} & 0.91$\pm$0.02 & \cellcolor[HTML]{FFFFC7}0.34$\pm$0.07 & 0.92$\pm$0.02 & \multicolumn{1}{c|}{\cellcolor[HTML]{C0C0C0}0.97} & 0.88$\pm$0.04 & \cellcolor[HTML]{FFFFC7}0.21$\pm$0.06 & \cellcolor[HTML]{C0C0C0}0.89$\pm$0.03 & \multicolumn{1}{c|}{0.84} & 0.76$\pm$0.08 & \cellcolor[HTML]{FFFFC7}0.10$\pm$0.05 & \cellcolor[HTML]{C0C0C0}0.78$\pm$0.07 & 0.12
\\
Elastic & { 0.84$\pm$0.04} & \cellcolor[HTML]{FFFFC7}{ 0.70$\pm$0.07} & \cellcolor[HTML]{C0C0C0}0.85$\pm$0.04 & \multicolumn{1}{c|}{0.70} & { 0.22$\pm$0.15} & \cellcolor[HTML]{C0C0C0}0.31$\pm$0.06 & { 0.28$\pm$0.15} & \multicolumn{1}{c|}{-0.03} & 0.74$\pm$0.06 & \cellcolor[HTML]{FFFFC7}0.42$\pm$0.07 & 0.76$\pm$0.06 & \multicolumn{1}{c|}{\cellcolor[HTML]{C0C0C0}0.87} & 0.82$\pm$0.05 & \cellcolor[HTML]{FFFFC7}-0.18$\pm$0.06 & 0.83$\pm$0.04 & \multicolumn{1}{c|}{\cellcolor[HTML]{C0C0C0}0.87} & 0.87$\pm$0.03 & \cellcolor[HTML]{FFFFC7}-0.48$\pm$0.06 & \cellcolor[HTML]{C0C0C0}0.89$\pm$0.03 & -0.31
\\
Pixelate & { 0.65$\pm$0.08} & { 0.72$\pm$0.06} & { 0.70$\pm$0.07} & \multicolumn{1}{c|}{\cellcolor[HTML]{C0C0C0}{ 0.73}} & { 0.64$\pm$0.08} & \cellcolor[HTML]{FFFFC7}{ 0.70$\pm$0.06} & { 0.72$\pm$0.06} & \multicolumn{1}{c|}{\cellcolor[HTML]{C0C0C0}0.79} & 0.46$\pm$0.09 & \cellcolor[HTML]{FFFFC7}0.47$\pm$0.07 & 0.76$\pm$0.05 & \multicolumn{1}{c|}{\cellcolor[HTML]{C0C0C0}0.87} & 0.11$\pm$0.10 & \cellcolor[HTML]{FFFFC7}0.08$\pm$0.07 & 0.74$\pm$0.06 & \multicolumn{1}{c|}{\cellcolor[HTML]{C0C0C0}0.89} & -0.22$\pm$0.09 & \cellcolor[HTML]{FFFFC7}-0.24$\pm$0.07 & 0.77$\pm$0.05 & \cellcolor[HTML]{C0C0C0}0.72
\\
JPEG & { 0.67$\pm$0.09} & { 0.75$\pm$0.07} & { 0.71$\pm$0.07} & \multicolumn{1}{c|}{\cellcolor[HTML]{C0C0C0}{ 0.81}} & { 0.67$\pm$0.09} & { 0.74$\pm$0.06} & { 0.69$\pm$0.07} & \multicolumn{1}{c|}{\cellcolor[HTML]{C0C0C0}0.76} & 0.66$\pm$0.08 & 0.74$\pm$0.06 & 0.71$\pm$0.07 & \multicolumn{1}{c|}{\cellcolor[HTML]{C0C0C0}0.78} & 0.66$\pm$0.08 & 0.73$\pm$0.06 & 0.71$\pm$0.07 & \multicolumn{1}{c|}{\cellcolor[HTML]{C0C0C0}0.79} & 0.65$\pm$0.08 & 0.72$\pm$0.06 & 0.69$\pm$0.07 & \cellcolor[HTML]{C0C0C0}0.78 \\ \hline
\multicolumn{21}{c}{\textbf{CIFAR-10-C}} \\ \hline
Gaussian Noise & 0.68$\pm$0.08 & 0.72$\pm$0.04 & 0.71$\pm$0.06 & \multicolumn{1}{c|}{\cellcolor[HTML]{C0C0C0}0.92} & 0.74$\pm$0.07 & \cellcolor[HTML]{C0C0C0}0.82$\pm$0.03 & 0.77$\pm$0.05 & \multicolumn{1}{c|}{0.78} & 0.81$\pm$0.04 & \cellcolor[HTML]{FFFFC7}0.82$\pm$0.03 & 0.83$\pm$0.03 & \multicolumn{1}{c|}{\cellcolor[HTML]{C0C0C0}0.84} & 0.82$\pm$0.04 & \cellcolor[HTML]{FFFFC7}0.80$\pm$0.03 & \cellcolor[HTML]{C0C0C0}0.84$\pm$0.03 & \multicolumn{1}{c|}{0.80} & 0.83$\pm$0.04 & \cellcolor[HTML]{FFFFC7}0.80$\pm$0.03 & \cellcolor[HTML]{C0C0C0}0.84$\pm$0.03 & 0.78
\\
Shot Noise & 0.75$\pm$0.06 & 0.80$\pm$0.03 & 0.77$\pm$0.05 & \multicolumn{1}{c|}{\cellcolor[HTML]{C0C0C0}0.94} & 0.65$\pm$0.08 & 0.69$\pm$0.05 & 0.69$\pm$0.07 & \multicolumn{1}{c|}{\cellcolor[HTML]{C0C0C0}0.87} & 0.77$\pm$0.06 & \cellcolor[HTML]{C0C0C0}0.82$\pm$0.03 & 0.80$\pm$0.04 & \multicolumn{1}{c|}{0.80} & 0.80$\pm$0.04 & \cellcolor[HTML]{FFFFC7}0.79$\pm$0.03 & 0.82$\pm$0.04 & \multicolumn{1}{c|}{\cellcolor[HTML]{C0C0C0}0.83} & 0.81$\pm$0.04 & \cellcolor[HTML]{FFFFC7}0.76$\pm$0.03 & \cellcolor[HTML]{C0C0C0}0.83$\pm$0.04 & 0.74
\\
Impulse Noise & 0.74$\pm$0.06 & 0.78$\pm$0.03 & 0.76$\pm$0.05 & \multicolumn{1}{c|}{\cellcolor[HTML]{C0C0C0}0.93} & 0.69$\pm$0.07 & 0.76$\pm$0.06 & 0.71$\pm$0.06 & \multicolumn{1}{c|}{\cellcolor[HTML]{C0C0C0}0.88} & 0.72$\pm$0.07 & \cellcolor[HTML]{FFFFC7}0.72$\pm$0.04 & 0.74$\pm$0.05 & \multicolumn{1}{c|}{\cellcolor[HTML]{C0C0C0}0.78} & 0.79$\pm$0.05 & \cellcolor[HTML]{FFFFC7}0.69$\pm$0.03 & \cellcolor[HTML]{C0C0C0}0.80$\pm$0.04 & \multicolumn{1}{c|}{0.66} & 0.81$\pm$0.04 & \cellcolor[HTML]{FFFFC7}0.77$\pm$0.03 & \cellcolor[HTML]{C0C0C0}0.83$\pm$0.03 & 0.67
\\
Defocus Blur & 0.82$\pm$0.04 & 0.87$\pm$0.02 & 0.83$\pm$0.04 & \multicolumn{1}{c|}{\cellcolor[HTML]{C0C0C0}0.97} & 0.81$\pm$0.04 & 0.87$\pm$0.03 & 0.82$\pm$0.04 & \multicolumn{1}{c|}{\cellcolor[HTML]{C0C0C0}0.93} & 0.79$\pm$0.04 & 0.81$\pm$0.03 & 0.80$\pm$0.04 & \multicolumn{1}{c|}{\cellcolor[HTML]{C0C0C0}0.93} & 0.80$\pm$0.04 & 0.82$\pm$0.04 & 0.82$\pm$0.04 & \multicolumn{1}{c|}{\cellcolor[HTML]{C0C0C0}0.93} & 0.82$\pm$0.04 & 0.86$\pm$0.03 & 0.83$\pm$0.03 & \cellcolor[HTML]{C0C0C0}0.90
\\
Frosted Glass Blur & 0.69$\pm$0.06 & 0.76$\pm$0.04 & 0.71$\pm$0.06 & \multicolumn{1}{c|}{\cellcolor[HTML]{C0C0C0}0.88} & 0.68$\pm$0.06 & 0.81$\pm$0.04 & 0.71$\pm$0.06 & \multicolumn{1}{c|}{\cellcolor[HTML]{C0C0C0}0.88} & 0.65$\pm$0.07 & 0.77$\pm$0.05 & 0.68$\pm$0.06 & \multicolumn{1}{c|}{\cellcolor[HTML]{C0C0C0}0.89} & 0.75$\pm$0.07 & \cellcolor[HTML]{C0C0C0}0.84$\pm$0.03 & 0.79$\pm$0.05 & \multicolumn{1}{c|}{0.81} & 0.72$\pm$0.08 & \cellcolor[HTML]{C0C0C0}0.86$\pm$0.03 & 0.77$\pm$0.05 & 0.80
\\
Motion Blur & 0.77$\pm$0.05 & 0.84$\pm$0.03 & 0.79$\pm$0.05 & \multicolumn{1}{c|}{\cellcolor[HTML]{C0C0C0}0.89} & 0.78$\pm$0.05 & \cellcolor[HTML]{FFFFC7}{0.78$\pm$0.03} & 0.79$\pm$0.04 & \multicolumn{1}{c|}{\cellcolor[HTML]{C0C0C0}0.94} & 0.79$\pm$0.04 & 0.81$\pm$0.04 & 0.81$\pm$0.04 & \multicolumn{1}{c|}{\cellcolor[HTML]{C0C0C0}0.92} & 0.80$\pm$0.04 & \cellcolor[HTML]{FFFFC7}0.81$\pm$0.04 & 0.82$\pm$0.04 & \multicolumn{1}{c|}{\cellcolor[HTML]{C0C0C0}0.92} & 0.81$\pm$0.04 & 0.84$\pm$0.03 & 0.82$\pm$0.04 & \cellcolor[HTML]{C0C0C0}0.94
\\
Zoom Blur & 0.78$\pm$0.04 & \cellcolor[HTML]{FFFFC7}{0.76$\pm$0.03} & 0.79$\pm$0.04 & \multicolumn{1}{c|}{\cellcolor[HTML]{C0C0C0}0.96} & 0.79$\pm$0.04 & \cellcolor[HTML]{FFFFC7}{0.80$\pm$0.03} & 0.81$\pm$0.04 & \multicolumn{1}{c|}{\cellcolor[HTML]{C0C0C0}0.97} & 0.80$\pm$0.04 & \cellcolor[HTML]{FFFFC7}0.81$\pm$0.03 & 0.82$\pm$0.04 & \multicolumn{1}{c|}{\cellcolor[HTML]{C0C0C0}0.96} & 0.81$\pm$0.04 & 0.83$\pm$0.03 & 0.82$\pm$0.04 & \multicolumn{1}{c|}{\cellcolor[HTML]{C0C0C0}0.95} & 0.81$\pm$0.04 & 0.85$\pm$0.03 & 0.82$\pm$0.04 & \cellcolor[HTML]{C0C0C0}0.92
\\
Snow & 0.79$\pm$0.05 & 0.81$\pm$0.03 & 0.80$\pm$0.04 & \multicolumn{1}{c|}{\cellcolor[HTML]{C0C0C0}0.94} & 0.78$\pm$0.05 & 0.80$\pm$0.04 & 0.80$\pm$0.04 & \multicolumn{1}{c|}{\cellcolor[HTML]{C0C0C0}0.95} & 0.76$\pm$0.05 & 0.78$\pm$0.04 & 0.78$\pm$0.04 & \multicolumn{1}{c|}{\cellcolor[HTML]{C0C0C0}0.93} & 0.76$\pm$0.05 & 0.81$\pm$0.04 & 0.78$\pm$0.04 & \multicolumn{1}{c|}{\cellcolor[HTML]{C0C0C0}0.94} & 0.76$\pm$0.05 & \cellcolor[HTML]{FFFFC7}0.75$\pm$0.04 & 0.78$\pm$0.05 & \cellcolor[HTML]{C0C0C0}0.94
\\
Frost & 0.79$\pm$0.05 & 0.86$\pm$0.02 & 0.81$\pm$0.04 & \multicolumn{1}{c|}{\cellcolor[HTML]{C0C0C0}0.93} & 0.79$\pm$0.05 & 0.85$\pm$0.03 & 0.80$\pm$0.04 & \multicolumn{1}{c|}{\cellcolor[HTML]{C0C0C0}0.93} & 0.78$\pm$0.05 & \cellcolor[HTML]{FFFFC7}0.77$\pm$0.03 & 0.80$\pm$0.04 & \multicolumn{1}{c|}{\cellcolor[HTML]{C0C0C0}0.91} & 0.77$\pm$0.06 & \cellcolor[HTML]{FFFFC7}0.76$\pm$0.03 & 0.79$\pm$0.05 & \multicolumn{1}{c|}{\cellcolor[HTML]{C0C0C0}0.94} & 0.72$\pm$0.06 & \cellcolor[HTML]{FFFFC7}0.72$\pm$0.04 & 0.75$\pm$0.05 & \cellcolor[HTML]{C0C0C0}0.83
\\
Fog & 0.82$\pm$0.04 & 0.88$\pm$0.02 & 0.83$\pm$0.04 & \multicolumn{1}{c|}{\cellcolor[HTML]{C0C0C0}0.97} & 0.84$\pm$0.04 & 0.88$\pm$0.02 & 0.85$\pm$0.03 & \multicolumn{1}{c|}{\cellcolor[HTML]{C0C0C0}0.97} & 0.84$\pm$0.04 & 0.89$\pm$0.03 & 0.85$\pm$0.03 & \multicolumn{1}{c|}{\cellcolor[HTML]{C0C0C0}0.96} & 0.85$\pm$0.03 & 0.89$\pm$0.02 & 0.87$\pm$0.03 & \multicolumn{1}{c|}{\cellcolor[HTML]{C0C0C0}0.96} & 0.84$\pm$0.04 & 0.86$\pm$0.02 & 0.86$\pm$0.03 & \cellcolor[HTML]{C0C0C0}0.86
\\
Brightness & 0.82$\pm$0.04 & 0.87$\pm$0.03 & 0.83$\pm$0.03 & \multicolumn{1}{c|}{\cellcolor[HTML]{C0C0C0}0.96} & 0.82$\pm$0.04 & 0.87$\pm$0.03 & 0.83$\pm$0.04 & \multicolumn{1}{c|}{\cellcolor[HTML]{C0C0C0}0.94} & 0.82$\pm$0.04 & 0.87$\pm$0.03 & 0.83$\pm$0.03 & \multicolumn{1}{c|}{\cellcolor[HTML]{C0C0C0}0.94} & 0.82$\pm$0.04 & 0.88$\pm$0.02 & 0.84$\pm$0.04 & \multicolumn{1}{c|}{\cellcolor[HTML]{C0C0C0}0.94} & 0.82$\pm$0.04 & 0.88$\pm$0.02 & 0.84$\pm$0.04 & \cellcolor[HTML]{C0C0C0}0.95
\\
Contrast & 0.82$\pm$0.04 & 0.88$\pm$0.03 & 0.83$\pm$0.04 & \multicolumn{1}{c|}{\cellcolor[HTML]{C0C0C0}0.96} & 0.84$\pm$0.04 & 0.89$\pm$0.02 & 0.85$\pm$0.03 & \multicolumn{1}{c|}{\cellcolor[HTML]{C0C0C0}0.95} & 0.84$\pm$0.04 & 0.87$\pm$0.03 & 0.85$\pm$0.03 & \multicolumn{1}{c|}{\cellcolor[HTML]{C0C0C0}0.93} & 0.83$\pm$0.04 & \cellcolor[HTML]{C0C0C0}0.86$\pm$0.03 & 0.85$\pm$0.03 & \multicolumn{1}{c|}{0.78} & 0.74$\pm$0.06 & 0.73$\pm$0.03 & \cellcolor[HTML]{C0C0C0}0.78$\pm$0.05 & 0.67
\\
Elastic & 0.80$\pm$0.04 & 0.86$\pm$0.03 & 0.82$\pm$0.04 & \multicolumn{1}{c|}{\cellcolor[HTML]{C0C0C0}0.95} & 0.80$\pm$0.04 & 0.84$\pm$0.03 & 0.81$\pm$0.04 & \multicolumn{1}{c|}{\cellcolor[HTML]{C0C0C0}0.92} & 0.78$\pm$0.05 & \cellcolor[HTML]{FFFFC7}0.79$\pm$0.03 & 0.80$\pm$0.04 & \multicolumn{1}{c|}{\cellcolor[HTML]{C0C0C0}0.94} & 0.76$\pm$0.05 & 0.78$\pm$0.03 & 0.78$\pm$0.04 & \multicolumn{1}{c|}{\cellcolor[HTML]{C0C0C0}0.92} & 0.66$\pm$0.09 & 0.78$\pm$0.04 & 0.69$\pm$0.07 & \cellcolor[HTML]{C0C0C0}0.90
\\
Pixelate & 0.80$\pm$0.04 & 0.86$\pm$0.02 & 0.82$\pm$0.04 & \multicolumn{1}{c|}{\cellcolor[HTML]{C0C0C0}0.94} & 0.78$\pm$0.05 & \cellcolor[HTML]{FFFFC7}{0.78$\pm$0.03} & 0.79$\pm$0.04 & \multicolumn{1}{c|}{\cellcolor[HTML]{C0C0C0}0.95} & 0.76$\pm$0.05 & 0.79$\pm$0.04 & 0.78$\pm$0.04 & \multicolumn{1}{c|}{\cellcolor[HTML]{C0C0C0}0.91} & 0.64$\pm$0.07 & 0.62$\pm$0.05 & 0.66$\pm$0.07 & \multicolumn{1}{c|}{\cellcolor[HTML]{C0C0C0}0.72} & 0.81$\pm$0.05 & \cellcolor[HTML]{FFFFC7}0.76$\pm$0.06 & \cellcolor[HTML]{C0C0C0}0.83$\pm$0.04 & 0.61
\\
JPEG & 0.73$\pm$0.06 & 0.80$\pm$0.03 & 0.75$\pm$0.05 & \multicolumn{1}{c|}{\cellcolor[HTML]{C0C0C0}0.89} & 0.66$\pm$0.08 & 0.72$\pm$0.04 & 0.69$\pm$0.06 & \multicolumn{1}{c|}{\cellcolor[HTML]{C0C0C0}0.94} & 0.63$\pm$0.10 & 0.74$\pm$0.04 & 0.65$\pm$0.07 & \multicolumn{1}{c|}{\cellcolor[HTML]{C0C0C0}0.91} & 0.59$\pm$0.09 & 0.71$\pm$0.05 & 0.62$\pm$0.07 & \multicolumn{1}{c|}{\cellcolor[HTML]{C0C0C0}0.92} & 0.54$\pm$0.10 & 0.70$\pm$0.06 & 0.58$\pm$0.09 & \cellcolor[HTML]{C0C0C0}0.91 \\ \hline
\end{tabular}
}
\end{table*}

\begin{figure}
    \centering
    \subfigure[iWildCam-OOD]{
    \includegraphics[scale=0.29]{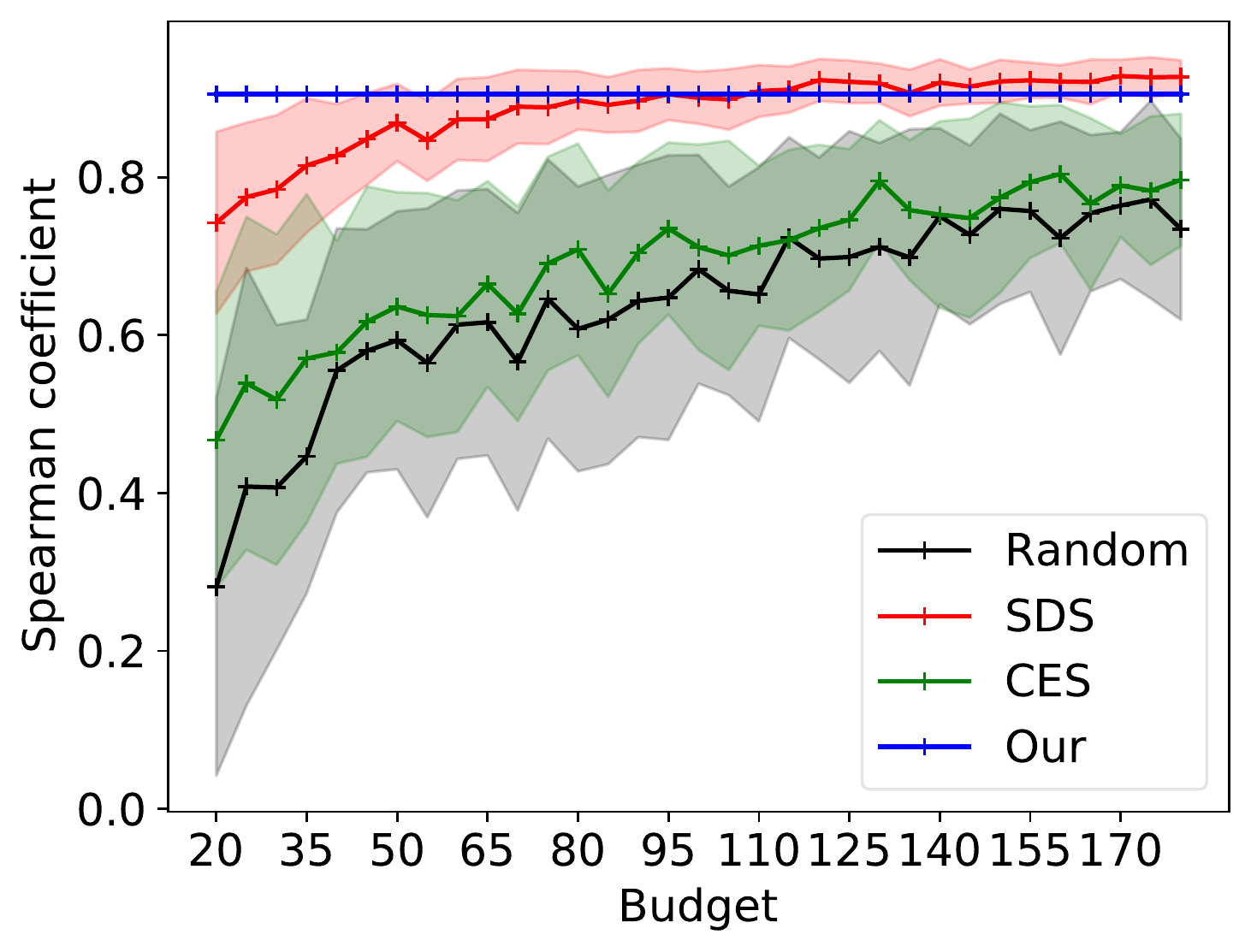}
    }
    \subfigure[Amazon-OOD]{
    \includegraphics[scale=0.29]{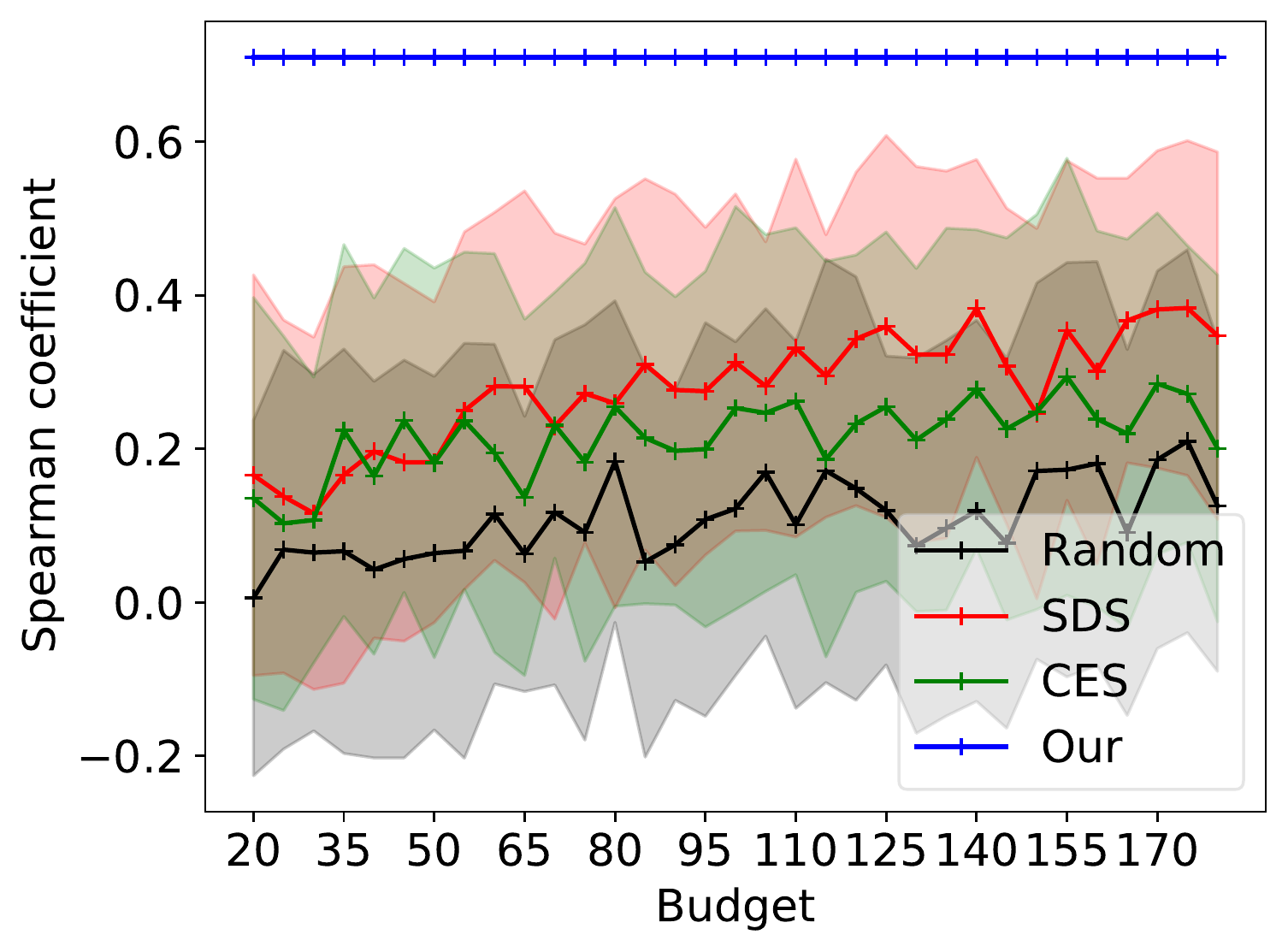}
    }
    \subfigure[Java250-OOD]{
    \includegraphics[scale=0.29]{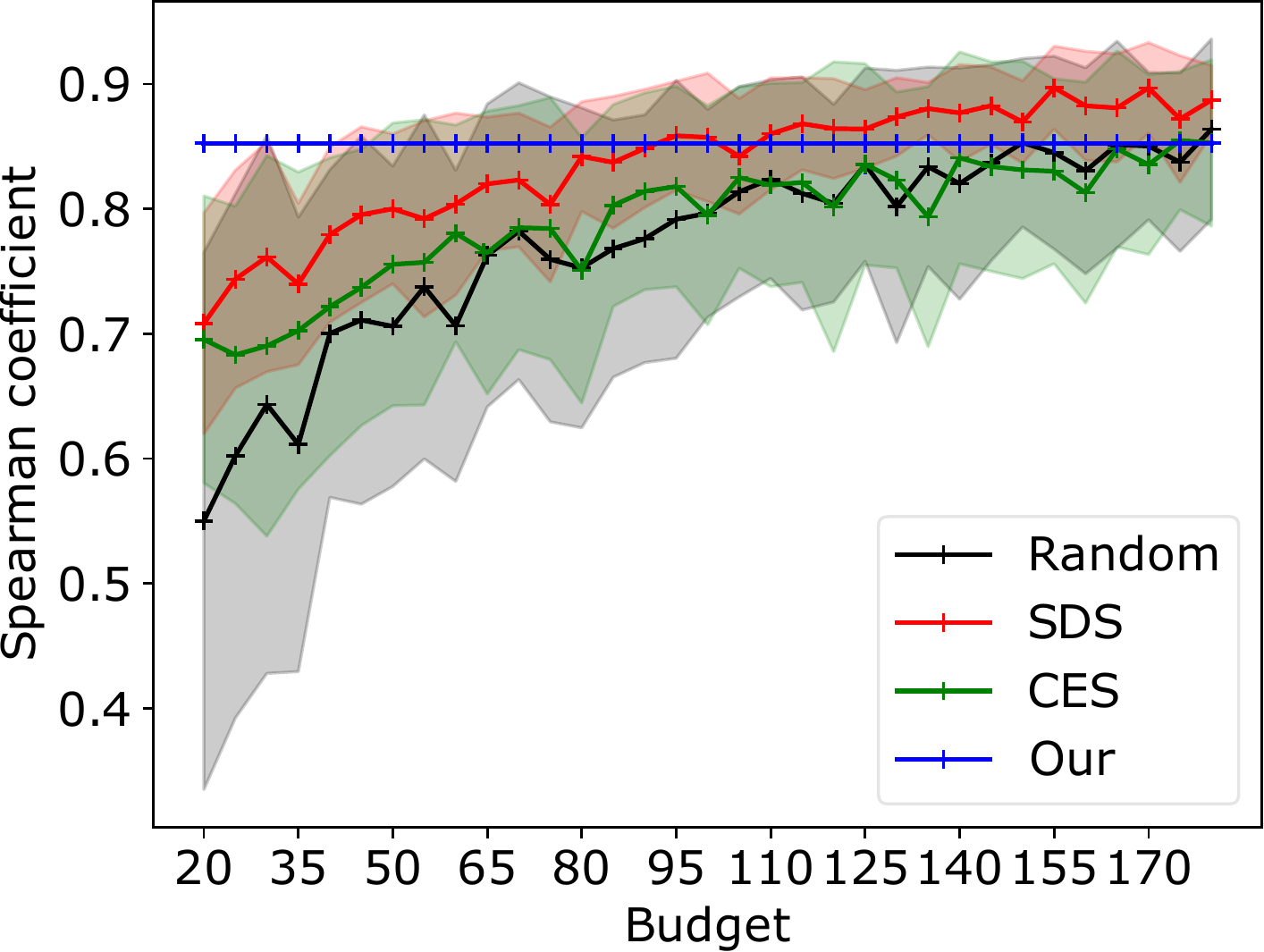}
    }
    \caption{Spearman's correlation coefficient of ranking results based on OOD test data. The higher the better. The shaded area represents the standard deviation. ``Budget'' is the number of labeled data.}
    \label{fig:ood-spearman}
\end{figure}

For the natural distribution shift, the results are shown in Figure \ref{fig:ood-spearman}. LaF can better distinguish the performance of DNNs than the baseline methods. In particular, in Amazon, LaF is significantly better by up to 0.70 based on Spearman's correlation. In addition, concerning the Jaccard similarity in Table \ref{tab:oodk}, LaF is consistently the best in identifying the top DNNs for iWildCam and Amazon.

\begin{table}[]
\caption{Jaccard similarity of ranking the top-$k$ DNNs concerning natural distribution shift. For baseline methods, we report the average results over all labeling budgets. The best performance is highlighted in gray. The higher the better.}
\label{tab:oodk}
\resizebox{0.6\columnwidth}{!}{
\begin{tabular}{llrrrr}
\hline
\textbf{Jaccard} & \textbf{Dataset} & \textbf{Random} & \textbf{SDS} & \textbf{CES} & \textbf{Our} \\ \hline
 & iWildCam & 0.66 & 0.96 & 0.68 & \cellcolor[HTML]{C0C0C0}1 \\
 & Amazon & 0.1 & 0.22 & 0.15 & \cellcolor[HTML]{C0C0C0}1 \\
\multirow{-3}{*}{$k$=1} & Java250 & 0.16 & 0.00 & \cellcolor[HTML]{C0C0C0}0.95 & 0.00\\ \hline
 & iWildCam & 0.4 & 0.63 & 0.41 & \cellcolor[HTML]{C0C0C0}1 \\
 & Amazon & 0.16 & 0.28 & 0.21 & \cellcolor[HTML]{C0C0C0}1 \\ 
 \multirow{-3}{*}{$k$=3} & Java250 & 0.27 & 0.11 & \cellcolor[HTML]{C0C0C0}0.76 & 0.00 \\ \hline
 & iWildCam & 0.44 & 0.61 & 0.47 & \cellcolor[HTML]{C0C0C0}0.67 \\
 & Amazon & 0.23 & 0.39 & 0.29 & \cellcolor[HTML]{C0C0C0}1 \\
 \multirow{-3}{*}{$k$=5} & Java250 & 0.36 & 0.28 & \cellcolor[HTML]{C0C0C0}0.78 & 0.25\\ \hline
 & iWildCam & 0.62 & \cellcolor[HTML]{C0C0C0}0.83 & 0.65 & 0.82 \\
 & Amazon & 0.41 & 0.44 & \cellcolor[HTML]{C0C0C0}0.45 & 0.43 \\ \multirow{-3}{*}{$k$=10} & Java250 & 0.61 & 0.72 & 0.67 & \cellcolor[HTML]{C0C0C0}1\\ \hline
\end{tabular}}
\end{table}

In addition, compared to the effectiveness given ID test data, the ranking by all methods is different since the performance of DNNs changes given OOD test data. However, we notice an opposite phenomenon happens. Given ID test data, LaF achieves 0.39, 0.80, and 0.96 concerning Spearman's coefficient for iWildCam, Amazon, and Java250, respectively. While given OOD test data, the results are 0.91, 0.71, and 0.85, respectively. In other words, the effectiveness improves on the OOD test data in iWildCam but degrades in Amazon and Java250. To make clear the reason behind this, we analyze the accuracy and robustness of multiple DNNs on ID and OOD test data (Table \ref{tab:data-summary}), respectively. In iWildCam, the performance difference of its 20 DNNs becomes larger on OOD test data, from 1.54\% to 11.52\%. In Amazon, the performance of all 20 DNNs degrades, e.g., from 74.84\% to 72.35\%. Besides, the performance difference in Amazon becomes smaller. Therefore, we believe that the model's ability and the performance difference among DNNs have an impact on the ranking effectiveness, which leads to the investigation in RQ3.

\vspace{1mm}
\noindent\fcolorbox{black}{answercolor}{\begin{minipage}{\columnwidth} \textbf{Answer to RQ2}: Under different distribution shifts, LaF still outstands in all ranking methods. Due to the distribution shift, DNNs' performance declines, which degrades the ranking effectiveness. Particularly, SDS is sensitive to artificial distribution shifts and fails to defeat random sampling and CES in many cases.
\end{minipage}}

\subsection{RQ3: Analysis of Impact factors}
\label{subsec:impact}
As mentioned in RQ2, by comparing the ranking effectiveness given ID and OOD test data, we raise the demand of investigating the two impact factors, the quality, and diversity of multiple DNNs. The quality refers to the model's performance given the test data and is calculated as the average accuracy or robustness over all DNNs on each dataset. For instance, in MNIST without distribution shift, the quality is the average accuracy of 30 DNNs on the ID test data, and in MNIST-C with Gaussian Noise (severity=1), the quality is the average robustness of 30 DNNs on the corresponding OOD data. The diversity indicates the performance difference among DNNs and is the standard deviation of accuracy or robustness over all DNNs on each dataset. 

\begin{figure} [h]
    \centering
    \includegraphics[scale=0.57]{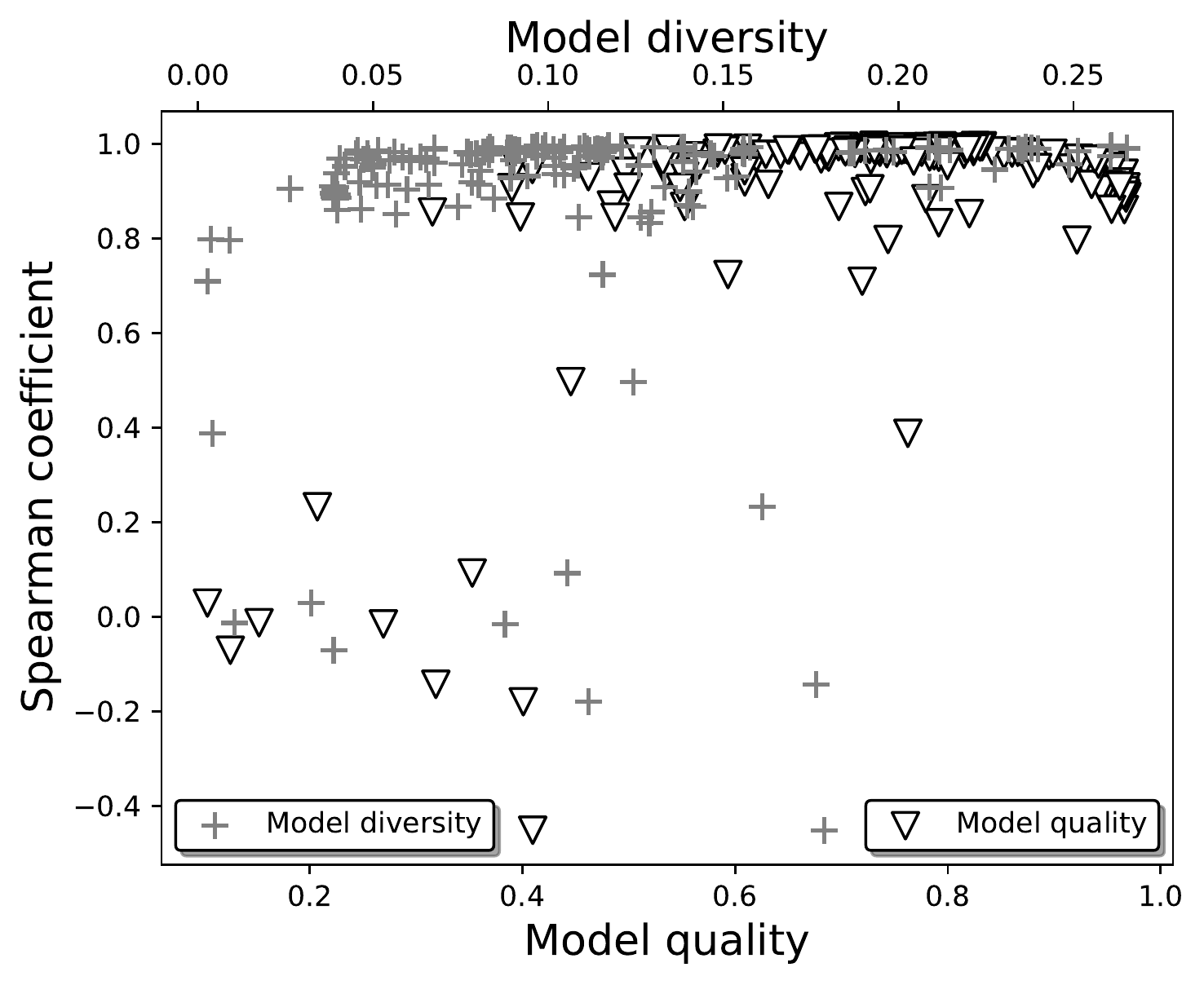}
    \caption{The impact of model quality/diversity on the ranking performance of LaF. Each point indicates the Spearman's correlation coefficient of a specific dataset and its DNNs. All 160 datasets are included.}
    \label{fig:impact}
\end{figure}

Figure \ref{fig:impact} plots the distribution of ranking performance concerning quality and diversity. Most good rankings happen with a high model quality (greater than 50\%). The reason is that in our scenario, we only have the access to the predictions of test data of multiple DNNs, which setups the initial inference of data difficulty and model specialty. Therefore, the learned Bayesian model can be more precise when the qualities of DNNs are high. Furthermore, this also explains why LaF outperforms the sampling-based methods. For example, SDS selects a few discriminative data to annotate to rank DNNs and the selection of data highly relies on the predicted labels. As a result, since the low qualities of DNNs always give a wrong estimation of the discrimination ability of data, the ranking performance is poorer. For instance, in Java250 and C++1000, SDS only reaches 0.82 and 0.79 on Spearman's correlation with 20 labeled data, respectively. However, LaF achieves 0.96 and 0.95 in two datasets, respectively, with no labeling effort. On the other hand, concerning the diversity, Figure \ref{fig:impact} reveals that there is a high chance of a good ranking when DNNs are diverse (greater than 5\%). Additionally, a poor ranking mostly happens when DNNs are too close to each other, which confirms the result of iWildCam with ID data (Figure \ref{fig:clean-camera}) that all ranking methods obtain poor ranking.

\vspace{1mm}
\noindent\fcolorbox{black}{answercolor}{\begin{minipage}{\columnwidth} \textbf{Answer to RQ3}: By investigating the impact factors, model quality, and model diversity, of ranking performance, we observe that when the multiple DNNs have high quality (e.g., the average accuracy/ robustness is over 50\%), the performance of DNNs can be discriminated better. On the other hand, there is a higher chance of a good ranking when DNNs are more diverse (larger than 18\%).
\end{minipage}}

\section{Threats to validity}
\label{sec:threats}

The internal threat is mainly due to the implementation of the baseline methods, our proposed approach, and the evaluation metrics. For SDS, we use the original implementation on GitHub provided by Meng \emph{et al.}~\cite{sds2021meng}. For random sampling and CES, we implement it based on the description in~\cite{sds2021meng} and carefully check the result to be consistent with that in~\cite{sds2021meng}. Regarding the evaluation metrics, we adopt popular libraries, such as SciPy~\cite{scipy2022}.

The external threat comes from the evaluated tasks, datasets, DNNs, and baseline methods. Regarding the classification tasks, we consider three different ones, image, text, and source code. For the datasets, we select the publicly available datasets. In particular, for datasets with the artificial distribution shift (15 types of natural corruptions) and natural distribution shift, we employ four public benchmarks. Concerning the DNNs, we collect them (either the off-the-shelf models or train with the provided scripts) from different repositories on GitHub. These models are with different architectures and parameters. For the comparison, we consider three sample-selection-based baseline methods and apply different numbers of labeling budgets to imply their performance.

The construct threat mainly lies in the sampling randomness in the baseline methods and the evaluation measures. To reduce the impact of randomness, for each baseline method, we repeat each experiment concerning the labeling budgets, datasets 50 times and report the results of both average and standard deviation. Since our proposed approach does not rely on sampling data to annotate, there is no sampling randomness. Considering the randomness (gradient ascent search) in the EM algorithm, we repeat LaF 50 times and found that the randomness was negligible (less than 1.84E-03). Regarding the evaluation measures, we consider three popular statistical analyses. Kendall's $\tau$ rank correlation and Spearman's rank-order correlation can infer the effectiveness of the methods concerning the general ranking, while the Jaccard similarity can specifically check the performance concerning the top-$k$ ranking. Besides, for the statistical analyses, we report the $p$-value to demonstrate the significance.

\section{Related Work}
\label{sec:related}

We review the related work from two aspects, deep learning testing and test selection for deep learning. 

\subsection{Deep Learning Testing}

Deep learning (DL) testing refers to evaluating the quality of developed deep neural networks (DNNs) for further deployment~\cite{zhang2020machine}. A simple and local testing strategy is to split a dataset into training, validation, and test sets. The training and validation sets contribute to the training process to tune parameters. The test set is untouched by the training process to provide an unbiased evaluation of the accuracy. Typically, this testing is built on the assumption that the training and test sets are independent and identically distributed. 

Instead of simple performance testing, multiple advanced testing techniques have been proposed in recent years. Pei~\emph{et al.,}~\cite{pei2017deepxplore} proposed neuron coverage-guided testing for deep learning systems which borrows the idea from code coverage-based testing in traditional software engineering. Here, the coverage is calculated base on the outputs of neurons in DNN. After that, based on the basic neuron coverage criterion, Ma~\emph{et al.,}~\cite{ma2018deepgauge} designed different types of coverage criteria to further explore the coverage-guided testing. Based on these coverage criteria, DeepTest~\cite{tian2018deeptest} and TACTIC~\cite{li2021testing} tended to test DNN-based self-driving systems. Both of them utilize the coverage information to guide the search algorithm to generate error-prone test sets to challenge the target systems. Besides, the famous technique -- fuzzing testing was also applied for testing deep learning models. Odena~\emph{et al.,}~\cite{odena2019tensorfuzz} proposed the first fuzzing testing framework by randomly injecting noise into the image to generate new tests to find the error inputs against the DNN model. Xie~\emph{et al.,}~\cite{xie2019deephunter} used neuron coverage as fitness to fuzz the data and generate tests for the DNN testing. More practical, Guo~\emph{et al.,}~\cite{guo2018dlfuzz} provided a tool to support fuzzing testing of DL models. 

Different from the above works which focus on a single DNN model and utilize test data to measure the quality of the model, our work studies multiple models and provides a new technique to rank multiple models without label information.

\subsection{Test Selection for Deep Learning}

The purpose of test selection for deep learning is to reduce the labeling effort during DL testing. Generally, test selection methods can be divided into two types, test selection for fault identification and test selection for performance estimation. 

Test selection for fault identification is to find the test data that are most likely been mis-predicted by the model. Multiple methods have been proposed in the last few years. Feng~\emph{et al.,}~\cite{feng2020deepgini}, and Ma~\emph{et al.,}~\cite{ma2021test} proposed metrics based on the uncertainty of output probabilities, and also demonstrated that these metrics can be used to select data and retrain the pre-trained model to further enhance its performance. Chen~\emph{et al.,}~\cite{wang2021prioritizing} utilized the technique of mutation testing to mutate input data and models and select the error-prone test data based on the killing score. More recently, Li~\emph{et al.,}~\cite{li2021testrank} proposed a learning-based method that uses graph neural networks to learn the difference between the fault data and normal data and then predicts the new faults. Gao~\emph{et al.,}~\cite{gao2022adaptive} considered the diversity of faults and selected faults from different fault patterns.

Test selection for performance estimation aims to select a subset of data that can represent the whole test set. In this way, we can only label and test this subset and know the performance of the model on the entire test data. Li~\emph{et al.,}~\cite{li2019boosting} proposed CES, which selects samples that have the minimum cross entropy
with the entire test set. Chen~\emph{et al.,}~\cite{chen2020pace} proposed a clustering-based method PACE that selects data from the center of each cluster. 

Even though test selection for performance estimation can be also used for selecting the best model during the model reusing process, we considered it as our comparison baseline. The major difference between test selection and our proposed method LaF is -- LaF is labeling free which means the model resuing process can be fully automated.

\section{Conclusion}
\label{sec:conclusion}
Observing the limitations (labeling effort, sampling randomness, and performance degradation on out-of-distribution data) of existing selection-based methods, we proposed a labeling-free approach to undertake the task of ranking multiple deep neural networks (DNNs) without the need of domain expertise to lighten the MLOps. The main idea is to build a Bayesian model given the predicted labels of data, which allows for free labeling and non-sampling randomness. The experimental results on various domains (image, text, and source code) and different performance criteria (accuracy and robustness against artificial and natural distribution shifts) demonstrate that LaF significantly outperforms the three baseline methods concerning both Spearman's correlation and Kendall's $\tau$. In addition, the results of the Jaccard similarity show the efficiency of LaF in identifying the top-$k$ ($k=1,3,5,10$) DNNs. 

This work currently only focuses on the classification task, we will explore it for other tasks, such as regression, in future work. Observing the ranking difference on ID and out-of-distribution (OOD) test data, our approach might be useful to detect the existence of distribution shifts. We will consider this in future work.

\bibliographystyle{ACM-Reference-Format}
\bibliography{ref}

\end{document}